\documentclass{ecai} 

\usepackage{latexsym}
\usepackage{amssymb}
\usepackage{amsmath}
\usepackage{amsthm}
\usepackage{booktabs}
\usepackage{enumitem}
\usepackage{graphicx}
\usepackage{color}

\usepackage{amsmath,amsfonts,bm}

\def\eqref#1{equation~\ref{#1}}
\def\Eqref#1{Equation~\ref{#1}}

\def\1{\bm{1}}

\DeclareMathAlphabet{\mathsfit}{\encodingdefault}{\sfdefault}{m}{sl}
\SetMathAlphabet{\mathsfit}{bold}{\encodingdefault}{\sfdefault}{bx}{n}

\def\sD{{\mathbb{D}}}

\DeclareMathOperator*{\argmin}{arg\,min}

\usepackage{float}
\usepackage{algorithm,tabularx}
\usepackage{algorithmic}
\usepackage{makecell}
\usepackage{multirow}
\usepackage{tocloft}
\usepackage[dvipsnames]{xcolor}

\newtheorem{theorem}{Theorem}
\newtheorem{assumption}{Assumption}

\newcommand{\BibTeX}{B\kern-.05em{\sc i\kern-.025em b}\kern-.08em\TeX}
\newcommand{\OurAlgoPFN}[0]{PS-PFN}
\newcommand{\OurAlgoPFNCA}[0]{PS-PFN$^{cost}$}
\newcommand{\OurAlgoPFNs}[0]{PS-PFN$_{mixed}$}
\newcommand{\OurAlgoPFNsCA}[0]{PS-PFN$_{mixed}^{cost}$}
\newcommand{\OurAlgoMax}[0]{PS-Max}
\newcommand{\OurAlgoMaxCA}[0]{PS-Max$^{cost}$}
\newcommand{\OurAlgo}[0]{\OurAlgoPFN}

\newcommand{\MaxUCB}{\textit{MaxUCB}}
\newcommand{\RisingBandits}{\textit{Rising~Bandits}}
\newcommand{\RisingBanditsCA}{\textit{Rising~Bandits}$^{cost}$}
\newcommand{\ThompsonSampling}{\textit{Thompson~Sampling}}
\newcommand{\randomsearch}{\textit{random~search}}
\newcommand{\SMAC}{\textit{SMAC}}
\newcommand{\combinedsearch}{\textit{combined CASH}}
\newcommand{\decomposedCASH}{\textit{decomposed CASH}}
\newcommand{\CASHplus}{CASH+}

\newcommand{\Complex}[1][]{Complex\ifx\relax#1\relax\else[#1]\fi}
\newcommand{\Yahpogym}[1][]{YaHPOGym\ifx\relax#1\relax\else[#1]\fi}
\newcommand{\Tabreporaw}[1][]{TabRepoRaw\ifx\relax#1\relax\else[#1]\fi}
\newcommand{\Tabrepo}[1][]{TabRepoRaw\ifx\relax#1\relax\else[#1]\fi}
\newcommand{\Reshuffling}[1][]{Reshuffling\ifx\relax#1\relax\else[#1]\fi}
\newcommand{\numberOfBenchmarks}{three}

\definecolor{pink}{RGB}{247, 129, 191}
\newcommand{\pinksolid}{{\color{pink}\textbf{--}}}
\newcommand{\pinkdashed}{{\color{pink}\textbf{-~-}}}
\definecolor{orange}{RGB}{255, 127, 0}
\newcommand{\orangesolid}{{\color{orange}\textbf{--}}}
\newcommand{\orangedashed}{{\color{orange}\textbf{-~-}}}
\definecolor{blue}{RGB}{55, 126, 184}
\newcommand{\bluesolid}{{\color{blue}\textbf{--}}}
\newcommand{\bluedashed}{{\color{blue}\textbf{-~-}}}
\definecolor{brown}{RGB}{166, 86, 40}
\newcommand{\brownsolid}{{\color{brown}\textbf{--}}}

\definecolor{purple}{RGB}{152, 78, 163}
\newcommand{\purplesolid}{{\color{purple}\textbf{--}}}
\definecolor{green}{RGB}{77, 175, 74}
\newcommand{\greensolid}{{\color{green}\textbf{--}}}

\begin{document}


\begin{frontmatter}

\paperid{4390} 
\title{In-Context Decision Making for Optimizing Complex AutoML Pipelines}


\author[A]{\fnms{Amir}~\snm{Rezaei Balef }\thanks{Corresponding Author. Email: amir.balef@inf.uni-tuebingen.de.}}
\author[A]{\fnms{Katharina}~\snm{Eggensperger }}
\address[A]{Department of Computer Science, University of Tübingen}
\begin{abstract}
Combined Algorithm Selection and Hyperparameter Optimization (CASH) has been fundamental to traditional AutoML systems. However, with the advancements of pre-trained models, modern ML workflows go beyond hyperparameter optimization and often require fine-tuning, ensembling, and other adaptation techniques. 
While the core challenge of identifying the best-performing model for a downstream task remains, the increasing heterogeneity of ML pipelines demands novel AutoML approaches.
This work extends the CASH framework to select and adapt modern ML pipelines. We propose \OurAlgo{} to efficiently explore and exploit adapting ML pipelines by extending Posterior Sampling (PS) to the max $k$-armed bandit problem setup. \OurAlgo{} leverages prior-data fitted networks (PFNs) to efficiently estimate the posterior distribution of the maximal value via in-context learning. We show how to extend this method to consider varying costs of pulling arms and to use different PFNs to model reward distributions individually per arm. Experimental results on one novel and two existing standard benchmark tasks demonstrate the superior performance of \OurAlgo{} compared to other bandit and AutoML strategies. We make our code and data available at \url{https://github.com/amirbalef/CASHPlus}.
\end{abstract}

\end{frontmatter}

\section{Introduction} 
\label{Sec:Introduction}
\begin{figure}[tbp]
\centering
\includegraphics[height=2.0cm]{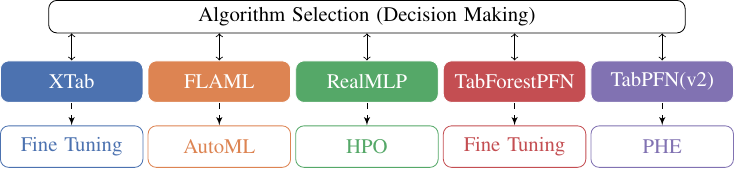}
\includegraphics[height=3.2cm]{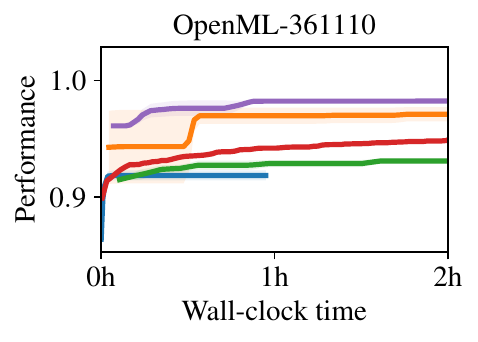}
\includegraphics[clip, trim=0.8cm 0cm 0cm 0.0cm, height=3.2cm]{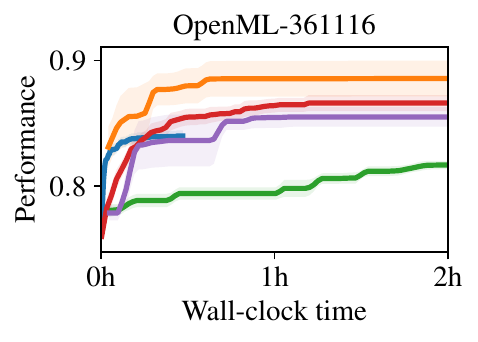}
\caption{(Top) AutoML needs to perform algorithm selection and resource allocation across heterogeneous optimization tasks, such as hyperparameter tuning, fine-tuning, ensembling, and more.
(Bottom) The performance of each workflow on two datasets demonstrates the variability of the optimization trajectories and the importance of algorithm selection.}
\label{fig:heterogeneous_pipeline}
\end{figure}

What model will perform best? The best-performing answer to this crucial question depends on the given task. We study this question in the context of predictive ML for tabular tasks, where possible model classes range from pre-trained models~\citep{Zhu2023XTabCP}, tree-based methods~\citep{chen-kdd16a,prokhorenkova-neurips18a}, modern deep learning approaches~\citep{holzmueller-neurips24a,gorishniy2021revisiting}, in-context learning with foundation models~\citep{hollmann2025accurate,ClsCondTabular}, and many more. Despite significantly different modelling approaches, with different implicit biases and strengths and weaknesses, all models have one thing in common: They must be adapted to the task at hand to perform best. This adaptation is typically an iterative procedure like optimizing hyperparameters, fine-tuning weights, learning embeddings, adapting the architecture, or running an AutoML system (see Figure~\ref{fig:heterogeneous_pipeline}).

A popular task for automatically allocating resources across workflows that include hyperparameter optimization (HPO) is the \textit{Combined Algorithm and Hyperparameter Optimization (CASH)}~\citep{thornton-kdd13a} problem. 
A common approach to the CASH problem is to model it as a single hierarchical HPO task over a joint search space that includes the choice among $K$ ML algorithms and their associated hyperparameters. This is typically done by treating the algorithm selection as a categorical hyperparameter and using conditional dependencies to define the model-specific subspaces.
State-of-the-art approaches either tackle this as a single-level optimization problem via Bayesian optimization on the combined hierarchical search space (we refer to this as \combinedsearch{}), as done by AutoML Systems~\citep{thornton-kdd13a,feurer-nips15a, komer-automl14a,kotthoff2017auto,feurer-jmlr22a}. However, this approach requires optimizing over a large, high-dimensional, and heterogeneous space using a single HPO method, which can be inefficient and difficult to scale. Alternatively, one can decompose it into a two-level optimization problem and leverage bandit methods at the top level and HPO methods at the lower level~\citep{li2020efficient,hu2021cascaded,balef2024towards, maxucb} (we refer to this as \decomposedCASH{}). This framework has also been recently adapted to other types of workflows, e.g., selecting and fine-tuning pre-trained models~\citep{arango2024quicktune, van2024selecting}. 

Here, we want to optimally allocate resources across different, heterogeneous workflows, with different performance distributions, optimization behaviours, and costs per iteration (see Figure~\ref{fig:heterogeneous_pipeline}). Existing \textit{combined and decomposed CASH} methods are not straightforward to extend since they were designed to perform well for a single type of workflow.

We address this issue using a bandit algorithm for resource allocation for complex AutoML pipelines that leverages flexible pre-trained machine learning models performing in-context learning to predict the posterior distribution for each arm. Our contributions are as follows:
\begin{enumerate}
    \item We introduce \CASHplus{} as a generalization of CASH, extending to any performance optimization strategy, covering modern ML workflows.
    \item We address this as a two-level optimization problem and at the top level as a Max $K$-armed Bandit (MKB) problem. We extend posterior sampling to optimize the maximal reward efficiently.
    \item We identify prior-data fitted networks (PFNs)~\citep{muller-iclr22a} as a flexible model class to enable posterior sampling in practice and analyze HPO trajectories and learning curves to design effective priors. We call the resulting method \OurAlgo{}.
    \item We extend \OurAlgo{} to handle varying costs and to use different PFNs per arm.
    \item Finally, we demonstrate superior empirical performance of \OurAlgo{} on \numberOfBenchmarks{} benchmarks where we compare \OurAlgo{} with state-of-the-art Multi-Armed Bandits (MAB) algorithms developed for AutoML.
\end{enumerate}

We start by discussing our problem setup and introducing \CASHplus{} in Section~\ref{Subsec:Problem_Formulation}. Section~\ref{sec:background} provides a short introduction to posterior sampling. Next, we introduce \OurAlgo{} in Section~\ref{Sec:Methodology}, and evaluate it in Section~\ref{Sec:Numerical_Experiments}. We conclude by discussing its limitations and future work in Section~\ref{Sec:Conclusion}.

\section{Combined Algorithm Selection and Optimization (\CASHplus{})}
\label{Subsec:Problem_Formulation}
The CASH problem, at the core of traditional AutoML pipelines~\citep{thornton-kdd13a}, involves selecting the optimal ML algorithm \( A^{(i^*)} \) and its hyperparameters \( \pmb{\lambda}^* \) to minimize a loss metric \( \mathcal{L} \) (e.g., validation error). Given a dataset \( \sD = \{ D_{train}, D_{valid} \} \) and a set of \( K \) candidate algorithms \( \mathcal{A} = \{A^{(1)}, ..., A^{(K)} \} \), each with its own hyperparameter space \( \pmb{\Lambda}^{(i)} \), the goal of CASH is to optimize over the joint algorithm-hyperparameter space. Here, we generalize this to any iterative optimization procedure where the goal is to select the optimal ML algorithm \( A^{(i^*)} \) and its state \( \pmb{s}^* \) as found by its optimization procedure.  Formally,

\begin{equation}
A^{(i^*)}_{\pmb{s}^{*}} \in \argmin_{ A^{(i)} \in \mathcal{A}, \pmb{s} \in \pmb{S}^{(i)}  } \mathcal{L}(A^{(i)}_{\pmb{s}}  ,\sD).
\end{equation}

\CASHplus{} extends the traditional AutoML pipeline by incorporating optimization techniques beyond standard model selection and hyperparameter tuning. We note that this is very similar to the original definition of CASH as introduced by~\citet{thornton-kdd13a}, but explicitly allows for including \textit{any} optimization method, thus covering methods like hyperparameter tuning, fine-tuning, and ensembling. These methods often involve highly diverse search spaces. For example, fine-tuning neural networks typically has a large and complex space, while Post-Hoc Ensembling (PHE) may not have a well-defined search space at all. In such cases, \combinedsearch{} approaches are no longer applicable due to the incompatibility of the search spaces and, thus, search methods. Instead, similar to prior work~\citep{li2020efficient,hu2021cascaded,balef2024towards, maxucb}, we tackle \CASHplus{} as a two-level optimization problem (as depicted in Figure~\ref{fig:heterogeneous_pipeline}), which we formally define as:

\begin{align}
&A^{(i^*)} \in \argmin_{A^{(i)} \in \mathcal{A} } \mathcal{L}(A^{(i)}_{\pmb{s}^*} ,\sD) 
\label{eq:upper_level_optimization} \\
&\text{s.t.  } \pmb{s}^{*} \in \argmin_{ \pmb{s} \in \pmb{S^{(i)}} } \mathcal{L}( A^{(i)}_{\pmb{s}}  ,\sD).
\label{eq:lower_level_optimization}
\end{align}
At the upper level, we aim to find the overall best-performing ML model $A^{(i^*)}$ by selecting model  $A^{(i)} \in \mathcal{A}$, and at the lower level, we optimize the selected model $A^{(i)}$ iteratively.

\paragraph{\CASHplus{} as a Bandit problem.} To approach the upper-level decision-making task (see Figure~\ref{fig:heterogeneous_pipeline}) with a bandit method, we denote $r_{i,t}$ to be the feedback to arm $i$ obtained by evaluating $\pmb{s}_t \in \pmb{S}^{(i)}$ at time step $t\le T$. To be consistent with the bandit literature, we consider the negative loss as the reward and maximize it: %
\begin{equation}
r_{i,t} =  -\mathcal{L}( A^{(i)}_{\pmb{s}_t}  ,\sD). 
\end{equation}
The goal is then to find the best-performing algorithm $A^{(i^*)}$ and its optimal state $\pmb{s}^*$ given a time horizon $T$. This can be framed as minimizing the regret $R(T)$ with $I_t$ being the selected arm at time $t$ by:
\begin{equation}
R(T) = \max_{k\leq K}\mathop{\mathbb{E}[\max r_{k,t}]}_{t \leq T} - \mathop{\mathbb{E}[\max r_{I_t,t}]}_{t \leq T}. 
\label{eq:extremebandit_regret}
\end{equation}

This regret describes the gap between the highest possible oracle reward that could be obtained by pulling only the arm with the highest performance (left part) and the actual observed rewards obtained by applying our strategy (right part). 

\section{Background}
\label{sec:background}
The two-level optimization problem allows each model to be optimized independently. Existing bandit algorithms used in the \decomposedCASH{} literature generally assume that all arms (i.e., optimization methods) share similar characteristics \citep{li2020efficient,hu2021cascaded,balef2024towards, maxucb}. However, in the \CASHplus{} setting, the arms can differ significantly in their reward processes and costs, making those assumptions invalid. For this, we explore posterior sampling as a much more flexible framework for developing our algorithm.

\subsection{Posterior Sampling} 
Posterior sampling, first introduced in 1933 \citep{thompson1933likelihood}, has become a fundamental technique in the field of MAB, which is also known as Thompson Sampling (TS). This approach maintains a posterior over reward models and samples from it at each decision point to select actions.  Specifically, the uncertainty in each arm's reward model is captured by a predictive distribution $p_{\theta_i}(r \mid \varPhi)$, where $\theta_i$ represents the parameters of arm $i$'s model and $\varPhi$ denotes the information observed from the environment. As observations accumulate, the agent updates its belief about the predictive distribution and selects the arm with the highest sampled value from the posterior.

Posterior sampling is highly flexible, as it does not require closed-form solutions or restrictive assumptions. It can be applied with any probabilistic model, provided that sampling from the (possibly approximated) posterior is feasible \citep{kveton2024online}.

In Algorithm~\ref{alg:pseudocode_Posterior_Sampling} at each round $t$, the agent maintains a dataset $D_i$ of all observed rewards from arm $i$ and models the posterior distribution accordingly. It then samples $s_i \sim p(\cdot \mid D_i, f(t))$ from the posterior distribution. The function $f(t)$ can account for potential temporal dependencies in the reward process, such as rested or non-stationary rewards \citep{liu2023nonstationary}, which can be viewed as using time information as context for modeling reward processes in contextual bandits \citep{Shen2023HyperBanditCB}. After pulling arm $I_t$, the agent observes the reward $r_{I_t, n_{I_t} + 1}$ and updates the dataset $D_{I_t}$ with the new observation.

Posterior sampling relies fundamentally on the agent's ability to reason about the posterior distribution over rewards. This leads to optimal performance in maximizing the expected \textit{cumulative} reward in classical bandit settings \citep{russo2014learning}. 
In contrast, the goal in our setting is to optimize the expected \textit{maximum} observed reward (see Equation~\ref{eq:extremebandit_regret}), which aligns with the MKB framework. This shift in objective requires reasoning about the posterior distribution of maximum values across arms, introducing new challenges. In this paper, we develop approximation methods designed to estimate this distribution effectively.

\begin{algorithm}[htbp]
\scriptsize
\caption{Posterior sampling (Thompson Sampling)}
\label{alg:pseudocode_Posterior_Sampling}
\begin{algorithmic}[1]
\REQUIRE time horizon $T$, time function $f(t)$, number of arms $K$
\FOR {each arm $i \in \{1, \dots, K\}$}
    \STATE Pull arm $i$ once; observe reward $r_{i,1}$
    \STATE Initialize dataset $D_i \gets \{(1, r_{i,1})\}$ and counter $n_i \gets 1$
\ENDFOR

\FOR {$t = K + 1$ to $T$}
    \FOR {each arm $i \in \{1, \dots, K\}$}
        \STATE Draw \textcolor{blue}{$s_i \sim p(\cdot \mid D_i, f(t))$} \COMMENT{Sample from posterior}
    \ENDFOR
    \STATE Select arm $I_t \gets \arg\max\limits_{i \in \{1,\dots,K\}} s_i$
    \STATE Pull arm $I_t$; observe reward $r_{I_t, n_{I_t} + 1}$
    \STATE $n_{I_t} \gets n_{I_t} + 1$
    \STATE Update $D_{I_t} \gets D_{I_t} \cup \{(n_{I_t}, r_{I_t, n_{I_t}})\}$
\ENDFOR
\end{algorithmic}
\end{algorithm}

\subsection{Model-based Posterior Approximation} 
Computing the exact posterior distribution is often analytically or computationally infeasible for complex reward models. Consequently, approximate inference methods such as Markov Chain Monte Carlo or Variational Inference are commonly employed to estimate posterior distributions in practice \citep{phan2019thompson}.

Recently, transformers have demonstrated the capability to perform in-context learning (ICL) when trained on large amounts of data~\citep{brown-neurips20a}. ICL methods are very sample efficient, yielding high accuracy with only a few observations, making them very suitable for tasks with short time horizons.

So-called Decision Transformers (DTs) reframe reinforcement learning (RL) as a sequence modeling task, by conditioning on desired returns, past states, and actions~\citep{chen2021decision}. DTs predict future actions and leverage ICL to generalize with minimal adaptation. \citet{Lin2024TransformersAD} studies the ICL capability of pretrained transformers for RL in theory. Recent work shows that Decision Pretrained Transformers can approximate Bayesian posterior sampling over actions, enabling principled exploration to solve bandit problems \citep{lee2023supervised}. While prior work focuses on predicting optimal actions, our approach instead models the posterior distribution over rewards to achieve better scalability, for example, with respect to the number of actions.

\section{Methodology} 
\label{Sec:Methodology}
Accurately estimating the posterior distribution of the maximum reward is crucial for effective decision-making in the MKB settings. We first show how to extend posterior sampling to the MKB setting by finding the posterior distribution of maximum values under the simplified assumption of i.i.d. rewards and sufficiently fast concentration around the maximum. Since this assumption can be limiting for addressing \CASHplus{}, we discuss ICL using prior-fitted networks as a more flexible and expressive approximation of the posterior.

\subsection{ \OurAlgoMax: Posterior Sampling for Max $K$-armed Bandit}
We need to accurately predict the posterior distribution of the maximum reward to extend classical Thompson Sampling to our setting. Based on Algorithm~\ref{alg:pseudocode_Posterior_Sampling} and the regret definition (\Eqref{eq:extremebandit_regret}), we introduce \OurAlgoMax{} for $K$ arms with a limited time horizon of $T$.  Let \(r_{i,1:t}\) denote sequence of rewards from time $1$ till time $t$.

\begin{assumption} We assume that the i.i.d. random variable \(r_{i,t}\), representing the reward of pulling arm $i$ at time $t$, follows a sub-Gaussian distribution with cumulative distribution function (CDF) $F(x)$.
\label{theorem:assumption_1}
\end{assumption}

\noindent When Assumption~\ref{theorem:assumption_1} is held, line 7 in Algorithm~\ref{alg:pseudocode_Posterior_Sampling} can be replaced by:
\begin{equation}
     s_{i} = F_i^{-1}(U^{1/t}), \quad U \sim \mathrm{Uniform}(0,1)
     \label{eq:posterior_of_max}
\end{equation}
where $f(t)=t$ and $F_i^{-1}$ is the inverse CDF of our sub-Gaussian prior for arm $i$. We call the resulting method \textbf{\OurAlgoMax}.

\begin{theorem}[Analysis of \OurAlgoMax]
\begin{enumerate}
     If the expected maximum satisfies:
    \[
    \mathbb{E}[ \max (r_{i,1:t})] \geq F_i^{-1}\left(1 - \frac{1}{t^2}\right),
    \]
    then the number of times sub-optimal arm $i$ is pulled grows logarithmically with the time horizon.
\end{enumerate}
\label{theorem:max_sampling}
\end{theorem}

\OurAlgoMax{} operates similarly to classical Thompson Sampling. At each iteration, it updates the sub-Gaussian prior based on observed rewards and then draws samples from the posterior distribution of the maxima (\Eqref{eq:posterior_of_max}) rather than from the posterior distribution of the sub-Gaussian.
For now, we choose  $f(t)=t$ and assume constant costs to ensure anytime performance, as this choice aligns with the objective of identifying the arm with the highest expected maximum at each time step $t$. \footnote{In MKB problems, there is no single best arm and the oracle arm depends on the budget $T$~\citep{nishihara2016no}. If $T$ is known and sufficiently large, with $f(t)=T$, the agent aims to achieve the best final performance. With $f(t)= n_i + T-t$, the agent accounts for how many more iterations it can pull a given arm.}

We provide the proof of Theorem~\ref{theorem:max_sampling} in Appendix~\ref{app:preliminaries:theorem_1_proof}. 
MKB algorithms can surely achieve optimal performance \textit{only} when the reward distribution aligns with their assumptions. The condition in Theorem~\ref{theorem:max_sampling} highlights that the distribution must have a very light tail near its maximum. In CASH problems, because of the left-skewed reward distribution and almost non-stationarity, the conditions of Assumption~\ref{theorem:assumption_1} and Theorem~\ref{theorem:max_sampling} are often satisfied \citep{maxucb}.

\begin{figure}[htbp]
\centering
\includegraphics[clip, trim=0.0cm 0.2cm 0cm 0.0cm,height=3.0cm]{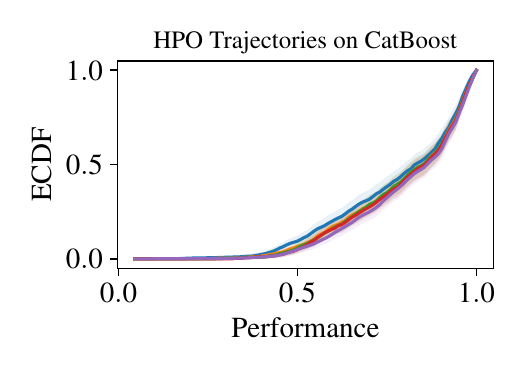}
\includegraphics[clip, trim=1.0cm 0.2cm 0cm 0.0cm, height=3.0cm]{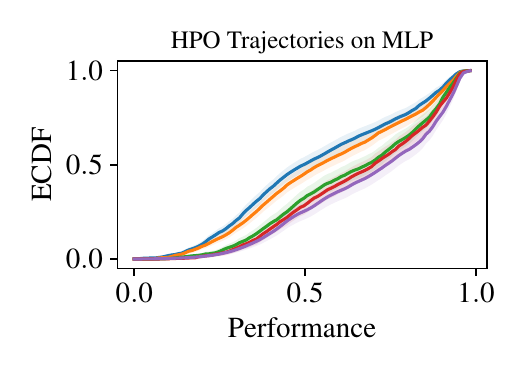}
\includegraphics[clip, trim=0.0cm 0.0cm 0.0cm 0.2cm, height=0.4cm]{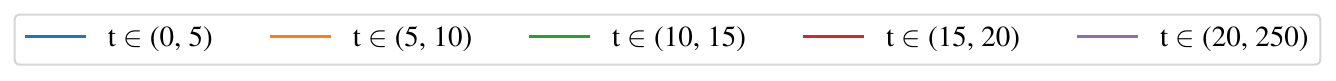}
\caption{Exemplary HPO trajectories (from the \Reshuffling{} benchmark) exhibit distributions that standard parametric models cannot capture.}
\label{fig:HPO_cdf}
\end{figure}

However, for \CASHplus{} tasks, the heterogeneity of the lower-level optimization methods results in rewards drawn from different distributions. This brings up three main limitations to overcome:
\begin{itemize}
    \item \textbf{The reward distributions do not follow a common form} and exhibit a highly negatively skewed distribution as shown on the left side of Figure \ref{fig:HPO_cdf}. 
    \item \textbf{The reward distribution varies across the arms} as different optimization methods can induce distinct reward processes.
    \item \textbf{The reward distribution may shift over time} when pulling the same arm (rested setting) as shown on the right side of Figure \ref{fig:HPO_cdf} where the reward distribution drifts over time. 
\end{itemize}

Addressing these limitations is vital to address \CASHplus{}. As a consequence, extending \OurAlgoMax{} for complex reward processes requires addressing three key challenges: 
(1) the presence of heterogeneous reward distributions \citep{baudry2021optimality}; (2) changes in the reward process over time \citep{fiandri2024thompson}; and (3) varying costs associated with pulling different arms \citep{Xia2015ThompsonSF}. 
To tackle them, we employ in-context learning through PFNs, as we discuss next.

\subsection{\OurAlgoPFN{}: Posterior sampling using PFNs} 
We need to accurately predict the posterior distribution of the maximum reward while accounting for changes in the reward distribution and left skewness of the distributions. We achieve this using PFNs~\citep{muller-iclr22a}.
PFNs have demonstrated strong performance, particularly for predictive ML tasks on small tabular datasets~\citep{hollmann-iclr23a,hollmann2025accurate}, and have been successfully applied in related tasks such as learning curve extrapolation~\citep{adriaensen2023efficient} and Bayesian optimization~\citep{muller2023pfns4bo}.
 
PFNs are pre-trained on synthetic data generated from a prior to perform approximate Posterior Predictive Distributions (PPD), without explicit retraining for new tasks. In other words, they are trained to predict some output $y$, conditioned on an input $n$ and a training set $D_{train}$ of given input-output examples. \citet{muller-iclr22a} proved that minimizing this loss over many sampled tasks $(n, y) \cup D_{train}$ directly coincides with minimizing the KL divergence between the PFN\textquotesingle{s} predictions and the true PPD. In essence, the PFN meta-learns to perform approximate posterior inference on synthetic tasks sampled from the prior, and at inference time, also does so for an unseen and real task.
We focus on this model class since PFNs particularly perform well on small data tasks, enable instant predictions due to in-context learning, can be efficiently trained entirely on synthetic data, and the prior design allows incorporating assumptions and prior knowledge about data we expect to observe in the real world.

\begin{figure}[htbp]
\centering
\includegraphics[height=4cm]{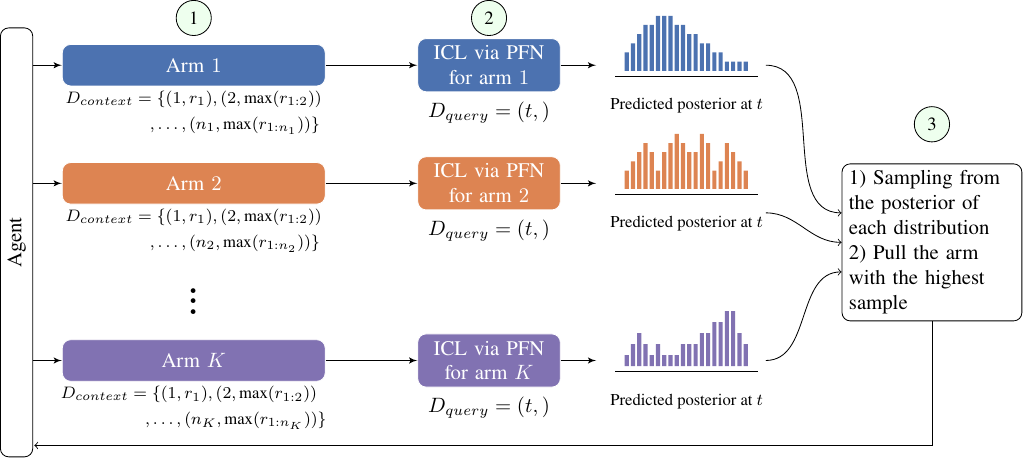}
\caption{A single iteration of \OurAlgo{}. (1) Context construction: Format observed rewards as PFN input; (2)  Posterior prediction: Query PFN for time step $t$; (3) Decision: Sample from posterior to select arm. }
\label{fig:ps_pfn}
\end{figure}

We use PFNs to model and estimate the unknown per-arm reward distributions as shown in Figure~\ref{fig:ps_pfn}. Specifically, we use the maximum of the reward as output $y = \max(r_{i,1:t})$. For a sequence of rewards (and potentially additional information) from arm $i$ with $n_i$ observations, we obtain $D_{i, 1:n_i}=\{(1,r_{i,1}), (2,\max(r_{i,1:2})),\ldots, (n_i,\max(r_{i,1:n_i}))\}$. We then apply in-context learning with PFNs, using the observed data $D_{i, 1:n_i}$ as the context to output the posterior distribution $p(\max(r_{i,1:t})\mid D_{i, 1:n}, t)$. This distribution is then used to guide the agent's next arm selection.

To apply PFNs to our setup, we need the key ingredients: (a) a pre-training task that matches our problem formulation and (b) a prior to generate synthetic reward sequences that are close to what we expect to observe in the real world.

\paragraph{Architecture and pre-training task for reward distribution estimation.} We use the same architecture used by \citet{muller-iclr22a} which is a sequence Transformer~\citep{vaswani-neurips17a}. For training, we use $D_{train}=\{(1,r_{1}), (2,\max(r_{1:2})),\ldots, (n_i,\max(r_{1:n_i}))\}$ as context and $D_{test}= \left(t,\right)$ as a test output. Following \citet{adriaensen2023efficient}, we use a similar setup; we omit positional encodings since the input already includes a positional feature, and we mask the attention matrix so that each position only attends to training data positions. 
However, instead of uniformly sampling the cutoff points to split trajectories into training and test datasets for each batch, we use a harmonic distribution to assign more weight to smaller cutoff points. This is because predictions with fewer observations are more crucial in our setting. We provide more details on our architecture and training in Appendix~\ref{app:priors_architecture}.

Once pre-trained, the PFN (with parameters $\theta$) outputs a discretized distribution $P_{\theta}(\max(r_{i,1:t}) \mid D_{\text{test}}, t)$ with a fixed number of bins, and predicting the probability mass for each bin~\citep{muller-iclr22a}. We will use this distribution to generate samples for line 7 in Algorithm~\ref{alg:pseudocode_Posterior_Sampling} using $s_i \sim P_{\theta}(\cdot \mid D_{i, 1:n_i}, t)$.

\begin{figure}[tb]
\centering
\begin{tabular}{@{\hskip 1mm} c @{\hskip 1mm} c @{\hskip 1mm} c}
\footnotesize {\quad Flat} & \footnotesize {Semi-flat} & \footnotesize {Curved} \\

\includegraphics[clip, trim=0.0cm 0.2cm 0.2cm 0.2cm,height=2.3cm]{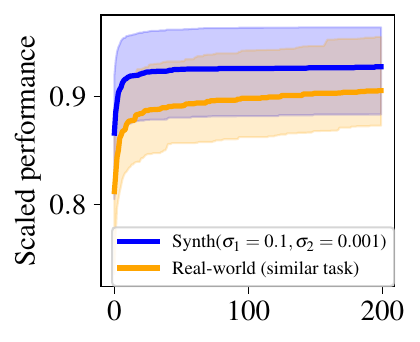} &
\includegraphics[clip, trim=0.8cm 0.2cm 0.2cm 0.2cm, height=2.3cm]{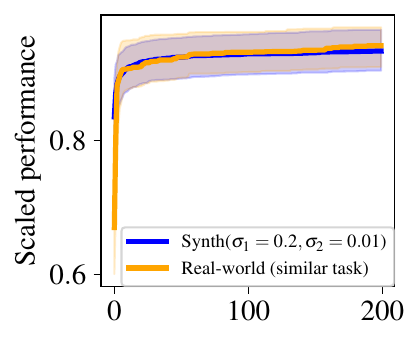} &
\includegraphics[clip, trim=0.8cm 0.2cm 0.2cm 0.2cm, height=2.3cm]{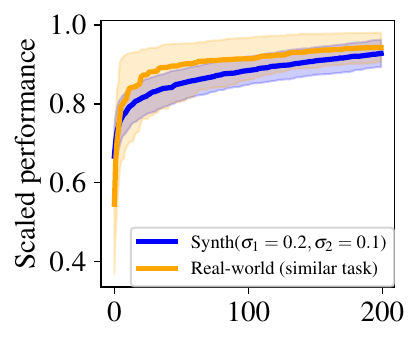} \\

\includegraphics[clip, trim=0.0cm 0.2cm 0.2cm 0.2cm, height=2.3cm]{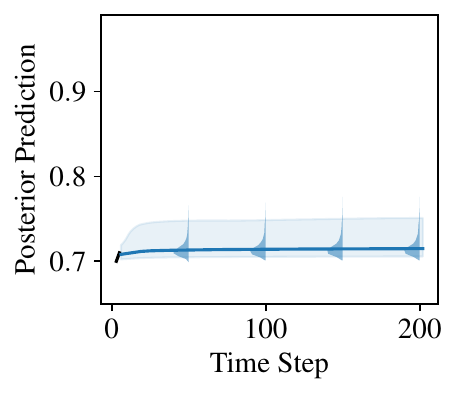} &
\includegraphics[clip, trim=0.8cm 0.2cm 0.2cm 0.2cm, height=2.3cm]{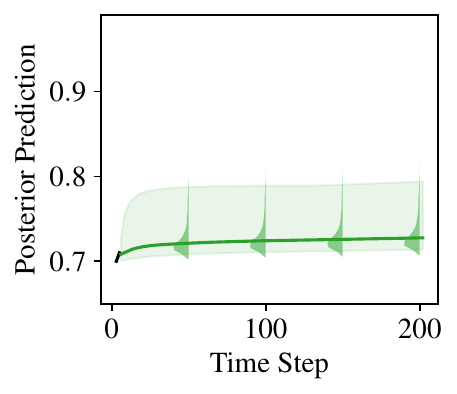} &
\includegraphics[clip, trim=0.8cm 0.2cm 0.2cm 0.2cm, height=2.3cm]{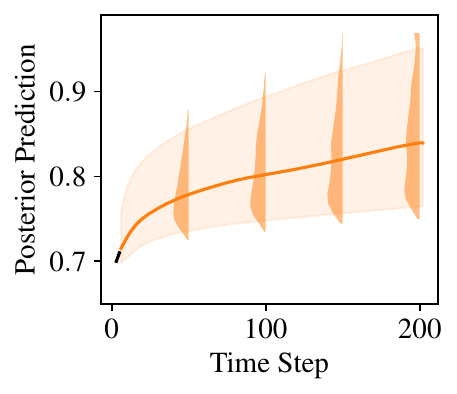}
\end{tabular}
\caption{(Top) Different optimization trajectories for \CASHplus{} tasks and their synthetically generated counterparts. (Bottom) Corresponding posterior predictions by PFNs given the same input (black line).}
\label{fig:HPO_landscapes}
\end{figure}

\paragraph{A prior for reward distributions.} 
The prior design is critical to the effectiveness of PFNs~\citep{breejen2024fine}, and our reward distributions heavily depend on the optimization landscape of the AutoML task and can vary considerably in structure and complexity; thus, we define three priors (flat, semi-flat, and curved) to generate synthetic data for training PFNs. The resulting PFNs cover common characteristics of optimization problems:
\begin{enumerate}
    \item The \textbf{flat prior} covers optimization landscapes common in HPO, which are characterized by large plateaus in the objective space, where many hyperparameter configurations yield similar performance~\citep{pushak-acm22a}. As a result, the distribution of observed rewards tends to be left-skewed, as most configurations cluster around suboptimal yet similar values. Importantly, there is no significant shift in the distribution over time (see Figure~\ref{fig:HPO_landscapes}, left).
\item The \textbf{semi-flat prior} covers trajectories where rewards gradually improve over time, reflecting a shift with a rapid decay in the distribution (see Figure~\ref{fig:HPO_landscapes}, middle), for example, when the search space for optimization is large. This means that the magnitude of the shift is larger in the initial steps and gradually decreases over time when the optimization converges.
\item The \textbf{curved prior} covers scenarios where the amount of shift is considerably high and does not decay rapidly (see Figure~\ref{fig:HPO_landscapes}, right).  This behavior can be observed when training neural networks from scratch.
\end{enumerate}

The semi-flat prior serves as a robust and broad prior covering most settings. The flat prior should be reserved for settings where the lower-level optimization converges fast and only little exploration is needed to approximate the posterior distribution. In contrast, the curved prior should be used for workflows that require more exploration to reduce uncertainty.

\paragraph{Generating synthetic reward trajectories.} To train our PFNs, we must generate synthetic trajectories for each prior. We implement this using two distributions $d_1$ and $d_2$.

The first distribution, $d_1$, models the rewards and captures characteristics of the optimization output and its uncertainty, i.e., left skewness when rewards are i.i.d.\, as is typical with random search. The second distribution, $d_2$, models the shift over time and introduces non-stationarity by modeling changes in the reward distribution over time. 
The reward sequence is generated by element-wise multiplication of the samples from $d_1$ with the sorted samples from $d_2$:

\begin{align}
    r_{1,1:t} &\sim d_1, \quad r_{2,1:t} \sim d_2, \\
    r_{1:t} &= r_{1,1:t} \cdot \mathrm{sort}(r_{2,1:t}).
    \label{eq:trajectories_generation}
\end{align}

Concretely, to generate a synthetic trajectory, we draw one sample for each parameter of the distributions, e.g., $\mu_1$. We then draw $t = 200$ samples from $d_1$ and $d_2$, denoted by $r_{1,1:t}$ and $r_{2,1:t}$, respectively. Next, we sort $r_{2,1:t}$ in ascending order and perform element-wise multiplication with $r_{1,1:t}$ to obtain the final reward sequence $r_{1:t}$.

Choosing appropriate distributions for $d_1$ and $d_2$ and the range of their parameters is challenging, and suitable values depend on the task at hand. We derive suitable ranges based on holdout tasks as detailed in Appendix~\ref{app:trajectory_analysis}.

\paragraph{Distribution \(d_1\) (Reward Uncertainty)}
The distribution \(d_1\) models the uncertainty in the rewards and is designed to be \textbf{left-skewed}.
We use a truncated skewed normal distribution over the interval \([0, 1]\), where the mean $\mu_1$ is sampled from \(\text{Uniform}(0, 1)\) (see Appendix~\ref{app:out_of_distribution}), the skewness is sampled from \(\text{Uniform}(-100, -20)\), and the standard deviation \(\sigma_1\) controls the level of model uncertainty.

\paragraph{Distribution \(d_2\) (Non-Stationarity)}
The distribution \(d_2\) captures non-stationarity by introducing time-dependent shifts in the reward distribution.
Similar to $d_1$, it is also modeled using a truncated skewed normal distribution over \([0, 1]\), with a fixed mean of \(1\), skewness sampled from \(\text{Uniform}(-100, -20)\), and standard deviation \(\sigma_2\) controlling the severity of the non-stationary behavior. 

For each prior, we specify different values for \(\sigma_1\) and \(\sigma_2\) as summarized in Table~\ref{tab:trajectory_types}. Figure~\ref{fig:HPO_landscapes} shows the posterior predictions of these three PFNs for the same input sequence. As seen, the output closely approximates the synthetic trajectories. Additionally, in Appendix~\ref{app:priors_importance}, we illustrate the sensitivity of our PFNs to prior selection by evaluating PS under different assumptions. Notably, the term $(1 - \mu_1)$ is used because reward processes with a low-performing mean tend to exhibit greater non-stationarity.

\begin{table}[ht]
\centering
\caption{Parameters for the distributions used to generate synthetic reward trajectories for each prior.
\label{tab:trajectory_types}}
\scriptsize
\begin{tabular}{lll}
\toprule
\textbf{Type} & \multicolumn{2}{l}{\textbf{Parameters}} \\
\midrule
Flat& \(\sigma_1 \sim \text{Uniform}(0, 0.1)\) & \(\sigma_2 \sim \text{Uniform}(0.001(1 - \mu_1),  0.001)\)  \\
Semi-flat& \(\sigma_1 \sim \text{Uniform}(0, 0.2)\) & \(\sigma_2 \sim \text{Uniform}(0.01 (1 - \mu_1),  0.01)\)  \\
Curved& \(\sigma_1 \sim \text{Uniform}(0, 0.2)\) & \(\sigma_2 \sim \text{Uniform}(0.1 (1 - \mu_1),  0.1)\)\\
\bottomrule
\end{tabular}
\end{table}

\subsection{Cost-awareness and handling budgets}
Finally, we need to address varying costs for pulling different arms. In classical bandit settings, cost-awareness or budgeted bandits are typically incorporated by normalizing the reward by the cost, i.e., maximizing the reward-to-cost ratio \citep{ding2013multi, xia2015thompson, xia2016budgeted}. This approach is justified because pulling the arm with the highest reward rate decreases the cumulative regret the most \citep{cayci2020budget}. A similar heuristic is used in Bayesian optimization, where improvement-based acquisition functions are divided by (the logarithm of) cost \citep{snoek2012practical}; however, it may underperform without proper controlling \citep{Xie2024CostawareBO}. %
In our setting, regret is only reduced when a higher reward than previously observed is obtained, making this strategy less appropriate.

Instead, we propose a different perspective: rather than adjusting the reward process by a cost estimate, we predict the posterior distribution of the maximal reward for a future time step adjusted by the remaining budget.

\begin{figure}[htp]
\centering
\includegraphics[height=3cm]{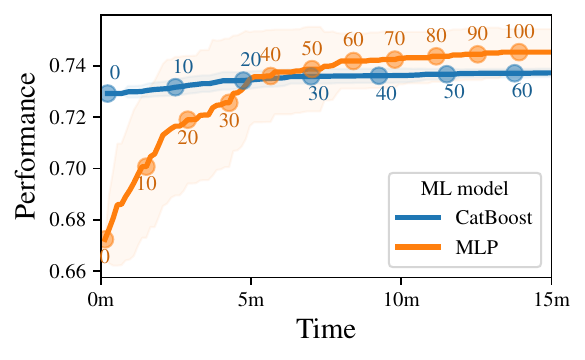}
\includegraphics[height=3cm]{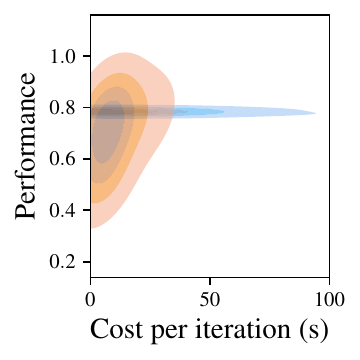}
\caption{(Left) for 30 iterations of HPO, CatBoost outperforms MLP, while with the same budget, it is possible to run MLP for 50 iterations, outperforming CatBoost. (Right) The cost of one iteration is noisy.}
\label{fig:cost_awareness}
\end{figure}

Figure~\ref{fig:cost_awareness} exemplarily shows that, for a fixed budget, e.g., $5$ minutes, MLP achieves over $30$ pulls (iterations), whereas CatBoost can be pulled only $20$ times since each iteration has a higher cost (runtime). This means that the maximum number of pulls, i.e., the time step $t$ for which we want to predict the distribution of maximal reward, varies across models.
To consider this in our algorithm and to estimate the effective posterior for each arm, we must modify $f(t)$ in Algorithm \ref{alg:pseudocode_Posterior_Sampling} based on the remaining pulls per arm.

To estimate costs per pull per arm, we introduce $B$ as the total (time) budget. Since runtimes can be stochastic, we model them as random variables and sample from their distribution e.g. log-normal distribution. Specifically, the cost of pulling arm $i$ is $c_{i,t}$, and we adjust our decision-making (exploration vs. exploitation) based on both the observed empirical cost and the spent budget $b$. Let $c \sim p_c(\cdot \mid \{c_{i,t}\}_{t=1}^{n_i}) $ denote a sample draw from the \textit{posterior distribution of the cost} for arm~$i$, given observed costs $\{c_{i,t}\}_{t=1}^{n_i} $. We then define the corrected time function as:

\begin{equation}
    f_i(t) =  n_i + \frac{B - b} { c \sim p_c(\cdot \mid \{c_{i,t}\}_{t=1}^{n_i} )}
    \quad\text{where } \quad b=\sum_{i \leq k}\sum_{t = 1}^{n_i} c_{i,t} 
\end{equation}

The intuition behind this definition is as follows: arm $i$ has already been pulled $n_i$ times, and $B-b$ represents the remaining budget. By dividing the remaining budget by the cost, we estimate the maximum number of additional pulls that can be performed. This allows us to predict the posterior distribution corresponding to that future iteration (see Appendix~\ref{app:cost_awareness} for more details).

\begin{figure*}[htb]
    \begin{tabular}{c c c c c}
    \hspace{-0.5em}\small\Complex{}
    & \hspace{-1em}\small\Tabreporaw{[SMAC]}
    &\hspace{-1.5em}\small\Yahpogym{[SMAC]}
    & \\
    \includegraphics[clip, trim=0.2cm 0cm 0.2cm 0.25cm ,height=4.3cm]{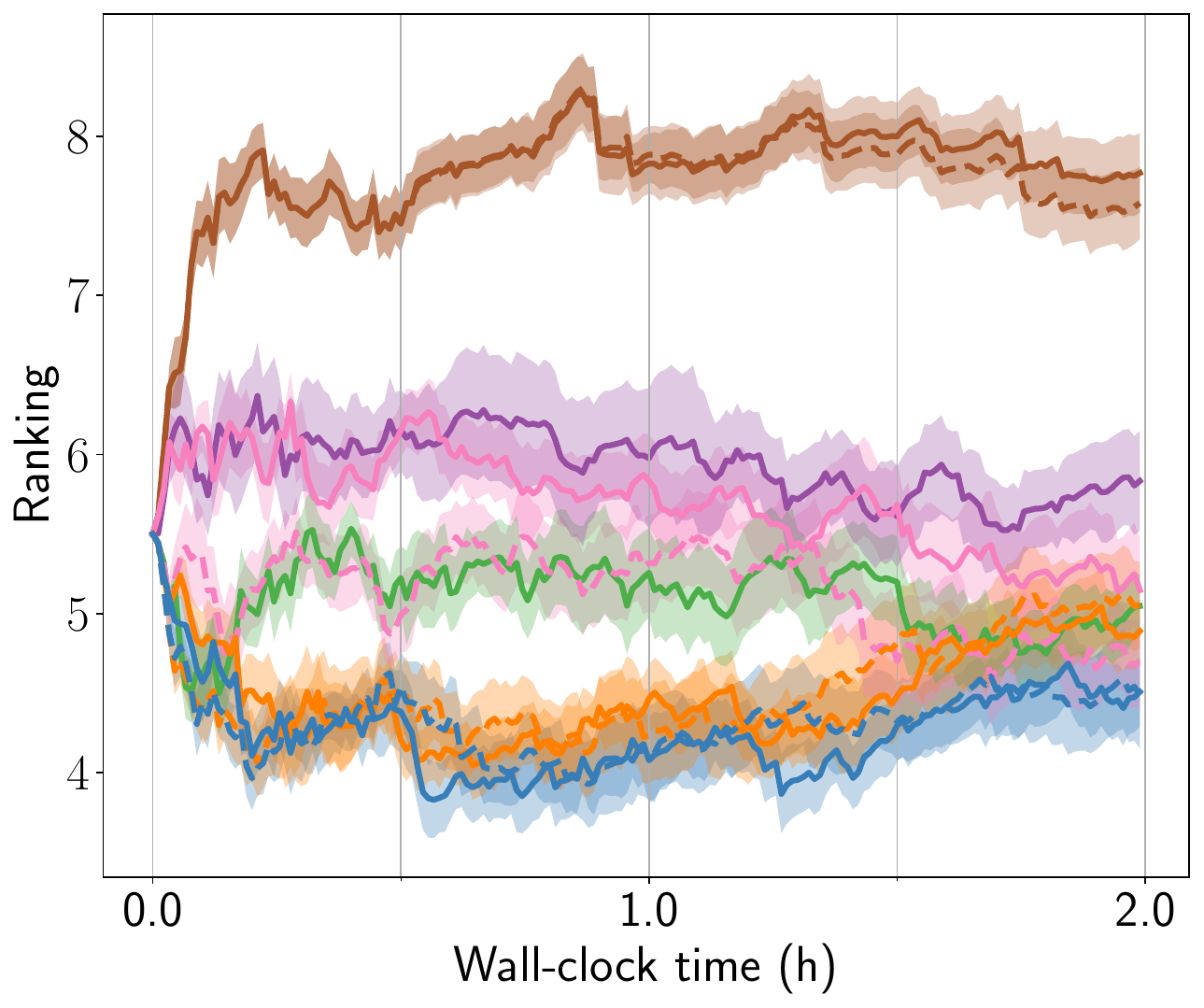}
    & \hspace{-1em} \includegraphics[clip, trim=1.1cm 0cm 0cm 0.25cm, height=4.3cm]{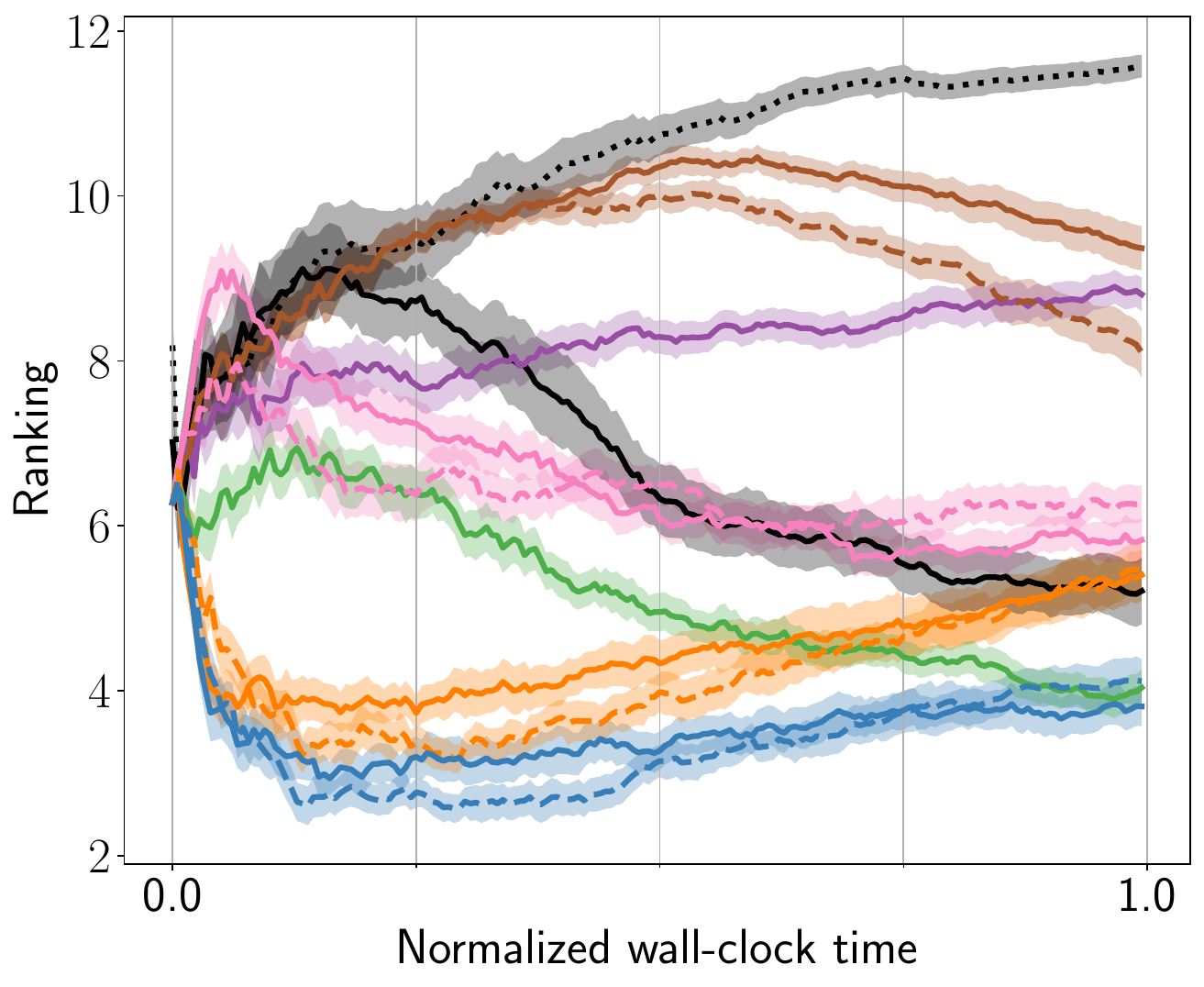}
    &  \hspace{-1em}\includegraphics[clip, trim=1.1cm 0cm 0cm 0.25cm, height=4.3cm]{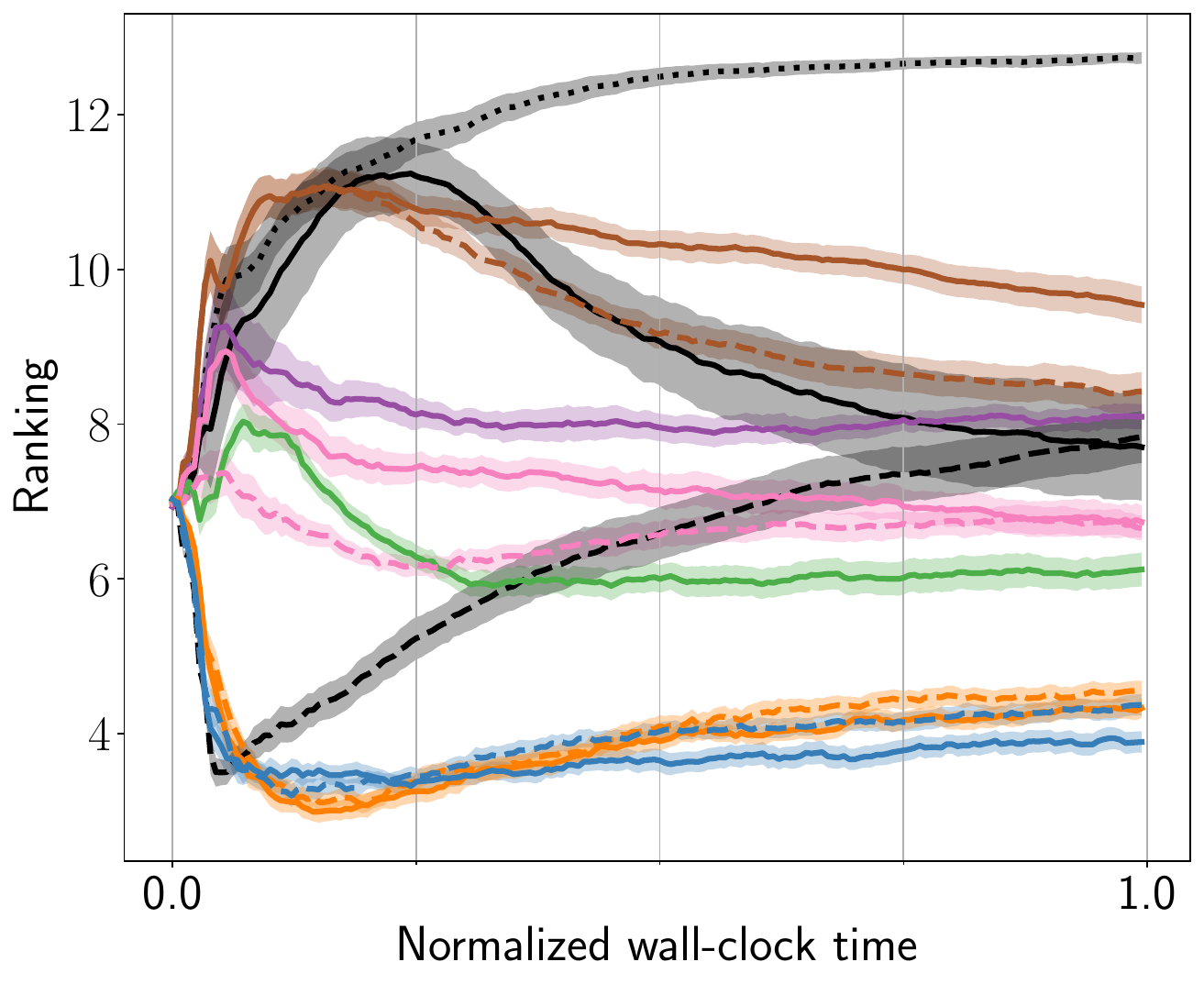}
    &\hspace{-1.4em} \includegraphics[clip, trim=0.3cm -3.5cm 0.0cm 0.0cm, height=4.2cm]{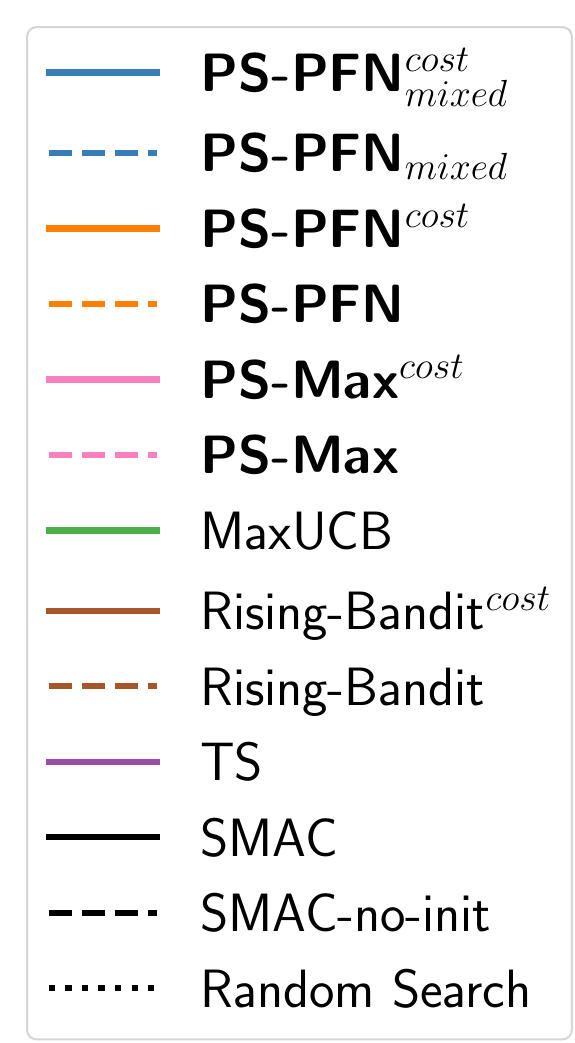}
    \end{tabular}
\caption{Average rank of algorithms on different benchmarks, lower is better. \SMAC{} and \randomsearch{} perform \combinedsearch{} across the joint space.}
\label{fig:average_rank}
\end{figure*}

\section{Experiments on AutoML tasks}
\label{Sec:Numerical_Experiments}
Next, we examine the empirical performance of our method in an AutoML setting by reporting the average ranking for each benchmark. (details in Appendix~\ref{app:benchmarks}). 
We will first briefly overview the experimental setup used across all experiments, and then discuss several research questions focusing on different aspects of this comparison.

\paragraph{Experimental setup.}
We use \numberOfBenchmarks{} AutoML benchmarks, implementing CASH and \CASHplus{} tasks for tabular supervised learning, differing in the considered ML models, optimization method, and datasets (see Table~\ref{app:tab:automltasks}). For the existing CASH benchmarks, \Tabreporaw{} and \Yahpogym{}, we use available pre-computed HPO trajectories~\citep{salinas2024tabrepo,maxucb}. For our newly developed \CASHplus{} task, \Complex{}, we run $5$ different methods across $30$ datasets implementing heterogeneous ML workflows as seen in Figure \ref{fig:heterogeneous_pipeline}.\footnote{\label{footnote:implementation} For all experiments, we used the AutoML Toolkit (AMLTK)~\citep{Bergman2024} and ran the optimization methods on a compute cluster equipped with Intel Xeon Gold 6240 CPUs and NVIDIA 2080 Ti GPUs, requiring $800$ CPU and $50$ GPU days.}

\paragraph{Baselines.} We compare our methods against \MaxUCB{}~\citep{maxucb}, \RisingBandits{} and \RisingBanditsCA{}~\citep{li2020efficient}, which have been developed for the decomposed CASH task. Additionally, we compare against \ThompsonSampling{} as a classical posterior sampling method. We use the default hyperparameter settings for all methods. As \combinedsearch{} baselines, we consider Bayesian optimization (\SMAC{}) and \randomsearch{}.\footnote{
Only available for \Tabreporaw{} and \Yahpogym{}.}

\paragraph{Our Methods.} We evaluate three different methods and their cost-aware variants:
\begin{itemize} 
    \item \textit{\OurAlgoMax{}} as the most basic baseline. We assume rewards follow a Gaussian distribution (see Appendix~\ref{app:priors_choices}).
    \item \textit{\OurAlgoPFN{}} uses the same PFN for modeling the reward distribution of each arm, pre-trained on the semi-flat prior.
    \item \textit{\OurAlgoPFNs{}} uses a different PFN for each arm (see Appendix~\ref{app:priors_choices}).
    \item \textit{\OurAlgoMaxCA{}, \OurAlgoPFNCA{}}, and  \textit{\OurAlgoPFNsCA{}} are the cost-aware extensions of the aforementioned methods. We assume costs follow a log-normal distribution (see Appendix~\ref{app:priors_choices}).
\end{itemize}

\begin{table}[h]
\caption{\OurAlgo{} compared to other \decomposedCASH{} bandit methods developed for AutoML tasks and TS. We report the number of wins, ties, and losses (w/t/l) and the normalized loss across different benchmarks and budgets. Additionally, we report p-values from a sign test to assess whether \OurAlgo{} significantly outperforms other methods and baselines. P-values below $\alpha=0.05$ are underlined, while those that remain below the adjusted $\alpha$ after multiple comparison correction are also boldfaced, indicating statistically significant superiority of \OurAlgo{}.}
\label{tab:sign_test}
\centering
\scriptsize
\begin{tabular}{@{\hskip.0cm}l@{\hskip.1cm}c@{\hskip.2cm}c
@{\hskip.2cm}r@{\hskip.2cm}r@{\hskip.2cm}r@{\hskip.2cm}r@{\hskip.2cm}r@{\hskip.0cm}}
\toprule
{} & Budget & \textbf{PS-PFN}  & vs  & \textbf{PS-Max} & MaxUCB & RisingBandit & TS \\
\midrule 
\parbox[t]{2mm}{\multirow{9}{*}{\rotatebox[origin=c]{90}{\Complex{}}}} 
 & \multicolumn{1}{l}{\multirow{3}{*}{1/3}} &   & p-value  & $\mathbf{\underline{0.00806}}$ & $\mathbf{0.04937}$ & $\mathbf{\underline{0.00003}}$ & $\mathbf{\underline{0.00806}}$ \\
&  &  & w/t/l  & $20$/$4$/$6$ & $18$/$4$/$8$ & $24$/$3$/$3$ & $20$/$3$/$7$ \\
& & $0.3643$& loss & $0.3745$ & $0.3698$ & $0.4358$ & $0.4219$\\
\cmidrule{2-8} 
 & \multicolumn{1}{l}{\multirow{3}{*}{2/3}} &   & p-value  & $\mathbf{\underline{0.00261}}$ & $\mathbf{0.04937}$ & $\mathbf{\underline{0.00000}}$ & $\mathbf{0.02139}$ \\
&  &  & w/t/l  & $21$/$4$/$5$ & $18$/$3$/$9$ & $27$/$0$/$3$ & $20$/$1$/$9$ \\
& & $0.2792$& loss & $0.2898$ & $0.2810$ & $0.3259$ & $0.2880$\\
\cmidrule{2-8} 
 & \multicolumn{1}{l}{\multirow{3}{*}{3/3}} &   & p-value  & $0.57223$ & $0.18080$ & $\mathbf{\underline{0.00072}}$ & $0.10024$ \\
&  &  & w/t/l  & $13$/$3$/$14$ & $16$/$4$/$10$ & $24$/$0$/$6$ & $18$/$2$/$10$ \\
& & $0.3126$& loss & $0.3005$ & $0.2975$ & $0.3131$ & $0.3045$\\
\midrule
\parbox[t]{2mm}{\multirow{9}{*}{\rotatebox[origin=c]{90}{\Tabreporaw{}}}} 
 & \multicolumn{1}{l}{\multirow{3}{*}{1/3}} &   & p-value  & $\mathbf{\underline{0.00000}}$ & $\mathbf{\underline{0.00003}}$ & $\mathbf{\underline{0.00000}}$ & $\mathbf{\underline{0.00000}}$ \\
&  &  & w/t/l  & $27$/$0$/$3$ & $26$/$0$/$4$ & $28$/$0$/$2$ & $29$/$0$/$1$ \\
& & $0.3545$& loss & $0.3627$ & $0.3589$ & $0.4183$ & $0.4241$\\
\cmidrule{2-8} 
 & \multicolumn{1}{l}{\multirow{3}{*}{2/3}} &   & p-value  & $\mathbf{\underline{0.00806}}$ & $0.42777$ & $\mathbf{\underline{0.00000}}$ & $\mathbf{\underline{0.00000}}$ \\
&  &  & w/t/l  & $22$/$0$/$8$ & $16$/$0$/$14$ & $28$/$0$/$2$ & $29$/$0$/$1$ \\
& & $0.2884$& loss & $0.3003$ & $0.2839$ & $0.3583$ & $0.3277$\\
\cmidrule{2-8} 
 & \multicolumn{1}{l}{\multirow{3}{*}{3/3}} &   & p-value  & $0.18080$ & $0.95063$ & $\mathbf{\underline{0.00016}}$ & $\mathbf{\underline{0.00000}}$ \\
&  &  & w/t/l  & $18$/$0$/$12$ & $11$/$0$/$19$ & $25$/$0$/$5$ & $27$/$0$/$3$ \\
& & $0.3255$& loss & $0.2804$ & $0.2727$ & $0.3335$ & $0.3141$\\
\midrule
\parbox[t]{2mm}{\multirow{9}{*}{\rotatebox[origin=c]{90}{\Yahpogym{}}}}
 & \multicolumn{1}{l}{\multirow{3}{*}{1/3}} &   & p-value  & $\mathbf{\underline{0.00000}}$ & $\mathbf{\underline{0.00000}}$ & $\mathbf{\underline{0.00000}}$ & $\mathbf{\underline{0.00000}}$ \\
&  &  & w/t/l  & $87$/$1$/$12$ & $75$/$1$/$24$ & $93$/$1$/$6$ & $86$/$1$/$13$ \\
& & $0.1494$& loss & $0.1937$ & $0.1723$ & $0.2384$ & $0.1822$\\
\cmidrule{2-8} 
 & \multicolumn{1}{l}{\multirow{3}{*}{2/3}} &   & p-value  & $\mathbf{\underline{0.00000}}$ & $\mathbf{\underline{0.00000}}$ & $\mathbf{\underline{0.00000}}$ & $\mathbf{\underline{0.00000}}$ \\
&  &  & w/t/l  & $78$/$1$/$21$ & $75$/$1$/$24$ & $82$/$0$/$18$ & $79$/$1$/$20$ \\
& & $0.1376$& loss & $0.1451$ & $0.1417$ & $0.1766$ & $0.1520$\\
\cmidrule{2-8} 
 & \multicolumn{1}{l}{\multirow{3}{*}{3/3}} &   & p-value  & $\mathbf{\underline{0.00000}}$ & $\mathbf{\underline{0.00009}}$ & $\mathbf{\underline{0.00000}}$ & $\mathbf{\underline{0.00000}}$ \\
&  &  & w/t/l  & $74$/$1$/$25$ & $68$/$1$/$31$ & $79$/$0$/$21$ & $78$/$1$/$21$ \\
& & $0.1260$& loss & $0.1339$ & $0.1195$ & $0.1390$ & $0.1410$\\
\bottomrule
\end{tabular}
\end{table}
\paragraph{How does \OurAlgoMax{} and \OurAlgoPFN{} compare against \ThompsonSampling{}?}
In Figure~\ref{fig:average_rank}, we compare the average rank over time of \OurAlgoMax{} (\pinkdashed{}) and \OurAlgoPFN{} (\orangedashed{}) to classical \ThompsonSampling{} (\purplesolid{}). While \OurAlgoMax{} and \ThompsonSampling{} initially perform similarly, this gap increases with higher budget, highlighting that targeting maximum values is crucial for effectively solving CASH problems.

\paragraph{How does using a PFN improve the performance?} Figure~\ref{fig:average_rank} shows a large improvement in ranking when comparing \OurAlgoMax{} (\pinkdashed{}) with \OurAlgoPFN{} (\orangedashed{}). Table~\ref{tab:sign_test} also shows that this difference is statistically significant. Furthermore, 
using different PFNs per arm, each trained on a different prior, can further improve ranking, especially for high budgets, as seen by \OurAlgoPFNs{} (\bluedashed) performing better than \OurAlgoPFN{} (\orangedashed{}). However, this gap is smaller for \Yahpogym{} and \Complex{}, underlining the importance of carefully choosing a prior. 

\paragraph{How does \OurAlgo{} compare against other \decomposedCASH{} bandit methods?} In Figure~\ref{fig:average_rank}, we observe that \RisingBandits{} (\brownsolid{}) perform substantially worse than our method (which the statistical results in Table~\ref{tab:sign_test} emphasize). While the performance of \MaxUCB{} (\greensolid{}) can be competitive, it tends to perform worse for low budgets. Since the non-PFN-based variant~\OurAlgoMax{} (\pinksolid{}) consistently performs worse than \MaxUCB{} (\greensolid{}), we attribute the superior performance not to the bandit framework, but to leveraging an ML model.

\paragraph{How does cost-awareness improve the performance?}
Finally, we evaluate our cost-aware extension.
For this, we compare the cost-aware variants (solid) to their counterpart (dashed) for \OurAlgoMax{} (\pinksolid{} vs \pinkdashed), \OurAlgoPFN{} (\orangesolid{} vs \orangedashed{}), and \OurAlgoPFNs{} (\bluesolid{} vs \bluedashed) in Figure~\ref{fig:average_rank}. The improvement is overall less pronounced, with cost-awareness sometimes even performing worse (e.g., \Tabrepo{}).
Table~\ref{tab:percentage_improvement} demonstrates that the number of arms pulled consistently increases when cost-awareness is enabled, indicating that the agent observes more rewards within the same amount of total budget. However, cost-awareness does not necessarily improve final performance if the less costly arms perform worse.

\begin{table}[h]
\centering
\caption{Percentage gain in the number of pulling arms for using the cost-aware variant of each method.}
\label{tab:percentage_improvement}
\begin{tabular}{lrrr}
\toprule
{} &  \OurAlgoPFNsCA{} &  \OurAlgoPFNCA{} &  \OurAlgoMaxCA{} \\
\midrule
\Yahpogym{}  &      7.50\% &   4.66\% &   2.94\% \\
\Tabreporaw{} &     15.36\% &   6.74\% &   7.55\% \\
\Complex{}    &     15.32\% &  17.35\% &   4.69\% \\
\bottomrule
\end{tabular}
\end{table}

\section{Discussion and Future Work} 
\label{Sec:Conclusion}

By extending the CASH framework to \CASHplus{}, we address the selection and adaptation of heterogeneous pipelines, covering modern ML workflows using fine-tuning, ensembling, and hyperparameter optimization. We address the resulting bi-level optimization problem as a max $k$-armed bandit problem and identify posterior sampling as a flexible framework for resource allocation. To efficiently model reward distributions that do not follow common forms, are different for each arm, and potentially shift over time, we exploit PFNs~\citep{muller2023pfns4bo}. The resulting method, \OurAlgoPFN{} and its extensions, outperforms prior approaches and paves the way for a data-driven development of AutoML solutions for \CASHplus{}. In the following, we briefly discuss limitations and avenues for future research.

\paragraph{Limitations.} \OurAlgoPFN{} is computationally more expensive than other baselines since it leverages in-context learning for estimating the posterior distribution. Thus, in cases where iterations of the lower-level (optimization) method are fast, \OurAlgoPFN{} may dominate compute costs. Additionally, quadratic scaling in context length limits the application for large budgets since we use reward observations as context. However, in \CASHplus{} tasks, iterations of the underlying method are typically cost-intensive, with tight budget constraints.
We note that theoretically analyzing \OurAlgoPFN{} might be challenging due to the heuristic synthetic data generation and using ML models to approximate the posterior.
Furthermore, while our cost-aware extension increases the number of arms pulled, performance does not constantly improve. We model the training cost with a log-normal distribution; however, it could be modeled more accurately based on the dataset and configurations. Lastly, though our method can be used beyond AutoML tasks, it is sensitive to the prior choice for training the PFNs, and applying it to other tasks might require careful adaptation.

\paragraph{Future directions.} Flexibility is the main advantage of \OurAlgoPFN{}. By defining different priors to train PFNs and, thus, also to model reward distributions, we can incorporate meta-knowledge derived by data analysis or from domain experts.
The optimal choice of priors improves performance; however, this selection can be challenging for new and unknown tasks. Instead of robust defaults, future work could investigate ways to automatically derive a prior in a data-driven way or fine-tune the PFNs on new observations. Additionally, our newly introduced benchmark provides the first step in covering heterogeneous pipelines. We expect with the advancement of pre-trained models for predictive tabular ML to encounter a larger and much more diverse choice of pipelines covering other iterative adaptation methods, such as context optimization~\citep{feuer-neurips24tunetables}, data augmentation~\citep{cui-arxiv24}, feature engineering using LLMs~\citep{hollmann-arxiv23a-v5}, and many more.




\begin{ack}
The authors are funded by the Deutsche Forschungsgemeinschaft (DFG, German Research Foundation) under Germany’s Excellence Strategy -– EXC number 2064/1 -– Project number 390727645. The authors also thank the International Max Planck Research School for Intelligent Systems (IMPRS-IS).
\end{ack}


\bibliography{strings, lib, references, shortproc,myshortproc}

\counterwithin{figure}{section}
\counterwithin{table}{section}
\counterwithin{algorithm}{section}

\clearpage            
\onecolumn            
\appendix            

\vspace*{1cm}
\section*{\centering Table of Contents for the Appendices}
\vspace{1cm}
\begin{itemize}[itemsep=0.8em, topsep=1em]
\item \textbf{Appendix A: Preliminaries} \dotfill \pageref{app:preliminiaries}
\begin{itemize}
  \item A.1 Proof of Theorem \dotfill \pageref{app:proof}
  \item A.2 Budget Constraint and Variable Costs \dotfill \pageref{app:cost_awareness}
  \item A.3 Importance of priors \dotfill \pageref{app:priors_importance}
  \item A.4 Out of distribution \dotfill \pageref{app:out_of_distribution}
  \item A.5 Trajectory Analysis \dotfill \pageref{app:trajectory_analysis}
\end{itemize}
\item \textbf{Appendix B: Benchmarks} \dotfill \pageref{app:benchmarks}
\item \textbf{Appendix C: PFN architecture} \dotfill \pageref{app:priors_architecture}
\item \textbf{Appendix D: Priors} \dotfill \pageref{app:priors_choices}
\item \textbf{Appendix E: Results in details} \dotfill \pageref{app:more_results}
\end{itemize}

\newpage
\section{Preliminiaries}
\label{app:preliminiaries}

\subsection{Proof of Theorem}
\label{app:proof}
\textbf{Theorem~\ref{theorem:max_sampling}.} (Analysis of \OurAlgoMax)
     If the expected maximum satisfies:
    \[
    \mathbb{E}[ \max (r_{i,1:t})] \geq F_i^{-1}\left(1 - \frac{1}{t^2}\right),
    \]
    Then the number of times sub-optimal arm $i$ is pulled grows logarithmically with the time horizon.
\begin{proof}
\label{app:preliminaries:theorem_1_proof}
\textbf{Sampling the Maximum.} Let $r_1, \dots, r_t$ be i.i.d. with CDF $F(x)$. The CDF of the maximum of reward $\max (r_{1:t})$ is:
\begin{equation}
    F_{\max}(x) = P(\max (r_{1:t}) \leq x) =\prod_{i=1}^{t} P(r_i \leq x) = [F(x)]^{t}.
    \label{app:preliminaries:lemma_eq1}
\end{equation}
To sample from $F_{\max}$, we use inverse transform sampling:
\begin{enumerate}
    \item Draw $U \sim \mathrm{Uniform}(0,1)$
    \item Solve $F_{\max}(x) = U$:
    \[
    [F(x)]^t = U \implies F(x) = U^{1/t} \implies x = F^{-1}(U^{1/t})
    \]
\end{enumerate}
Thus, $\max (r_{1:t}) = F^{-1}(U^{1/t})$ follows the correct distribution.

\textbf{Convergence Analysis.} For the convergence analysis, let \( F_n(x) \) denote the empirical CDF estimated from \( n \) i.i.d.\ samples. Using the Dvoretzky-Kiefer-Wolfowitz (DKW) inequality:
\begin{equation}
\mathbb{P}\left(\sup_x |F_n(x) - F(x)| \geq \epsilon \right) \leq \delta, \quad \text{with } \epsilon = \sqrt{\frac{\ln(2/\delta)}{2n}}.
\label{eq:dkw_bound}
\end{equation}

Now we want to bound $|F^t_n(x) - F^t(x)|$. We know:
\[
|F^t_n(x) - F^t(x)| \leq t |F_n(x) - F(x)|
\]

Hence,
\begin{equation}
\mathbb{P}\left(\sup_x |F^t_n(x) - F^t(x)| \geq t \epsilon \right) \leq \delta.
\label{eq:dkw_power_bound}
\end{equation}

This implies a worst-case deviation that grows linearly with \( t \), leading to potentially linear regret. However, we are particularly interested in the concentration around the maximum $\max (r_{1:t})$, where tighter bounds can be established under some conditions.

We know the variance of the empirical CDF satisfies:
\[
\mathrm{Var}(F_n(x)) = \frac{F(x)(1 - F(x))}{n} \leq \frac{1 - F(x)}{n}.
\]
Applying Bernstein's inequality, for any \( \lambda > 0 \) and  near the right tail $x = \max (r_{1:t})$:
\[
\mathbb{P}\left( |F_n(\max (r_{1:t})) - F(\max (r_{1:t}))| \geq \lambda \right) \leq 2 \exp\left( -\frac{n \lambda^2}{2(1 - F(\max (r_{1:t}))) + \frac{2}{3} \lambda} \right).
\]
With rewriting it as the form of \( \epsilon \) at confidence level \( \delta \), we obtain:
\begin{equation}
\mathbb{P}\left( |F_n(\max (r_{1:t})) - F(\max (r_{1:t}))| \geq \epsilon) \right) \leq \delta, \quad \text{where } \epsilon = \sqrt{\frac{2(1 - F(\max (r_{1:t})) \ln(2/\delta)}{n}}.
\label{eq:bernstein_tail}
\end{equation}
This shows that the deviation is significantly smaller near the maximum when
\[
1 - F(\max (r_{1:t})) \leq \frac{1}{t^2}.
\]

By applying \Eqref{eq:bernstein_tail} to \Eqref{eq:dkw_power_bound}, we obtain the following refined bound under the additional condition:
\begin{align}
    \mathbb{P}\left( |F^t_n(\max (r_{1:t})) - F(\max (r_{1:t}))| \geq \epsilon) \right) \leq \delta, \quad \text{where } \epsilon = \sqrt{\frac{\ln(2/\delta)}{n}}, \quad \ \text{and } \mathbb{E}[\max (r_{1:t})] \geq F^{-1}\left(1 - \frac{1}{t^2}\right)
\label{eq:new_dkw_power_bound}
\end{align}

under which the concentrations are fast enough. With this fast concentration, following the classical Thompson sampling proof (see Theorem 36.2 from \citep{lattimore2020bandit}), we can show that the number of times sub-optimal arms are pulled grows logarithmically with the time horizon. 
\end{proof}

\newpage

\subsection{Budget Constraint and Variable Costs}
\label{app:cost_awareness}
In addition to the main paper, we provide more details on implementing cost-awareness in our method. For this, we consider \CASHplus{} with a time budget constraint $B$ and variable costs per arm, i.e. the evaluation of the model for the model state $A^{(i)}_{\pmb{s}}$ for dataset $\sD$ costs $\mathcal{C}(A^{(i)}_{\pmb{s}},\sD)$, with $\pmb{L_S}^{(i)}$ being the list of previously evaluated configurations for model $i$. Formally,

\begin{align}
&A^{(i^*)} \in \argmin_{A^{(i)} \in \mathcal{A} } \mathcal{L}(A^{(i)}_{\pmb{s}^*} ,\sD) 
\label{eq:upper_level_optimzation_b} \\
&\text{s.t.  } \pmb{S}^{*} \in \argmin_{ \pmb{s} \in \pmb{S^{(i)}} } \mathcal{L}( A^{(i)}_{\pmb{s}}  ,\sD).
\label{eq:lower_level_optimzation_b}\\
&\text{s.t.  } \sum_{i \in K} \sum_{ \pmb{s} \in \pmb{L_S^{(i)}} } \mathcal{C}(A^{(i)}_{\pmb{s}} ,\sD) \leq B
\end{align}

Given an overall budget of $B$, we define $\pmb{s}_t$ to be the model state proposed by the optimizer in the lower level and $r_{i,t}$ to be the feedback to arm $i$ obtained by evaluating $\pmb{s}_t$. To be consistent with the bandit literature, we maximize the negative loss with cost $c_{i,t}$: 
\begin{align}
&r_{i,t} =-\mathcal{L}( A^{i}_{\pmb{s}_t}  ,\sD) \notag \\
&c_{i,t} = \mathcal{C}(A^{(i)}_{\pmb{s}_t } ,\sD)
\end{align}
The goal is then to find the best-performing algorithm $A^*$ and its state $\pmb{s}^*$ with spending a time budget $B$, and we assume the cost of pulling arm $i$ follows a sub-Gaussian distribution with expected mean $c_i$.

\begin{align}
R(B) &= \max_{i \leq K} \mathbb{E}\left[ \max_{t \leq \lfloor B/c_i \rfloor} r_{i,1:t} \right] 
       - \mathbb{E}\left[ \max_{t \leq T} r_{I_t,1:t} \right] \\
\text{s.t.} \quad & \sum_{t=1}^T c_{I_t,t} \leq B
\end{align}

Each arm experiences a different time horizon $T_i = \frac{B}{c_i}$. If we sort arms based on optimality (denoting the optimal arm as $1$ and the most sub-optimal arm as $K$), we have:

\begin{align}
\mathbb{E}\left[ \max_{t \leq \lfloor B/c_1 \rfloor} r_{1,1:t} \right] > 
\mathbb{E}\left[ \max_{t \leq \lfloor B/c_2 \rfloor} r_{2,1:t} \right] > 
\cdots > 
\mathbb{E}\left[ \max_{t \leq \lfloor B/c_K \rfloor} r_{K,1:t} \right]
\end{align}

Let at time step $t$, arm $i$ be pulled for $n_i$ times and the spent budget be $b = \sum_{i \leq k}\sum_{t = 1}^{n_i} c_{i,1:t}$. The remaining budget at time step $t$ is $B-b$. And arm $i$ has been pulled for $n_i$ times. In the base case, we can only pull this arm for $\frac{B-b}{c_i}$ times. Meaning that instead of $f(t)=t$ we need to predict the posterior at $f(t)= n_i + \frac{B-b}{c_i}$. However, the cost of pulling arms is noisy (e.g., the cost of pulling an arm implementing hyperparameter optimization will depend on the model size and training hyperparameters for each configuration), and we need to estimate $c_i$. We model these costs as a random distribution to estimate $c_i$, and sample the cost from this distribution.

\begin{equation}
    f_i(t) =  n_i + \frac{B - b} { c \sim p_c(\cdot \mid \{c_{i,t}\}_{t=1}^{n_i} )}
    \quad\text{where } \quad b=\sum_{i \leq k}\sum_{t = 1}^{n_i} c_{i,t}.
\end{equation}

Principally, without having prior information about maximum budget $B$, one can use $f(t)$ as follows:

\begin{equation}
    f_i(t) =  \frac{b} { c \sim  p_c(\cdot \mid \{c_{i,t}\}_{t=1}^{n_i} )}
    \quad\text{where } \quad b=\sum_{i \leq k}\sum_{t = 1}^{n_i} c_{i,t}.
\end{equation}

This is the cost-scaled version of $f(t)=t$.

\subsection{Importance of priors}
\label{app:priors_importance}
Here, we study the sensitivity of the PFN with respect to its prior. For this, we pre-trained three separate PFNs using rewards generated from a truncated skewed normal distribution. The priors mainly differ in the parameter range used to sample from to initiate the distribution
, as shown in Table~\ref{app:tab:priors_importance}. We then evaluated them as a model for \OurAlgoPFN{} on a synthetic multi-armed bandit task with $7$ arms, with the reward distributions characterized by the mean  $\mu$ and standard deviation $\sigma$ values listed in Table~\ref{app:tab:priors_importance_synthetic_dataset}. To assess generalization across skewness, we varied the skewness parameter $a$ within the range $[-100,100]$ in increments of $5$.

\begin{table}[h]
    \centering
    \begin{minipage}[t]{0.5\textwidth}
        \centering
        \begin{tabular}{l|l}
            \textbf{Dataset} & \textbf{Parameters} \\
            \hline
            \multirow{3}{*}{Neg} 
            & \scriptsize $a \sim \mathcal{U}(-100, -20)$ \\
            & \scriptsize $\mu \sim \mathcal{U}(0.0, 1.0)$ \\
            & \scriptsize $\sigma \sim \mathcal{U}(0, 0.2)$ \\
            \hline
            \multirow{3}{*}{Pos} 
            & \scriptsize $a \sim \mathcal{U}(20, 100)$ \\
            & \scriptsize $\mu \sim \mathcal{U}(0.0, 1.0)$ \\
            & \scriptsize $\sigma \sim \mathcal{U}(0, 0.2)$ \\
            \hline
            \multirow{3}{*}{Mix} 
            & \scriptsize $a \sim \mathcal{U}(-100, 100)$ \\
            & \scriptsize $\mu \sim \mathcal{U}(0.0, 1.0)$ \\
            & \scriptsize $\sigma \sim \mathcal{U}(0, 0.2)$ \\
        \end{tabular}
        \caption{Parameters of the priors per dataset type used to train the PFNs.}
        \label{app:tab:priors_importance}
    \end{minipage}
     \hfill   
    \begin{minipage}[t]{0.45\textwidth}
        \centering
        \begin{tabular}{l|c|c}
            \textbf{Arm} & \textbf{Parameter $\mu$} & \textbf{Parameter $\sigma$} \\
            \hline
            1 & 0.80 & 0.05 \\
            2 & 0.75 & 0.05 \\
            3 & 0.70 & 0.05 \\
            4 & 0.60 & 0.05 \\
            5 & 0.70 & 0.10 \\
            6 & 0.60 & 0.10 \\
            7 & 0.50 & 0.10 \\
        \end{tabular}
        \caption{Reward distribution parameters for the synthetic MAB tasks.}
        \label{app:tab:priors_importance_synthetic_dataset}
    \end{minipage}
\end{table}

We analyze the impact of prior distributions by evaluating both ranking and normalized regret plots. We define three types of environments based on the skewness parameter $a$: \textbf{Neg} for strongly left-skewed distributions ($a \in [-100,-45]$), \textbf{Mix} for mildly skewed or symmetric distributions ($a \in [-45,45]$), and \textbf{Pos} for right-skewed distributions ($a \in [45,100]$).

\begin{figure*}[htb]
    \begin{tabular}{c c c c c}
    \hspace{-0.5em}\small Neg
    & \hspace{-1em}\small Mix
    &\hspace{-1.5em}\small Pos
    & \\
    \includegraphics[clip, trim=0.2cm 0cm 0.3cm 0.25cm,height=4.3cm]{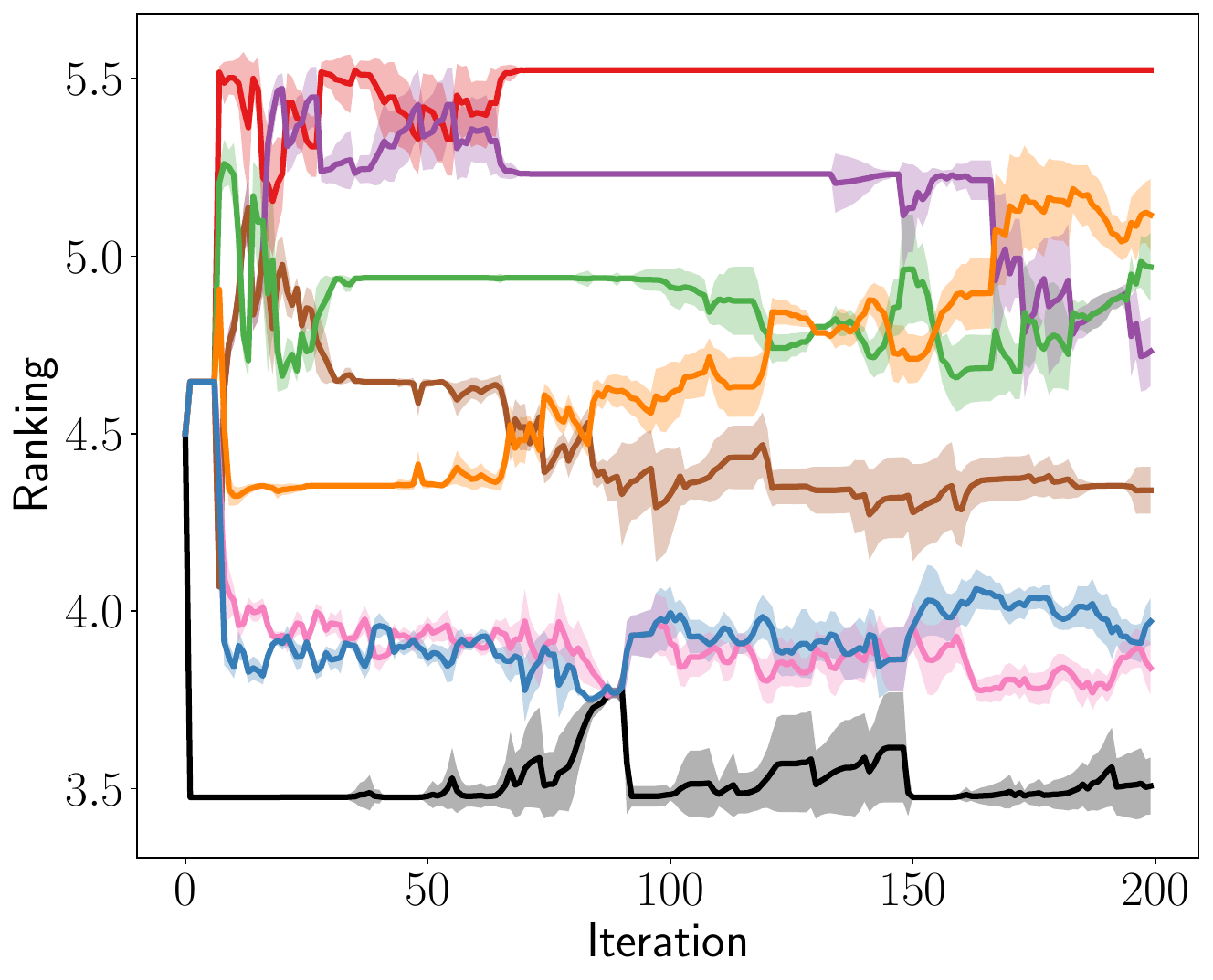}
    & \hspace{-1em} \includegraphics[clip, trim=1.0cm 0cm 0cm 0.25cm, height=4.3cm]{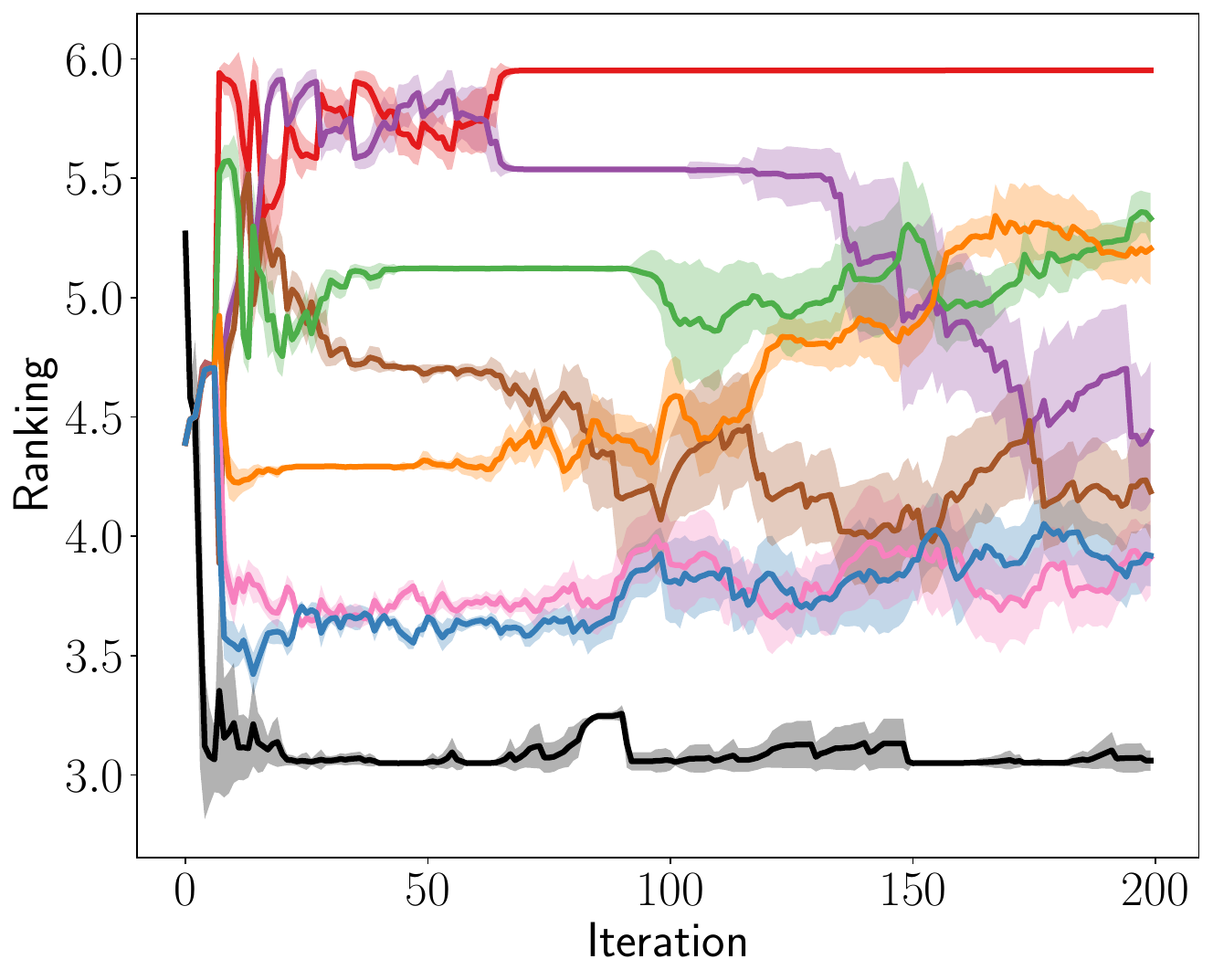}
    &  \hspace{-1em}\includegraphics[clip, trim=1.0cm 0cm 0cm 0.25cm, height=4.3cm]{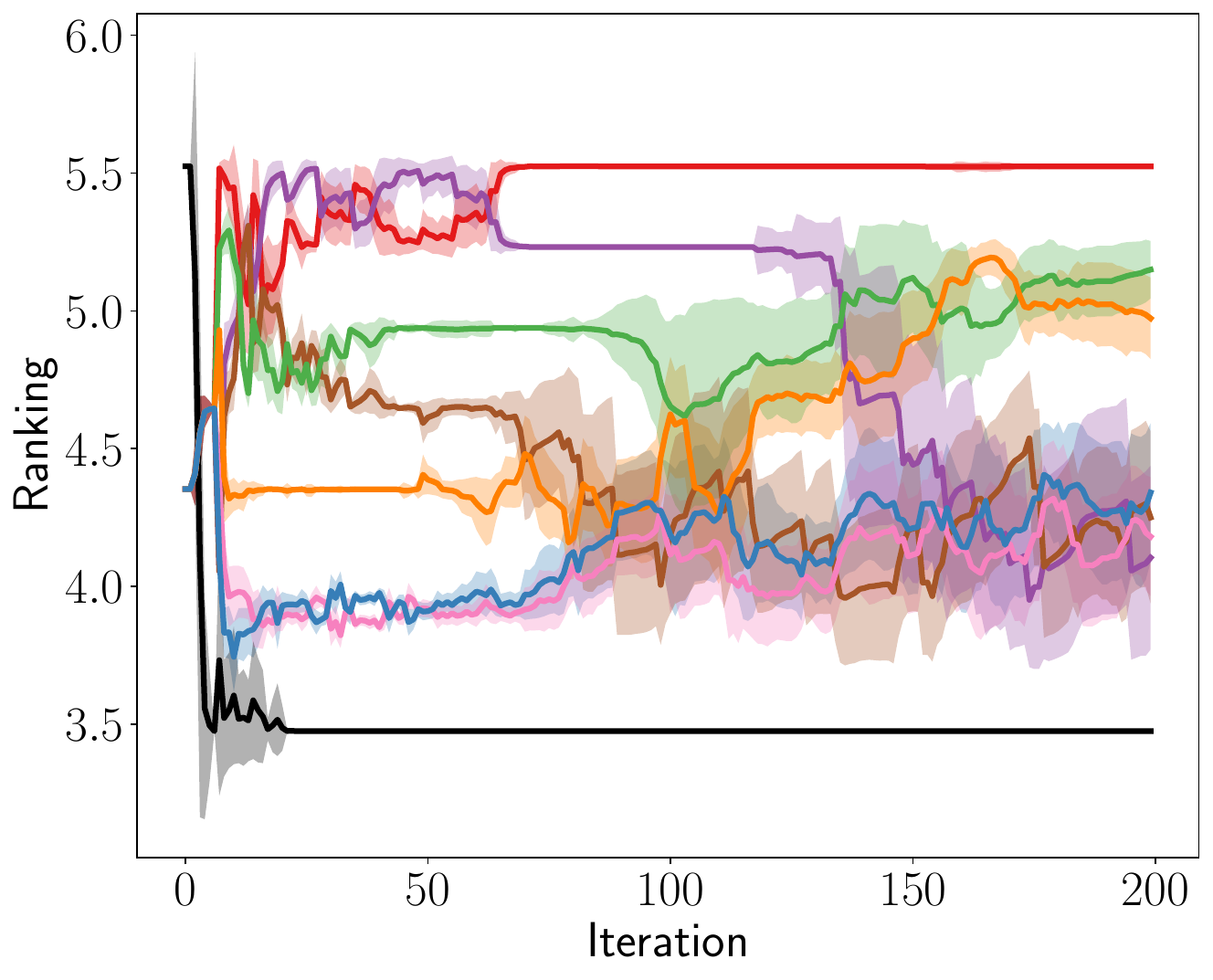}
    &\hspace{-1.4em} \includegraphics[clip, trim=0.3cm -3.5cm 0.0cm 0.0cm, height=4.2cm]{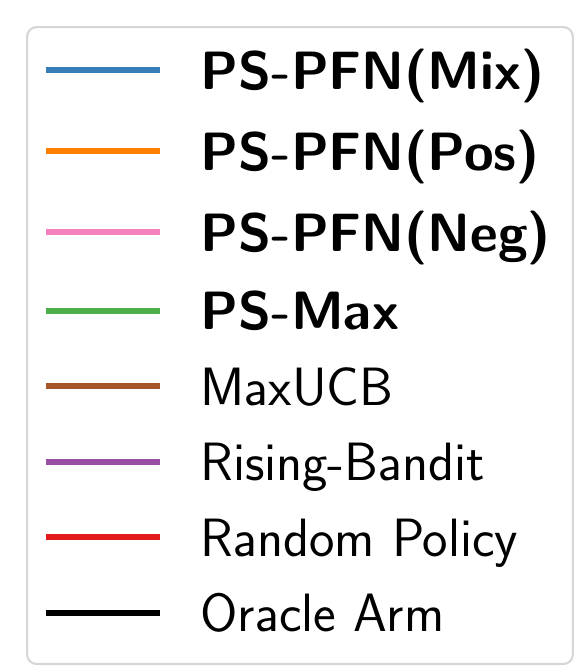}
    \end{tabular}
\caption{Comparing ranking of \textit{PS-PFN} with different priors on synthetic tasks.}
\label{app:priors_importance_rank}
\end{figure*}

\begin{figure*}[htb]
    \begin{tabular}{c c c c c}
    \hspace{-0.5em}\small Neg
    & \hspace{-1em}\small Mix
    &\hspace{-1.5em}\small Pos
    & \\
    \includegraphics[clip, trim=0.2cm 0cm 0.3cm 0.25cm,height=4.3cm]{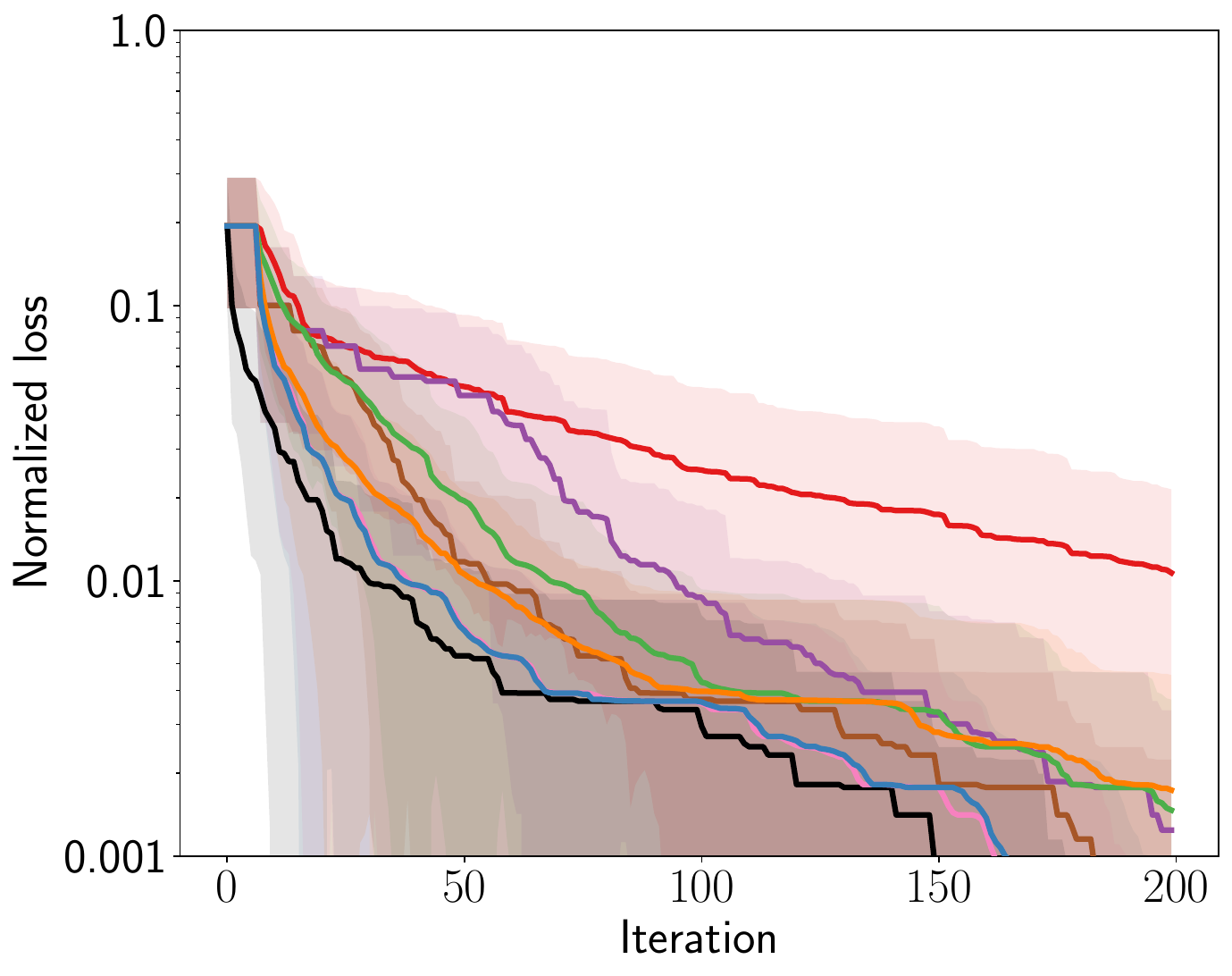}
    & \hspace{-1em} \includegraphics[clip, trim=1.0cm 0cm 0cm 0.25cm, height=4.3cm]{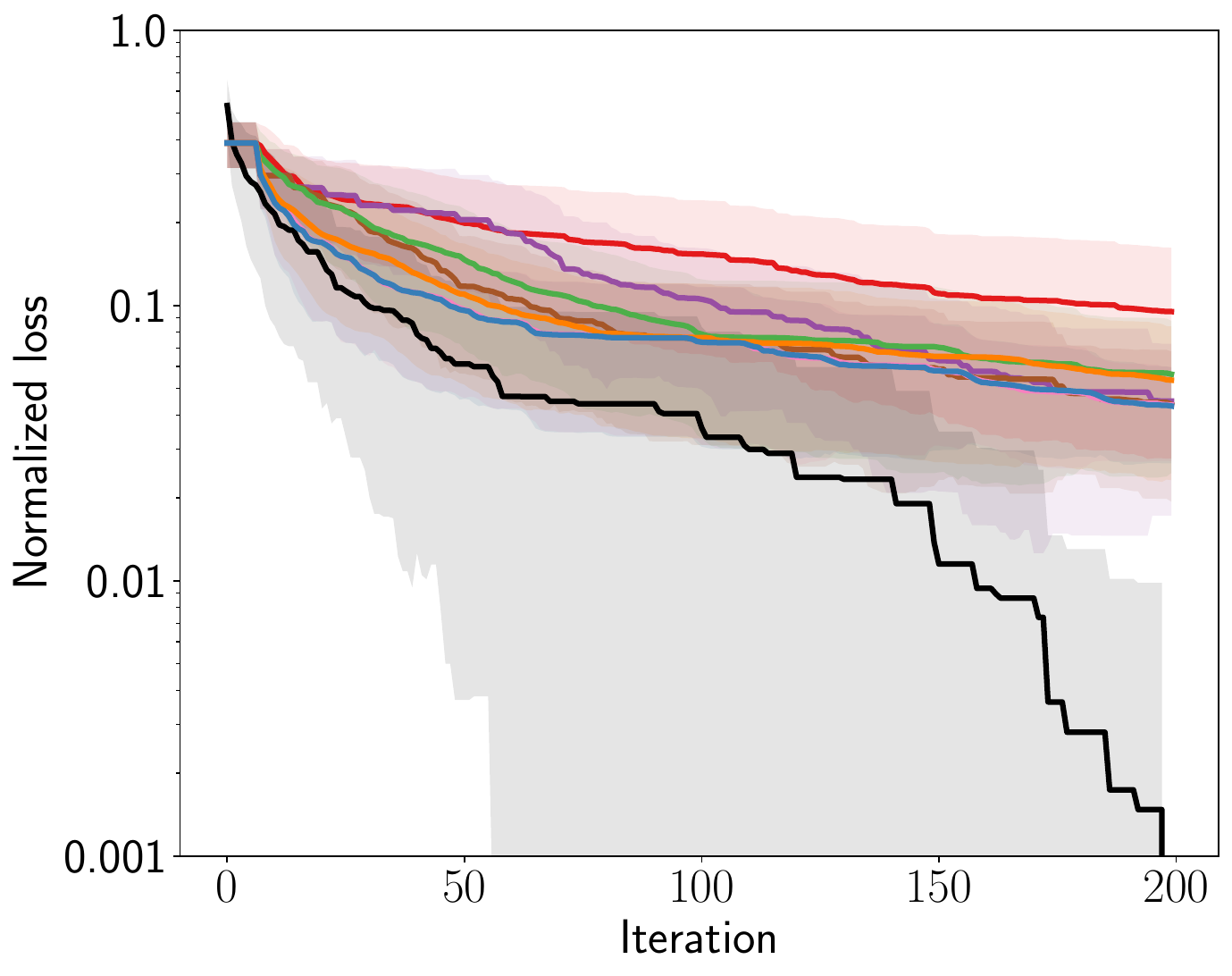}
    &  \hspace{-1em}\includegraphics[clip, trim=1.0cm 0cm 0cm 0.25cm, height=4.3cm]{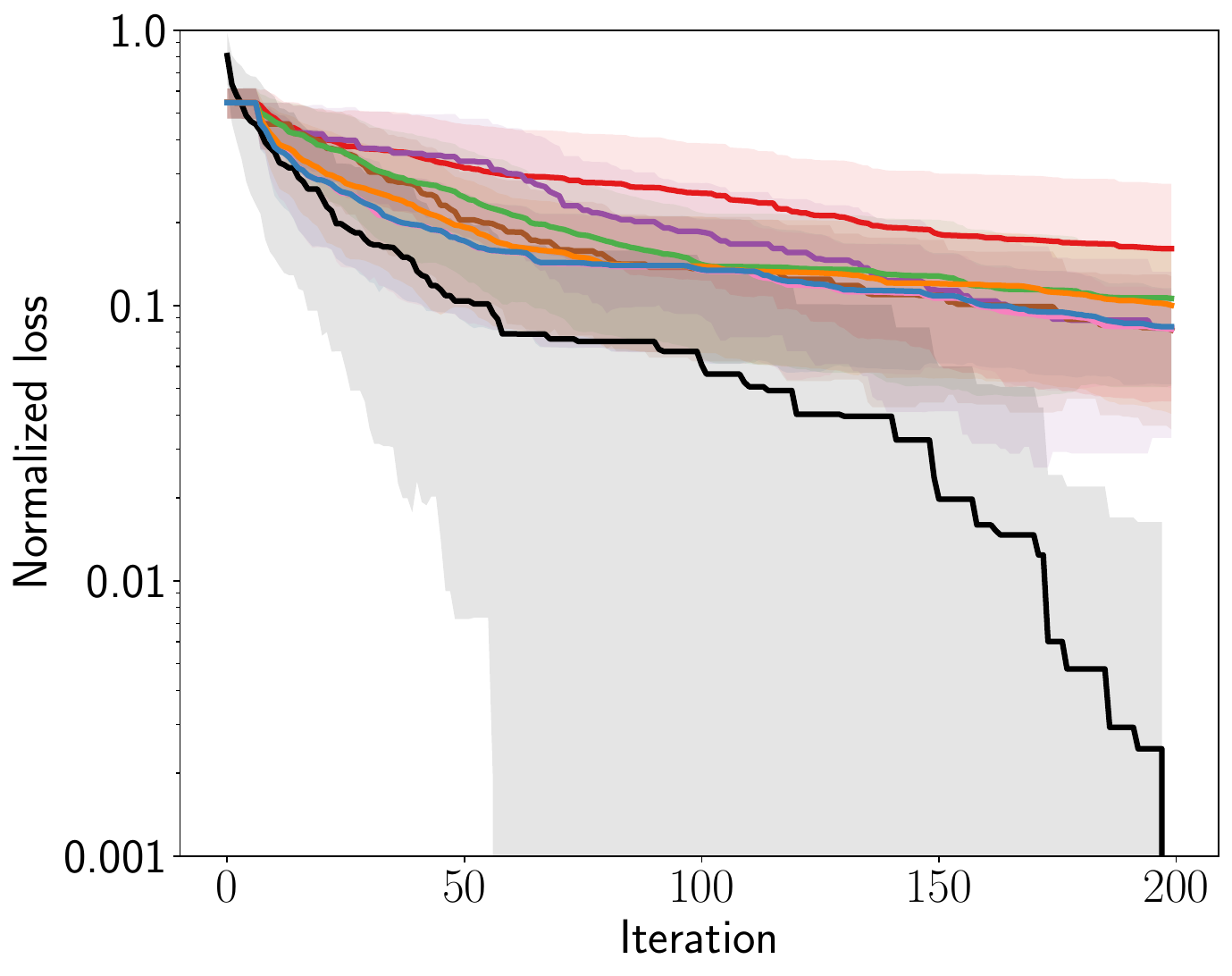}
    &\hspace{-1.4em} \includegraphics[clip, trim=0.3cm -3.5cm 0.0cm 0.0cm, height=4.2cm]{priors_importance_synth_ablation_legend.pdf}
    \end{tabular}
\caption{Comparing normalized loss of \textit{PS-PFN} with different priors on synthetic tasks.}
\label{app:priors_importance_regret}
\end{figure*}

As shown in the ranking plot (Figure~\ref{app:priors_importance_rank}), an interesting observation is that \textit{PS-PFN(Pos)} underperforms even in positively skewed environments. This is likely because right-skewed reward distributions make the problem more exploration-heavy, requiring more time horizon than the available budget allows. This trend is also reflected in the normalized regret plot (Figure~\ref{app:priors_importance_regret}), where \textit{PS-PFN(Pos)} exhibits higher regret.

Furthermore, the number of arm pulls reported in Figure~\ref{app:priors_importance_regret} shows that \textit{PS-PFN(Pos)} explores more frequently than the other models. Finally, the heatmap in Figure~\ref{app:priors_importance_heatmap} highlights that \textit{PS-PFN(Neg)} performs best in environments with left-skewed reward distributions.

\begin{figure*}[htb]
    \begin{tabular}{c c c c c}
    \hspace{-0.5em}\small Neg
    & \hspace{-1em}\small Mix
    &\hspace{-1.5em}\small Pos
    & \\
    \includegraphics[clip, trim=0.2cm 0cm 0.3cm 0.25cm,height=3.5cm]{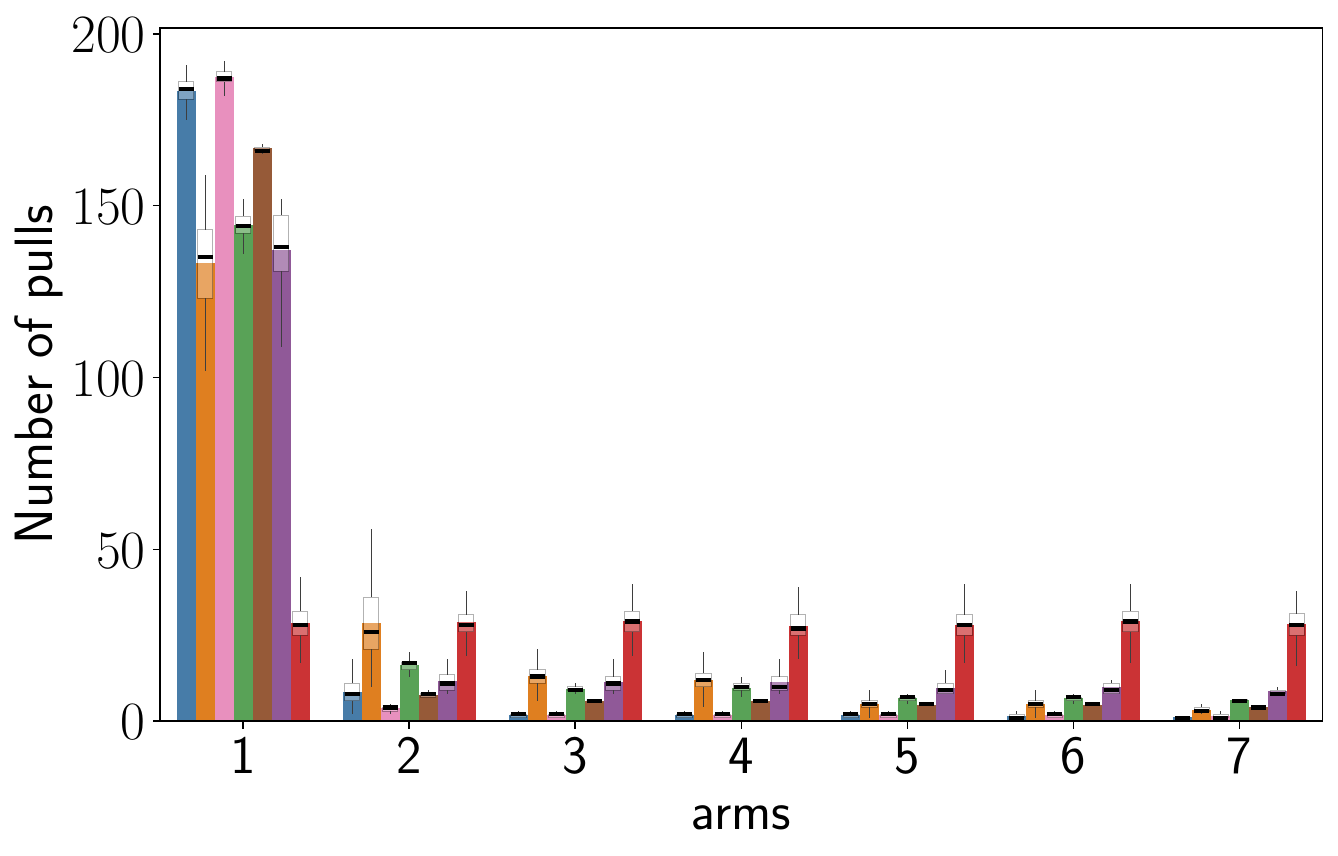}
    & \hspace{-1em} \includegraphics[clip, trim=1.0cm 0cm 0cm 0.25cm, height=3.5cm]{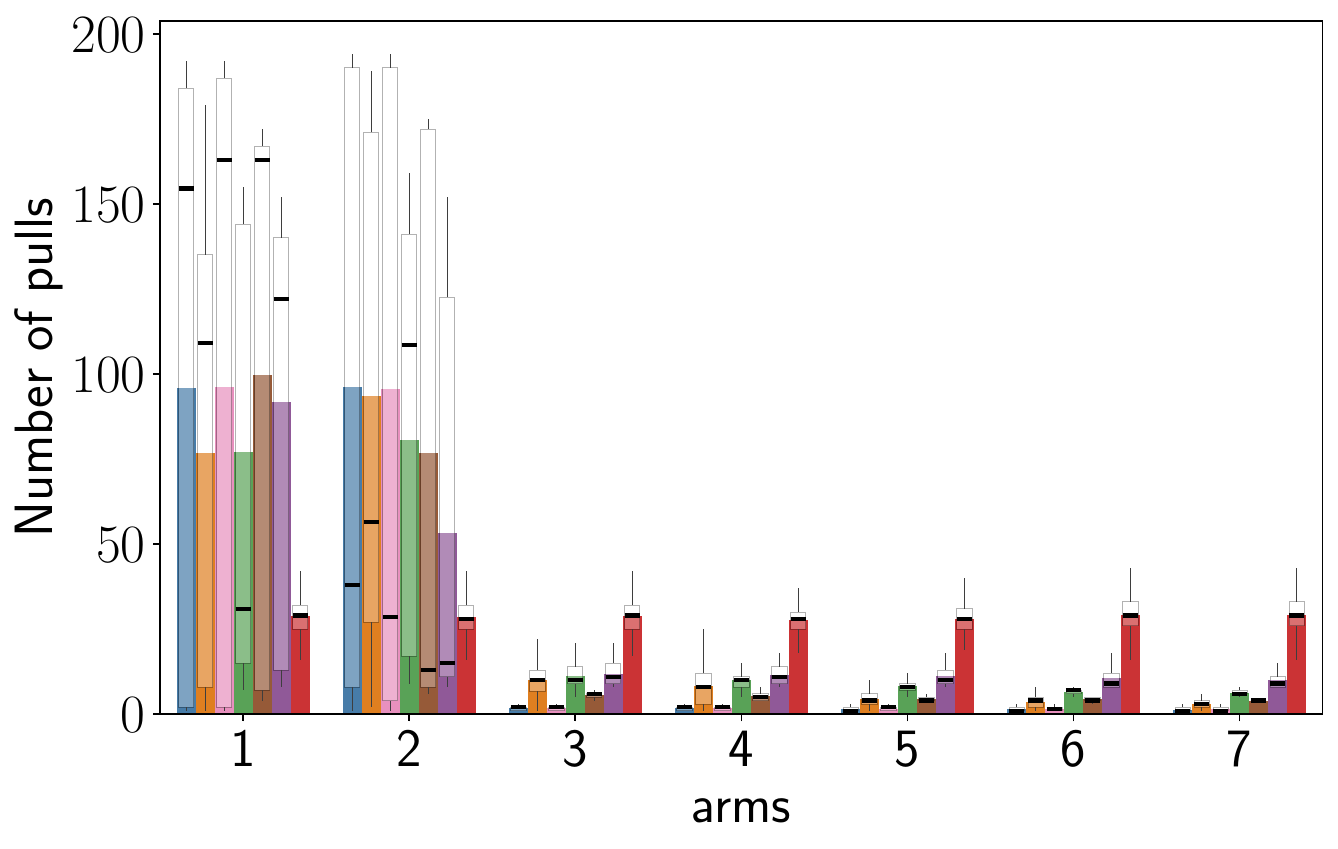}
    &  \hspace{-1em}\includegraphics[clip, trim=1.0cm 0cm 0cm 0.25cm, height=3.5cm]{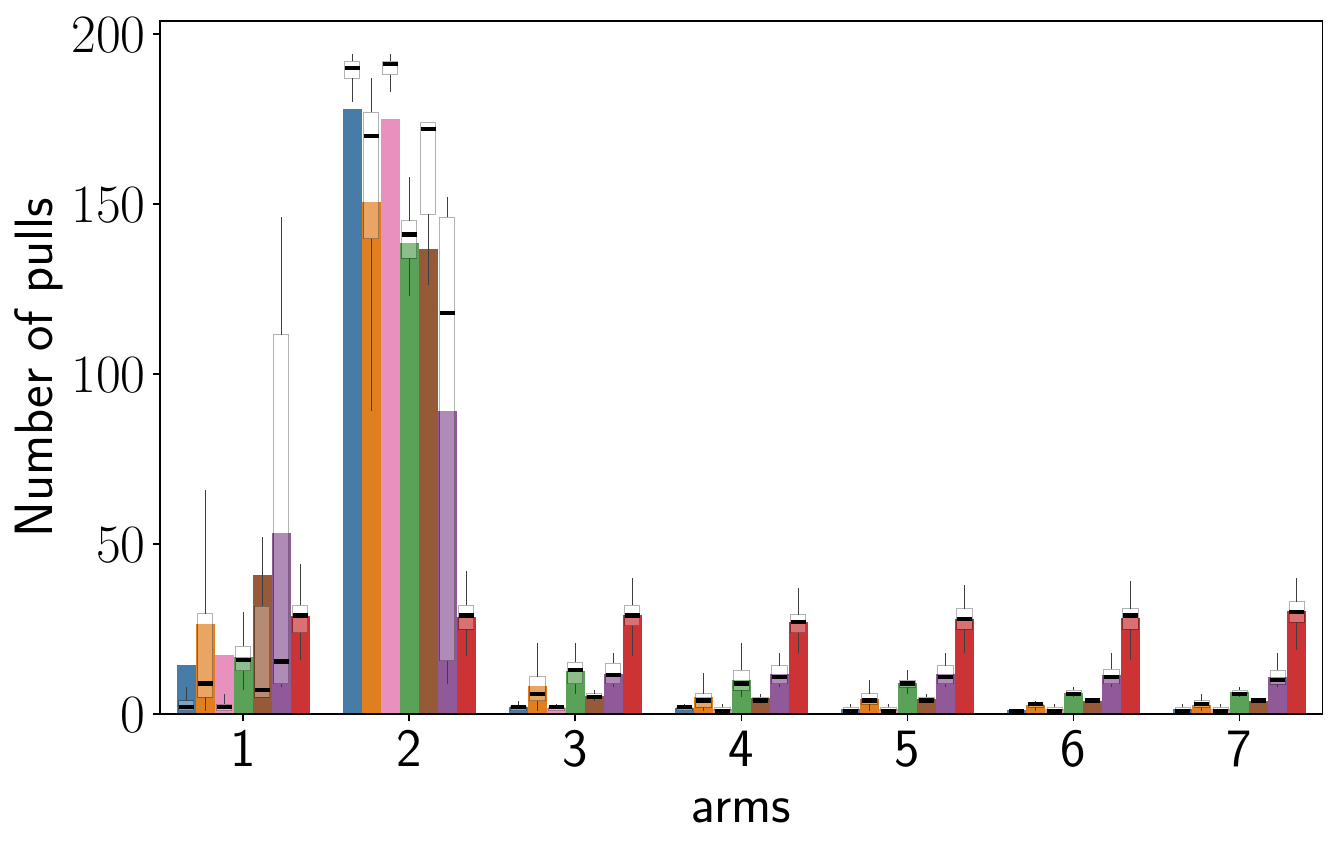}
    &\hspace{-1.4em} \includegraphics[clip, trim=0.3cm -3.5cm 0.0cm 0.0cm, height=3cm]{priors_importance_synth_ablation_legend.pdf}
    \end{tabular}
\caption{Comparing the number of pulls for each arm for different methods, arms are sorted by optimality.}
\label{app:priors_importance_pulls}
\end{figure*}

\begin{figure*}[htb]
\includegraphics[clip, trim=0.0cm 0cm 0cm 0.0cm, height=3.5cm]{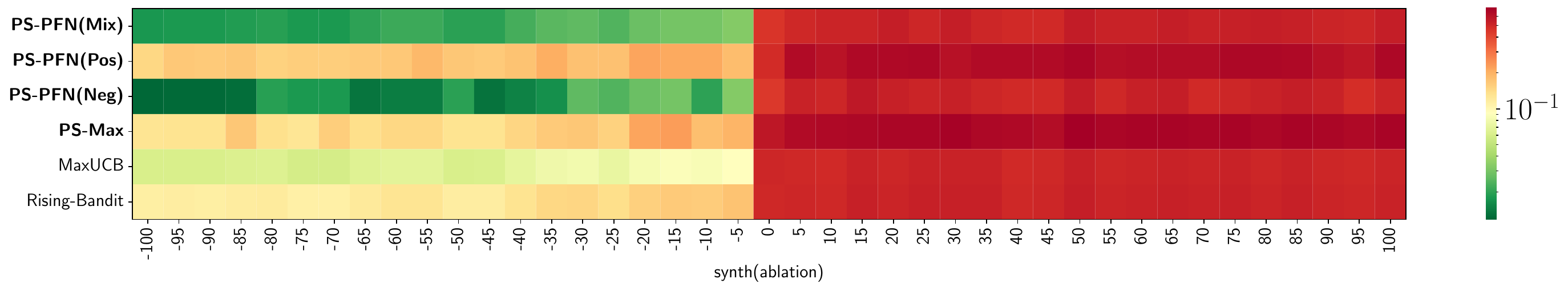}
\caption{Heatmap shows the final regret of each algorithm for different skewness in the arms, lower (green) is better.}
\label{app:priors_importance_heatmap}
\end{figure*}

\subsection{Out of distribution}
\label{app:out_of_distribution}
\begin{figure}[htb]
\centering
\begin{tabular}{c c}
\begin{minipage}{0.3\linewidth}
   \begin{tabular}{l|l}
    \scriptsize \textbf{Dataset} & \textbf{Parameters}\\
    \hline
    \multirow{3}{*}{Neg} 
    & \scriptsize $a \sim \mathcal{U}(-100, -20)$ \\
    & \scriptsize $\mu \sim \mathcal{U}(0, 1)$ \\
    & \scriptsize $\sigma \sim \mathcal{U}(0, 0.2)$ \\
    \hline
    \multirow{3}{*}{Neg(limited)} 
    & \scriptsize $a \sim \mathcal{U}(-100, -20)$ \\
    & \scriptsize $\mu \sim \mathcal{U}(0.8, 1)$ \\
    & \scriptsize $\sigma \sim \mathcal{U}(0, 0.2)$ \\
    \hline
    \end{tabular}
\end{minipage}
&\begin{minipage}{0.6\linewidth}
    \centering
    \includegraphics[clip, trim=0.0cm 0cm 0cm 0.0cm, height=4cm]{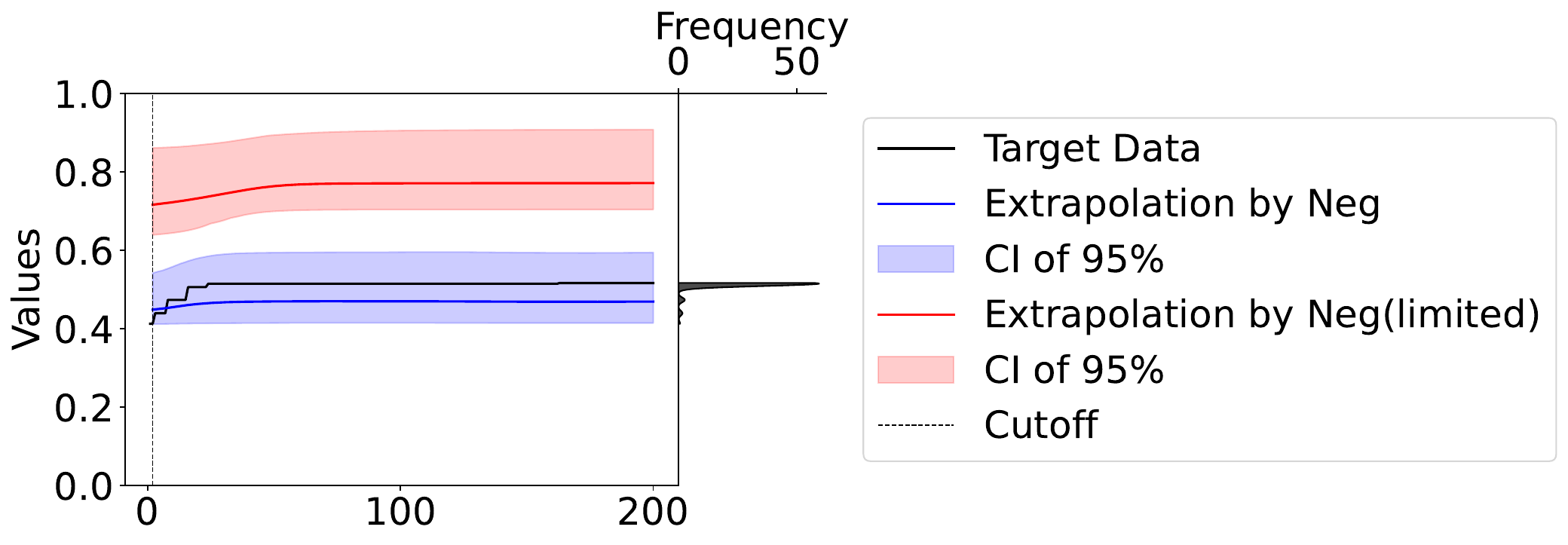}
\end{minipage}
\end{tabular}
\caption{out of distribution analysis}
\label{app:OOD_PFN}
\end{figure}

Here, we examine two PFNs trained on a negatively skewed distribution. However, one of them rarely sees values lower than $0.8$, we call it \textit{Neg(limited)}. In the prediction, as seen in Figure~\ref{app:OOD_PFN} it cannot extrapolate properly. Therefore, for generating synthetic trajectories, we always choose the parameter distribution such that the trajectories cover the whole range $[0,1]$ during pre-training PFNs.

\clearpage
\subsection{Trajectory Analysis}
\label{app:trajectory_analysis}

Here we detail how we designed the priors for training our PFNs, based on analyzing the optimization trajectories of the \Reshuffling{} benchmark. We aim to find robust default parameters for $d_1$ and $d_2$. For simplicity and clarity, we define $\hat\sigma_1$ and  $\hat\sigma_2$ as parameters to control range of $\sigma_1$ and $\sigma_2$ such that  \(\sigma_1 \sim \text{Uniform}(0, \hat\sigma_1)\) and \(\sigma_2 \sim \text{Uniform}(\hat\sigma_2 (1 - \mu_1),  \hat\sigma_2)\).  We aim to identify suitable values for $\hat\sigma_1$  and  $\hat\sigma_2$.

\begin{figure*}[tb]
    \begin{tabular}{c c c c}
    \includegraphics[clip, trim=0.0cm 0cm 0.0cm 0.0cm,height=3.5cm]{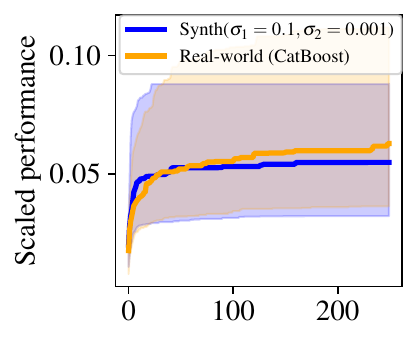}
    & \includegraphics[clip, trim=0.8cm 0cm 0.0cm 0.0cm,height=3.5cm]{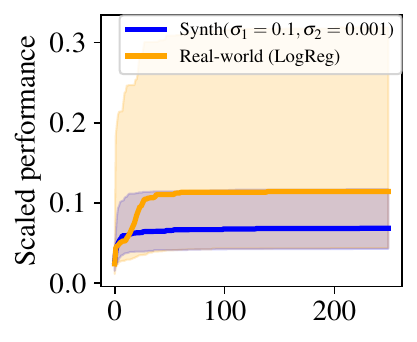}
    & \includegraphics[clip,  trim=0.8cm 0cm 0.0cm 0.0cm,height=3.5cm]{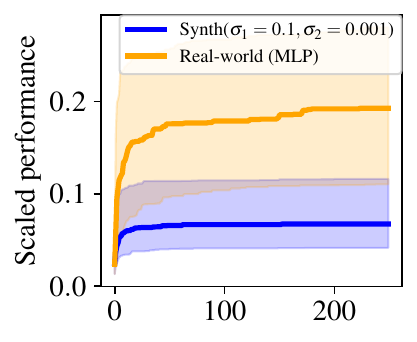}
    &\includegraphics[clip,  trim=0.8cm 0cm 0.0cm 0.0cm,height=3.5cm]{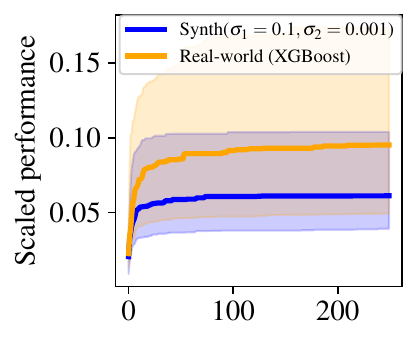}\\
    \includegraphics[clip, trim=0.0cm 0cm 0.0cm 0.0cm,height=3.5cm]{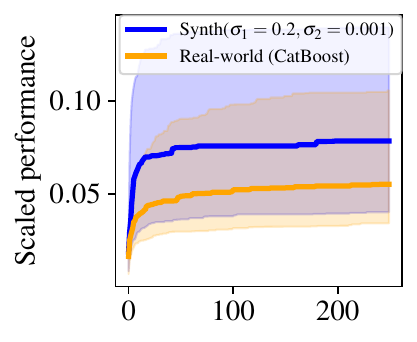}
    & \includegraphics[clip, trim=0.8cm 0cm 0.0cm 0.0cm,height=3.5cm]{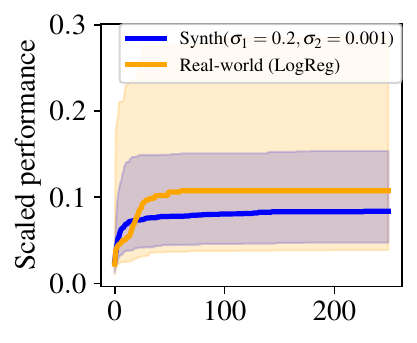}
    & \includegraphics[clip,  trim=0.8cm 0cm 0.0cm 0.0cm,height=3.5cm]{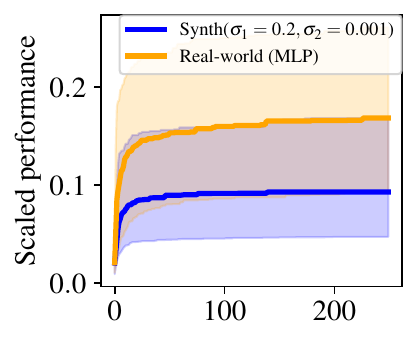}
    &\includegraphics[clip,  trim=0.8cm 0cm 0.0cm 0.0cm,height=3.5cm]{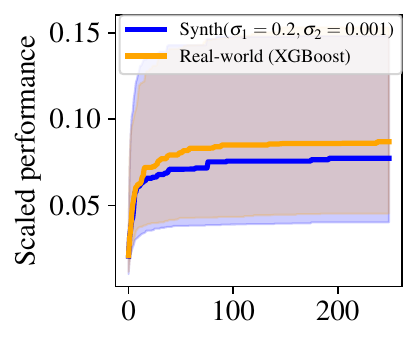}\\
    \includegraphics[clip, trim=0.0cm 0cm 0.0cm 0.0cm,height=3.5cm]{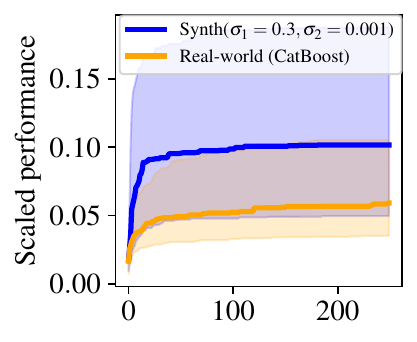}
    & \includegraphics[clip, trim=0.8cm 0cm 0.0cm 0.0cm,height=3.5cm]{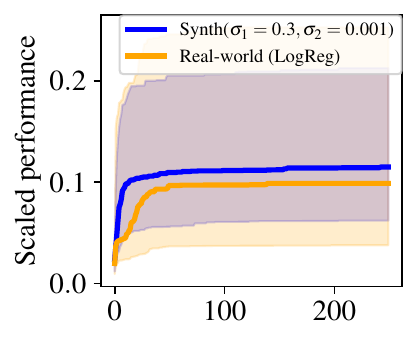}
    & \includegraphics[clip,  trim=0.8cm 0cm 0.0cm 0.0cm,height=3.5cm]{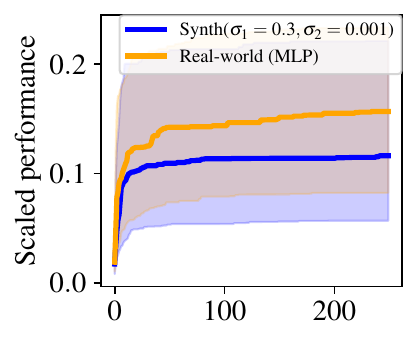}
    &\includegraphics[clip,  trim=0.8cm 0cm 0.0cm 0.0cm,height=3.5cm]{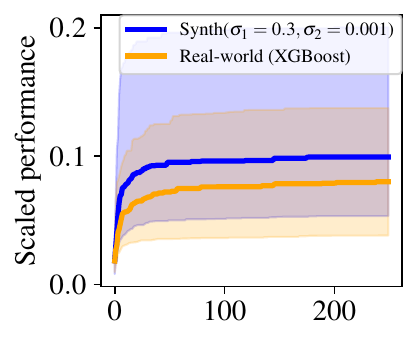}\\
\end{tabular}
\caption{Studying the effect of $\hat\sigma_1$  }
\label{app:priors_importance_variance_sigma_1}
\end{figure*}

We start by using $\hat{\sigma}_2 = 0.001$, which is close to zero. This eliminates the effect of $d_2$ and yields stationary trajectories. In Figure~\ref{app:priors_importance_variance_sigma_1}, we show the effect of $\hat{\sigma}_1$. Each row uses the same value for $\hat\sigma_1$, and each column shows results for a different ML pipeline (arm). Ideally, the blue area fully covers the yellow area, and the shape of the average trajectory is similar in terms of flatness or incremental trend. Comparing the blue curve (average of synthetic trajectories) with the yellow curve (average of real-world trajectories), we found that most often $\hat\sigma_1=0.3$ is too high (last row).  For CatBoost, $\hat\sigma_1=0.1$ yields good results, while for XGBoost, $\hat\sigma_1=0.2$ fits better. We use $\hat\sigma_1=0.1$ and $\hat\sigma_2 = 0.001$ for \textbf{flat} prior.

\begin{figure*}[htb]
    \begin{tabular}{c c c c}
    \includegraphics[clip, trim=0.0cm 0cm 0.0cm 0.0cm,height=3.5cm]{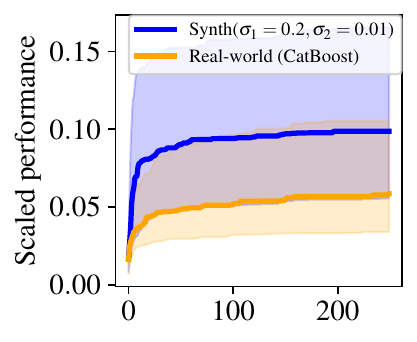}
    & \includegraphics[clip, trim=0.8cm 0cm 0.0cm 0.0cm,height=3.5cm]{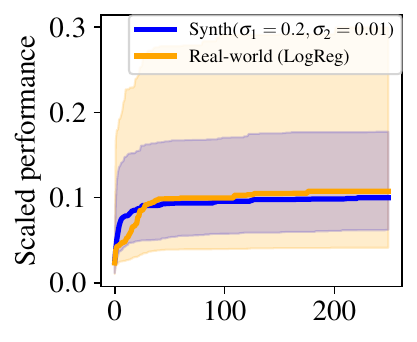}
    & \includegraphics[clip,  trim=0.8cm 0cm 0.0cm 0.0cm,height=3.5cm]{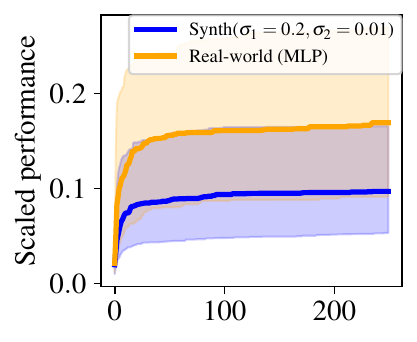}
    &\includegraphics[clip,  trim=0.8cm 0cm 0.0cm 0.0cm,height=3.5cm]{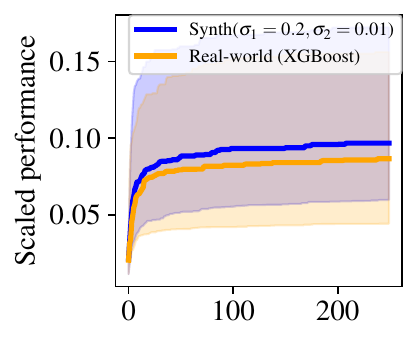}\\
    \includegraphics[clip, trim=0.0cm 0cm 0.0cm 0.0cm,height=3.5cm]{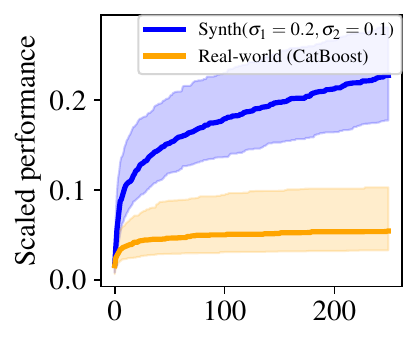}
    & \includegraphics[clip, trim=0.8cm 0cm 0.0cm 0.0cm,height=3.5cm]{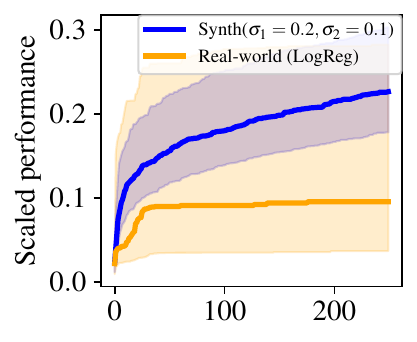}
    & \includegraphics[clip,  trim=0.8cm 0cm 0.0cm 0.0cm,height=3.5cm]{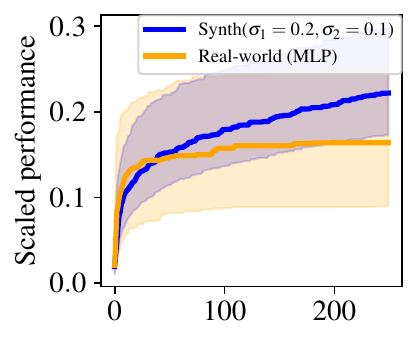}
    &\includegraphics[clip,  trim=0.8cm 0cm 0.0cm 0.0cm,height=3.5cm]{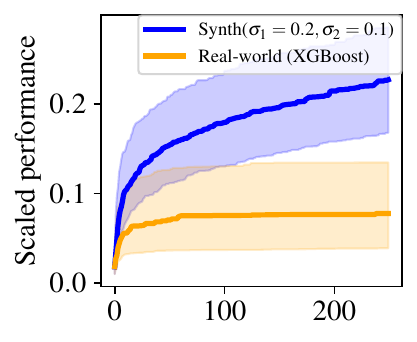}\\
\end{tabular}
\caption{Studying the effect of $\hat\sigma_2$}
\label{app:priors_sigma_2}
\end{figure*}

We now increase the amount of non-stationarity by changing $\hat\sigma_2$, while keeping $\ hat\sigma_1$ fixed at 0.2. As seen in Figure \ref{app:priors_sigma_2}, $\hat\sigma_2 =0.01$ covers better slight non-stationarity. For instance, in the case of MLP and LogReg, the increasing trend of the synthetic trajectories with $\hat\sigma_2 =0.01$ aligns more closely with the real-world trajectories compared to $\hat\sigma_2 =0.001$ (see Figure \ref{app:priors_importance_variance_sigma_1}). We use $\hat\sigma_1=0.2$ and $\hat\sigma_2 =0.01$ for \textbf{semi-flat} prior.

Notably, we also use $\hat\sigma_1=0.2$ and $\hat\sigma_2 =0.1$ for our \textbf{curved} prior as seen in Figure \ref{app:priors_sigma_2}.

\clearpage
\section{Benchmarks}
\label{app:benchmarks}
In our empirical evaluation, we use \numberOfBenchmarks{} AutoML benchmarks (\Tabreporaw{}, \Yahpogym{} and \Complex{}) as described in Table \ref{app:tab:automltasks}. Additionally, we make use of the \Reshuffling{} benchmark during development (as detailed in Section~\ref{app:trajectory_analysis}). We describe all of them in more detail here.
\paragraph{\Complex{}} is a novel~\CASHplus{} benchmark which we build for this work, covering optimization methods for $5$ different workflows covering fine-tuning XTAB~\citep{Zhu2023XTabCP} and TabForestPFN~\citep{breejen2024fine}, running the AutoML system FLAML~\citep{wang-mlsys21a}, optimizing hyperparameters of a shallow neural network~\citep{holzmueller-neurips24a} and PHETabPFNv2~\citep{hollmann2025accurate}. Table~\ref{app:tab:complex_space} provides details for each model, including the optimization method and the corresponding hyperparameter search space. Additionally, Table~\ref{app:tab:complex_space_datasets} lists the datasets from OpenML \citep{bischl2025} used in these experiments. We ran the optimization with four different random seeds, using seven folds to train the ML model and its optimization process, and the remaining three folds to evaluate the loss. Each run produced 12 distinct optimization trajectories as repetitions.

\paragraph{\Yahpogym{}}~\citep{pfisterer-automl22a} is a surrogate benchmark that covers six ML models (details in Table \ref{app:tab:yahpogym_spaces}) on $103$ datasets and uses a regression model (surrogate model) to predict performances for queried hyperparameter settings. We re-use HPO trajectories provided by \citet{maxucb}. In our empirical evaluation, we exclude the last $3$ datasets we used to derive our priors (see Section~\ref{app:trajectory_analysis}).

\paragraph{\Tabreporaw{}} uses the non-discretized search space from \Tabrepo{} ~\citep{salinas2024tabrepo} (details in Table \ref{app:tab:tabrepo_spaces}). We re-use HPO trajectories provided by \citet{maxucb}, which cover $30$ datasets (see context name \verb|D244_F3_C1530_30| in the original \Tabrepo{} code repository). We run Bayesian optimization for $8$ more datasets to study trajectories, which we exclude from the evaluation.

\paragraph{\Reshuffling{}} is an existing benchmark provided by~\citet{nagler2024reshuffling} from which we analyze HPO trajectories to develop our PFNs. It covers $4$ ML models (details in Table \ref{app:tab:hebo_spaces}) across $10$ datasets, with $10$ repetitions and $3$ different validation split ratios within a budget of $250$ iterations. As an HPO method, it uses \textit{HEBO}~\citep{cowenrivers-jair22a}. Since this benchmark does not contain cost information, we do not use it in our empirical evaluation.

\begin{table*}[htbp]
\centering
\footnotesize
\begin{tabular}{lccllcl}
     name & \#models & \#tasks & type & HPO meth. (rep.) & budget & reference \\
     \midrule
     YaHPOGym & 6 & 103 & surrogate & \SMAC{} ($32$) & 200 iterations & \citep{pfisterer-automl22a} \\
     TabRepoRaw & 7 & 30 & raw & \SMAC{} ($32$) & 200 iterations & - \\
     Complex & 5 & 30 & raw & Various ($12$) & 2 hours & - \\
     \bottomrule
\end{tabular}
\caption{Overview of AutoML tasks}
\label{app:tab:automltasks}
\end{table*}

\begin{table}[htbp]
\caption{ML models in \Complex{}.}
\label{app:tab:complex_space}
\centering
\begin{tabular}{llllll}
\toprule
ML model &  Optimization method & Hyperparameters & Search Space & Range & Info \\
\midrule
{XTab} & {Fine-tuning (AdamW)} & & & &  \\
\midrule
\multirow{28}{*}{FLAML } & \multirow{9}{*}{HPO (LightGBM)} 
    & num\_leaves & integer & [4, 32768] & log \\
    & & max\_depth & integer & [-1, 15] & -1 = no limit \\
    & & learning\_rate & continuous & [0.0001, 1.0] & log \\
    & & n\_estimators & integer & [16, 32768] & log \\
    & & min\_child\_samples & integer & [1, 100] & \\
    & & subsample & continuous & [0.1, 1.0] & \\
    & & colsample\_bytree & continuous & [0.1, 1.0] & \\
    & & reg\_alpha & continuous & [0.0, 10.0] & \\
    & & reg\_lambda & continuous & [0.0, 10.0] & \\
\cmidrule{2-6} 
 & \multirow{7}{*}{HPO (XGBoost)} 
    & max\_depth & integer & [3, 15] & \\
    & & learning\_rate & continuous & [0.0001, 1.0] & log \\
    & & n\_estimators & integer & [16, 32768] & log \\
    & & subsample & continuous & [0.1, 1.0] & \\
    & & colsample\_bytree & continuous & [0.1, 1.0] & \\
    & & reg\_alpha & continuous & [0.0, 10.0] & \\
    & & reg\_lambda & continuous & [0.0, 10.0] & \\
\cmidrule{2-6} 
 & \multirow{8}{*}{HPO (CatBoost)} 
    & learning\_rate & continuous & [0.0001, 1.0] & log \\
    & & depth & integer & [4, 10] & \\
    & & l2\_leaf\_reg & continuous & [1.0, 10.0] & \\
    & & iterations & integer & [16, 32768] & log \\
    & & rsm & continuous & [0.5, 1.0] & \\
    & & border\_count & integer & [32, 255] & \\
    & & bootstrap\_type & categorical & \{Bayesian, Bernoulli, MVS\} & \\
    & & bagging\_temperature & continuous & [0, 1] & for Bayesian \\
\cmidrule{2-6} 
 & \multirow{3}{*}{HPO (Random Forest)} 
    & max\_depth & integer & [2, 15] & \\
    & & n\_estimators & integer & [16, 32768] & log \\
    & & max\_features & categorical & \{sqrt, log2, None\} & \\
\cmidrule{2-6} 
 & \multirow{1}{*}{HPO (Logistic Regression)} 
    & C & continuous & [1e-6, 1e6] & log \\
\midrule
\multirow{10}{*}{RealMLP} & \multirow{10}{*}{HPO (SMAC)} 
     & num\_emb\_type & categorical & \{none, pbld, pl, plr\} & \\ 
    & & add\_front\_scale & categorical & \{True, False\} & \\ 
    & & lr & continuous & [0.02, 0.3] & log \\ 
    & & p\_drop & categorical & \{0.00, 0.15, 0.30\} & \\ 
    & & act & categorical & \{selu, relu, mish\} & \\ 
    & & hidden\_sizes & categorical & \{[256,256,256], [64,64,64,64,64], [512]\} & \\ 
    & & wd & ordinal & \{0.0, 0.02\} & \\ 
    & & plr\_sigma & continuous & [0.05, 0.5] & log \\ 
    & & ls\_eps & ordinal & \{0.0, 0.1\} & \\ 
    & & epochs & ordinal & \{16, 32, 64, 128, 256\} & default=256 \\
    & & batch\_size & ordinal & \{32, 64, 128, 256\} & default=256 \\
\midrule
{TabForestPFN} & {Fine-tuning (AdamW)} & &  & & \\
\midrule
{TabPFN (v2)} & {Post-Hoc Ensembling} &  & & & \\
\bottomrule
\end{tabular}
\end{table}

\begin{table}[htbp]
\caption{Dataset in \Complex{}.}
\label{app:tab:complex_space_datasets}
\begin{tabular}{lrlrrrr}
\toprule
index & task id & dataset name & number of samples & number of features & number of categorical features & number of calsses \\
\midrule
1 & 3593 & 2dplanes & 40768 & 11 & 1 & 2 \\
2 & 3627 & cpu-act & 8192 & 22 & 1 & 2 \\
3 & 3688 & houses & 20640 & 9 & 1 & 2 \\
4 & 3844 & kdd-internet-usage & 10108 & 69 & 69 & 2 \\
5 & 3882 & pendigits & 10992 & 17 & 1 & 2 \\
6 & 3904 & jm1 & 10885 & 22 & 1 & 2 \\
7 & 7295 & Click-prediction-small & 39948 & 10 & 1 & 2 \\
8 & 9933 & volcanoes-c1 & 28626 & 4 & 1 & 5 \\
9 & 9965 & skin-segmentation & 245057 & 4 & 1 & 2 \\
10 & 9977 & nomao & 34465 & 119 & 30 & 2 \\
11 & 9986 & gas-drift & 13910 & 129 & 1 & 6 \\
12 & 9987 & gas-drift-different-concentrations & 13910 & 130 & 1 & 6 \\
13 & 10106 & CreditCardSubset & 14240 & 31 & 1 & 2 \\
14 & 14965 & bank-marketing & 45211 & 17 & 10 & 2 \\
15 & 34537 & PhishingWebsites & 11055 & 31 & 31 & 2 \\
16 & 34539 & Amazon-employee-access & 32769 & 10 & 10 & 2 \\
17 & 361056 & california & 20634 & 9 & 1 & 2 \\
18 & 361063 & house-16H & 13488 & 17 & 1 & 2 \\
19 & 361071 & jannis & 57580 & 55 & 1 & 2 \\
20 & 361110 & electricity & 38474 & 9 & 2 & 2 \\
21 & 361111 & eye-movements & 7608 & 24 & 4 & 2 \\
22 & 361112 & KDDCup09-upselling & 5032 & 46 & 12 & 2 \\
23 & 361113 & covertype & 423680 & 55 & 45 & 2 \\
24 & 361114 & rl & 4970 & 13 & 8 & 2 \\
25 & 361115 & road-safety & 111762 & 33 & 4 & 2 \\
26 & 361116 & compass & 16644 & 18 & 10 & 2 \\
27 & 361282 & albert & 58252 & 32 & 11 & 2 \\
28 & 362098 & Dota2-Games-Results-Data-Set & 102944 & 117 & 117 & 2 \\
29 & 362407 & adult & 48790 & 15 & 9 & 2 \\
30 & 363550 & cdc-diabetes & 253680 & 22 & 1 & 2 \\
\bottomrule
\end{tabular}
\end{table}

\begin{table}[htbp]
\caption{Hyperparameter spaces for ML models in \Yahpogym{}.}
\label{app:tab:yahpogym_spaces}
\centering
\begin{tabular}{lllll}
\toprule
ML model &  Hyperparameter & Type & Range & Info \\
\midrule
\multirow{2}{*}{-} & trainsize & continuous & [0.03, 1] &  =0.525 (fixed) \\ 
    & imputation & categorical & \{mean, median, hist\} &  =mean (fixed)\\ 
\midrule
\multirow{2}{*}{Glmnet} & alpha & continuous & [0, 1] & \\
    & s & continuous & [0.001, 1097] & log \\ 
\midrule
\multirow{4}{*}{Rpart} & cp & continuous & [0.001, 1] & log \\ 
    & maxdepth & integer & [1, 30] &  \\ 
    & minbucket & integer & [1, 100] &  \\
    & minsplit & integer & [1, 100] &  \\ 
\midrule
\multirow{5}{*}{SVM} & kernel & categorical & \{linear, polynomial, radial\} &  \\ 
    & cost & continuous & [4.5e-05, 2.2e4] & log \\ 
    & gamma & continuous & [4.5e-05, 2.2e4] & log, kernel \\ 
    & tolerance & continuous & [4.5e-05, 2] & log \\ 
    & degree & integer & [2, 5] &  kernel \\ 
\midrule
\multirow{5}{*}{AKNN} & k & integer & [1, 50] &  \\ 
    & distance & categorical & \{l2, cosine, ip\} &  \\ 
    & M & integer & [18, 50] &  \\ 
    & ef & integer & [7, 403] & log \\ 
    & ef\_construction & integer & [7, 403] & log \\ 
\midrule
\multirow{7}{*}{Ranger} & num.trees & integer & [1, 2000] &  \\ 
    & sample.fraction & continuous & [0.1, 1] &  \\ 
    & mtry.power & integer & [0, 1] &  \\ 
    & respect.unordered.factors & categorical & \{ignore, order, partition\} &  \\ 
    & min.node.size & integer & [1, 100] &  \\ 
    & splitrule & categorical & \{gini, extratrees\} &  \\ 
    & num.random.splits & integer & [1, 100] & splitrule \\ 
\midrule
\multirow{13}{*}{XGBoost} & booster & categorical & \{gblinear, gbtree, dart\} &  \\ 
    & nrounds & integer & [7, 2980] & log \\ 
    & eta & continuous & [0.001, 1] & log, booster \\ 
    & gamma & continuous & [4.5e-05, 7.4] & log, booster\\ 
    & lambda & continuous & [0.001, 1097] & log \\ 
    & alpha & continuous & [0.001, 1097] & log \\ 
    & subsample & continuous & [0.1, 1] &  \\ 
    & max\_depth & integer & [1, 15] &  booster \\ 
    & min\_child\_weight & continuous & [2.72, 148.4] & log,  booster \\ 
    & colsample\_bytree & continuous & [0.01, 1] &   booster \\ 
    & colsample\_bylevel & continuous & [0.01, 1] &  booster \\ 
    & rate\_drop & continuous & [0, 1] &  booster \\ 
    & skip\_drop & continuous & [0, 1] &  booster \\ 
\bottomrule
\end{tabular}
\end{table}

\begin{table}[htbp]
\caption{Hyperparameter spaces for ML models in \Tabreporaw{}.}
\label{app:tab:tabrepo_spaces}
\centering
\begin{tabular}{llllll}
\toprule
ML model &  Hyperparameter & Type & Range & Info  & Default value \\
\midrule
\multirow{7}{*}{NN(PyTorch)} & learning rate & continuous & [1e-4, 3e-2] & log  & 3e-4 \\
    & weight decay & continuous & [1e-12, 0.1] & log  & 1e-6 \\
    & dropout prob & continuous & [0, 0.4] &  & 0.1 \\
    & use batchnorm & categorical & False, True  &  & \\  
    & num layers & integer & [1, 5]  &  & 2\\  
    & hidden size & integer & [8, 256]  &  & 128\\  
    & activation & categorical & relu, elu &  & \\  
\midrule
\multirow{9}{*}{NN(FastAI)}  & learning rate & continuous & [5e-4, 1e-1] & log  & 1e-2 \\
    & \multirow[t]{3}{*}{layers}  & \multirow[t]{3}{*}{categorical}   & [200], [400], [200, 100],  &  & \\  
    &   &    &  [400, 200], [800, 400], &  & \\  
    &   &  &   [200, 100, 50], [400, 200, 100] &  & \\  
    & emb drop & continuous & [0.0, 0.7] &   & 0.1 \\
    & ps & continuous &  [0.0, 0.7] &   & 0.1 \\
    & bs  & categorical & 256, 128, 512, 1024, 2048  &  & \\  
    & epochs & integer & [20, 50]  &  & 30\\    
\midrule
\multirow{6}{*}{CatBoost} & learning rate & continuous & [5e-3 ,0.1] & log  & 0.05 \\
    & depth & integer & [4, 8]  &  & 6\\  
    & l2 leaf reg & continuous & [1, 5] &  & 3\\
    & max ctr complexity & integer & [1, 5]  &  & 4\\  
    & one hot max size & categorical & 2, 3, 5, 10 &  & \\  
    & grow policy & categorical & SymmetricTree, Depthwise &  & \\  
\midrule
\multirow{5}{*}{LightGBM} & learning rate & continuous & [5e-3 ,0.1] & log  & 0.05 \\
    & feature fraction & continuous & [0.4, 1.0] &   & 1.0\\
    & min data in leaf & integer & [2, 60]  &  & 20\\  
    & num leaves & integer & [16, 255]  &  & 31\\  
    & extra trees & categorical & False, True &  & \\  
\midrule
\multirow{5}{*}{XGBoost} & learning rate & continuous & [5e-3 ,0.1] & log  & 0.1 \\
    & max depth & integer & [4, 10]  &  & 6\\  
    & min child weight & continuous & [0.5, 1.5] &   & 1.0\\
    & colsample bytree & continuous & [0.5, 1.0] &   & 1.0\\
    & enable categorical & categorical & False, True &  & \\  
\midrule
\multirow{3}{*}{Extra-trees} &  max leaf nodes& integer & [5000, 50000]  &  & \\ 
    & min samples leaf & categorical & 1, 2, 3, 4, 5, 10, 20, 40, 80 &  & \\  
    & max features & categorical & sqrt, log2, 0.5, 0.75, 1.0 &  & \\  
\midrule
\multirow{3}{*}{Random-forest} &  max leaf nodes& integer & [5000, 50000]  &  & \\ 
    & min samples leaf & categorical & 1, 2, 3, 4, 5, 10, 20, 40, 80 &  & \\  
    & max features & categorical & sqrt, log2, 0.5, 0.75, 1.0 &  & \\  
\bottomrule
\end{tabular}
\end{table}

\begin{table}[htbp]
\caption{Hyperparameter spaces for ML models in \Reshuffling{}.}
\label{app:tab:hebo_spaces}
\centering
\begin{tabular}{llllll}
\toprule
ML model &  Hyperparameter & Type & Range & Info  \\
\midrule
\multirow{6}{*}{Funnel-Shaped MLP} & learning rate & continuous & [1e-4, 1e-1] & log  \\
    & num layers & integer & [1, 5]  & \\ 
    & max units & categorical & 64, 128, 256, 512  & \\ 
    & batch size & categorical. & {16, 32, ..., max\_batch\_size}  & \\ 
    & momentum & continuous. & [0.1, 0.99] &  \\
    & alpha & continuous. &  [1e-6, 1e-1] & log \\
\midrule
\multirow{2}{*}{Elastic Net} & C & continuous & [1e-6, 10e4] & log  \\
     & l1 ratio & continuous & [0.0, 1.0] &   \\
\midrule
\multirow{4}{*}{XGBoost} & max depth & integer & [2, 12]  &  log \\  
    & alpha & continuous & [1e-8, 1.5] &  log\\
    & lambda & continuous & [1e-8, 1.0] &  log\\
    & eta & continuous & [0.01, 0.3] &  log\\
\midrule
\multirow{3}{*}{CatBoost} & learning rate & continuous & [0.01 ,0.3] & log  \\
    & depth & integer & [2, 12]  & \\  
    & l2 leaf reg & continuous & [0.5, 30] & \\
\bottomrule
\end{tabular}
\end{table}

\clearpage
\section{PFN architecture}
\label{app:priors_architecture}

The PFN architecture inherits standard Transformer hyperparameters: the number of layers (\texttt{nlayers}), attention heads (\texttt{nheads}), embedding dimension (\texttt{emsize}), and hidden dimension (\texttt{nhidden}). Specifically, our configuration uses $6$ layers, $4$ attention heads, and a hidden size of $512$.  For training, we adopt the Adam optimizer~\citep{kingma-iclr15a} (learning rate $10^{-4}$, batch size $100$) with cosine annealing~\citep{loshchilov-iclr17a}, including a linear warmup phase over the first 25\% of epochs. We set $m = 200$, meaning the PFN is trained to extrapolate sequences of up to $200$ steps (e.g., HPO or fine-tuning iterations). The PFN output is a discretized mass distribution with a fixed number of bins. We set the number of bins to $1,000$ as a hyperparameter. The model is trained for $1,000$ epochs, resulting in training with $100,000$ synthetic trajectories, which takes approximately one GPU-day.

\section{Priors}
\label{app:priors_choices}

We use three types of priors throughout our work: (a) a prior to efficiently estimate costs to make our methods cost-aware, (b) a prior to estimate reward distributions to evaluate~\OurAlgoMax{}, and (c) a prior to generate synthetic reward trajectories to train PFNs to be used for~\OurAlgoPFNs{}. We discuss the use of these in the following.

\paragraph{Priors for cost-awareness.} We assume that the cost (i.e., runtime) follows a log-normal distribution, with corresponding parameters provided in Table~\ref{app:tab:prior_params_cost}.  The log-normal distribution is a natural choice for modeling the cost of training machine learning models, as training costs often exhibit a skewed, right-tailed behavior where higher costs are less frequent but more extreme. This makes the log-normal distribution a good fit for such scenarios (as illustrated in Figure 1 from~\citep{HansLee2020}). We truncated the log-normal distribution to avoid unreasonably high or low costs, as shown in Table~\ref{app:tab:prior_params_cost}.

\begin{table}[h]
\centering
\begin{tabular}{|c|c|c|}
\hline
\textbf{Likelihood Model} & \textbf{Prior Parameters} & \textbf{Notes} \\
\hline
 Log-Normal &
\begin{tabular}{l}
$\mu_0 = 1.0$ \\
$\lambda_0 = 1.0$ \\
$\alpha_0 = 1.0$ \\
$\beta_0 = 0.0$
\end{tabular}
&
\begin{minipage}[t]{0.3\linewidth}
Log-cost modeled with \\
Normal-Inverse-Gamma prior. \\
Costs truncated to: \\
$0.1 \leq \text{cost} \leq B/10$\\
\end{minipage}
\\
\hline
\end{tabular}
\caption{Prior distributions and posterior sampling procedures for cost model.}
\label{app:tab:prior_params_cost}
\end{table}

\paragraph{Priors for \OurAlgoMax{}.} We assume that the rewards follow a Gaussian distribution, with initial parameter values specified in Table~\ref{app:tab:prior_params_reward}. 

\begin{table}[h]
\centering
\begin{tabular}{|c|c|c|}
\hline
\textbf{Likelihood Model} & \textbf{Prior Parameters} & \textbf{Notes} \\
\hline
Gaussian &
\begin{tabular}{l}
$\mu_0 = 1.0$ \\
$\lambda_0 = 1.0$ \\
$\alpha_0 = 1.0$ \\
$\beta_0 = 0.0$
\end{tabular}
&
\begin{minipage}[t]{0.3\linewidth}
Normal-Inverse-Gamma prior. \\
Posterior updates:
\begin{itemize}[leftmargin=*, nosep]
    \item $\sigma^2 \sim \text{InvGamma}(\alpha_n, \beta_n)$
    \item $\theta \sim \mathcal{N}(\mu_n, \sigma^2 / \lambda_n)$\\
\end{itemize}
\end{minipage}
\\
\hline
\end{tabular}
\caption{Prior distributions and posterior sampling procedures for reward model.}
\label{app:tab:prior_params_reward}
\end{table}

\renewcommand\fbox{\fcolorbox{green}{green!10}}
\setlength{\fboxsep}{1pt} 
\begin{figure*}[htb]
\centering
    \begin{tabular}{c c c}
    \includegraphics[clip, trim=0.0cm 0cm 0.0cm 0.0cm,height=3.2cm]{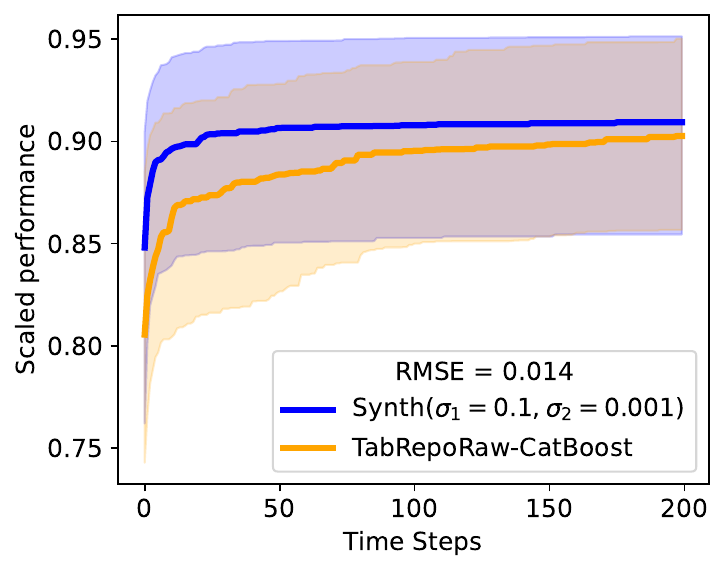}
    & \fbox{\includegraphics[clip, trim=0.8cm 0cm 0.0cm 0.0cm,height=3.2cm]{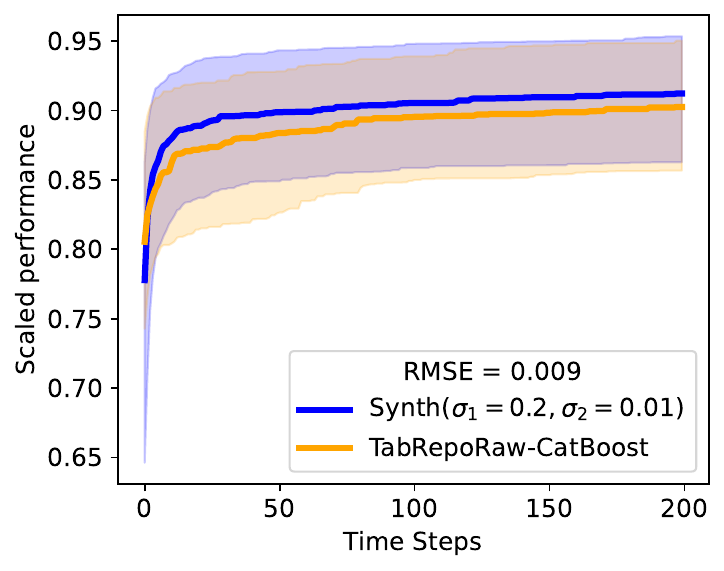}}
    & \includegraphics[clip,  trim=0.8cm 0cm 0.0cm 0.0cm,height=3.2cm]{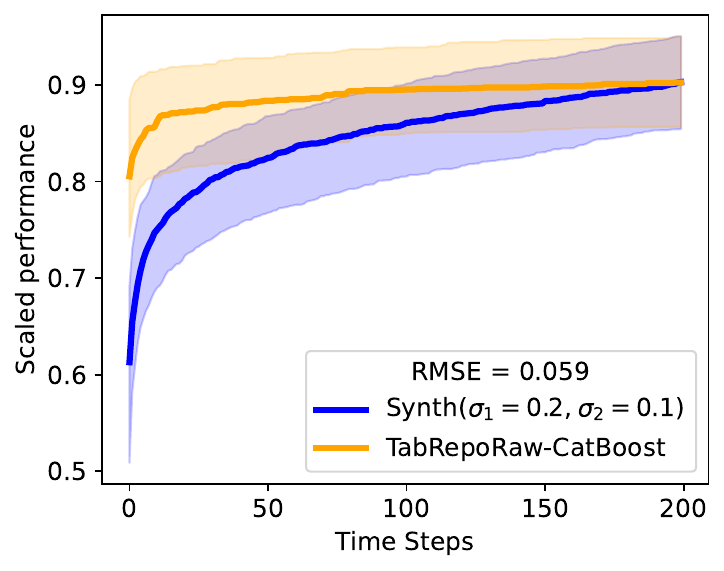}\\
    \includegraphics[clip,  trim=0.0cm 0cm 0.0cm 0.0cm,height=3.2cm]{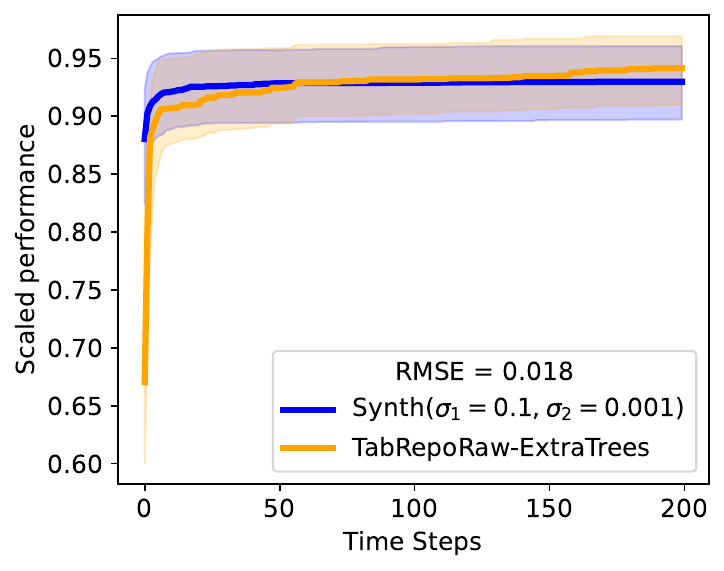}
    & \fbox{\includegraphics[clip, trim=0.8cm 0cm 0.0cm 0.0cm,height=3.2cm]{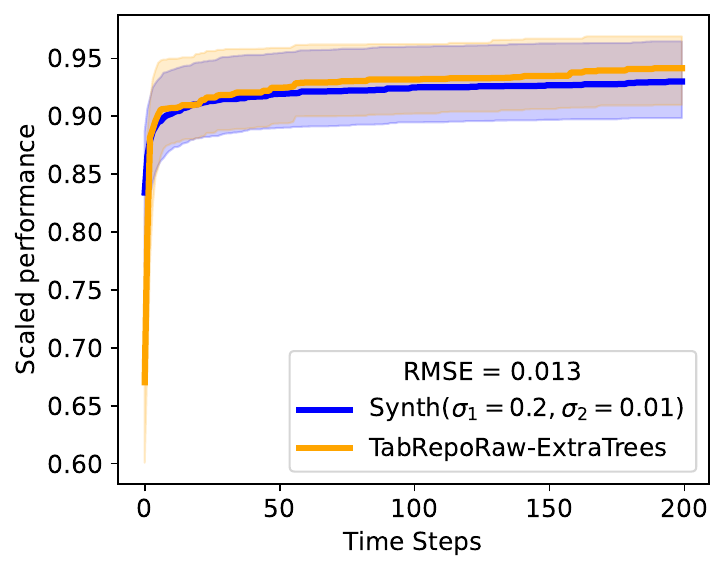}}
    & \includegraphics[clip, trim=0.8cm 0cm 0.0cm 0.0cm,height=3.2cm]{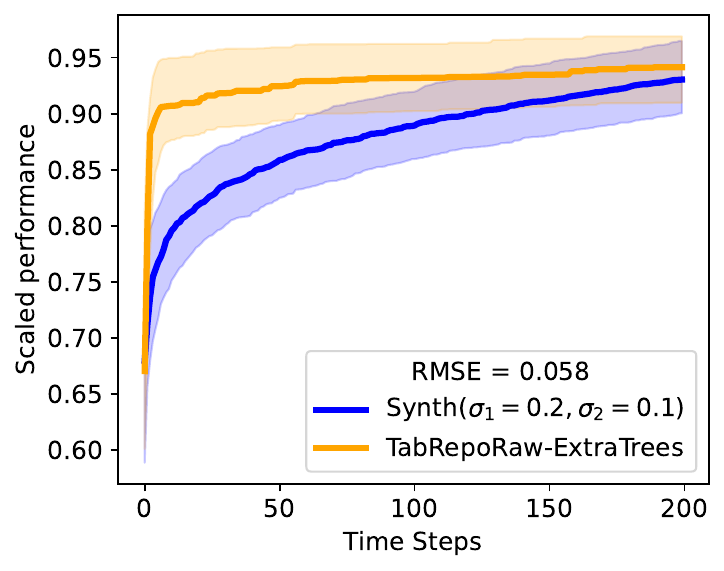}\\
    \includegraphics[clip,  trim=0.0cm 0cm 0.0cm 0.0cm,height=3.2cm]{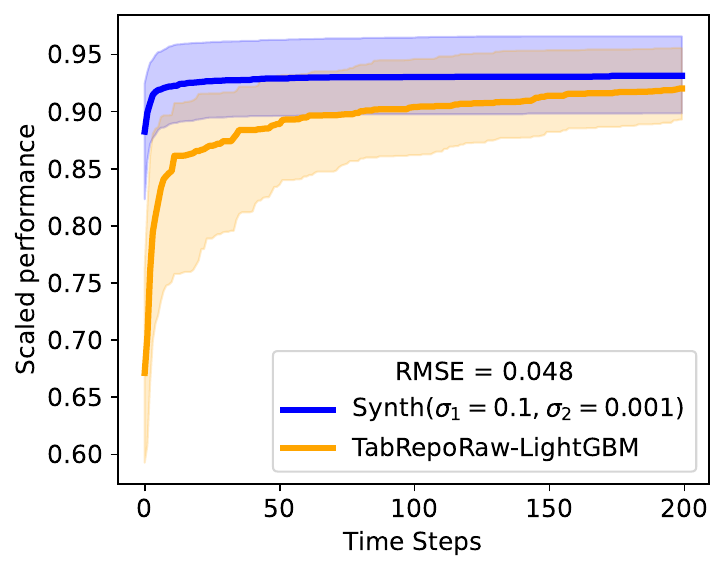}
    &\includegraphics[clip, trim=0.8cm 0cm 0.0cm 0.0cm,height=3.2cm]{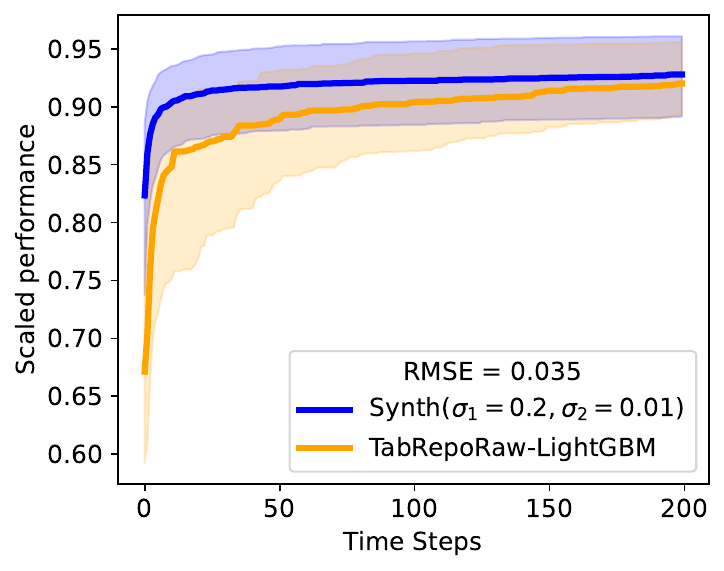}
    &  \fbox{\includegraphics[clip, trim=0.8cm 0cm 0.0cm 0.0cm,height=3.2cm]{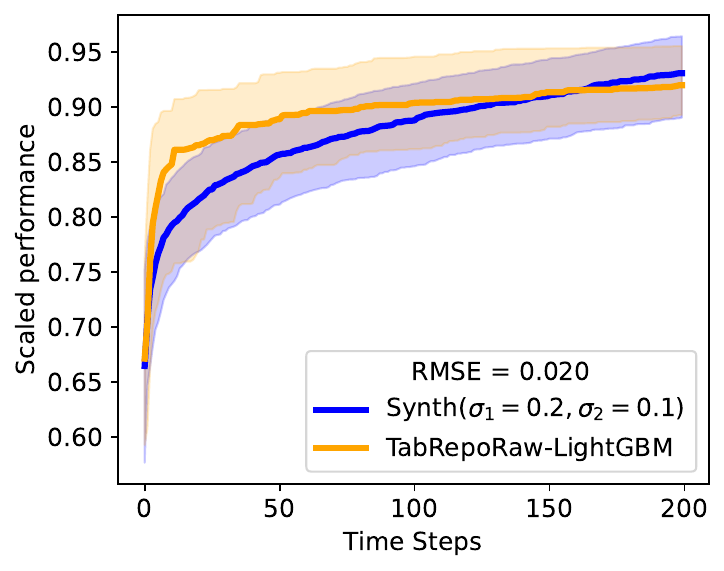}}\\
    \includegraphics[clip,  trim=0.0cm 0cm 0.0cm 0.0cm,height=3.2cm]{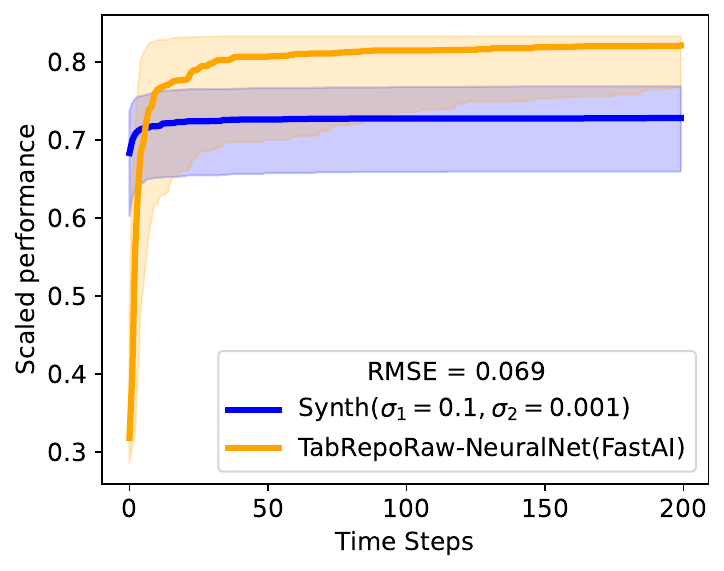}
    &\includegraphics[clip, trim=0.8cm 0cm 0.0cm 0.0cm,height=3.2cm]{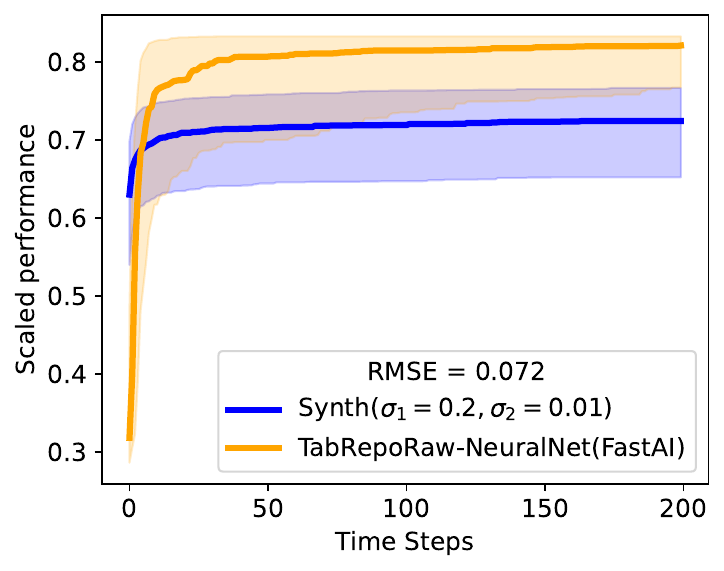}
    &  \fbox{\includegraphics[clip, trim=0.8cm 0cm 0.0cm 0.0cm,height=3.2cm]{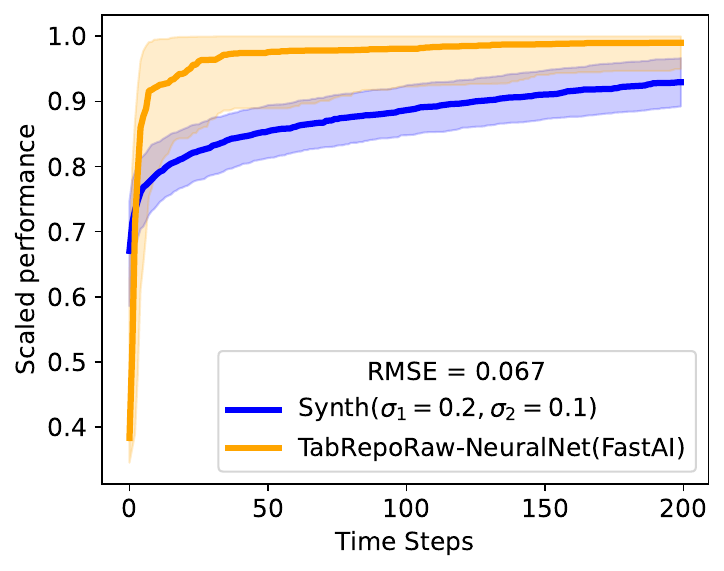}}\\
    \includegraphics[clip,  trim=0.0cm 0cm 0.0cm 0.0cm,height=3.2cm]{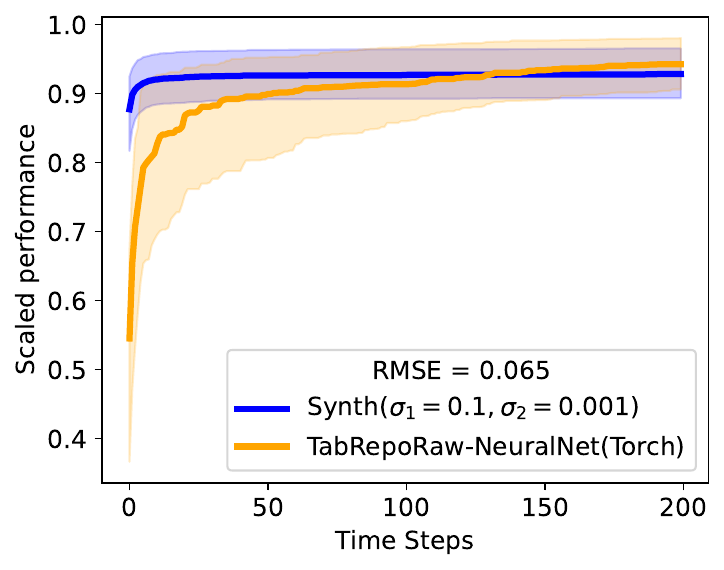}
    &\includegraphics[clip, trim=0.8cm 0cm 0.0cm 0.0cm,height=3.2cm]{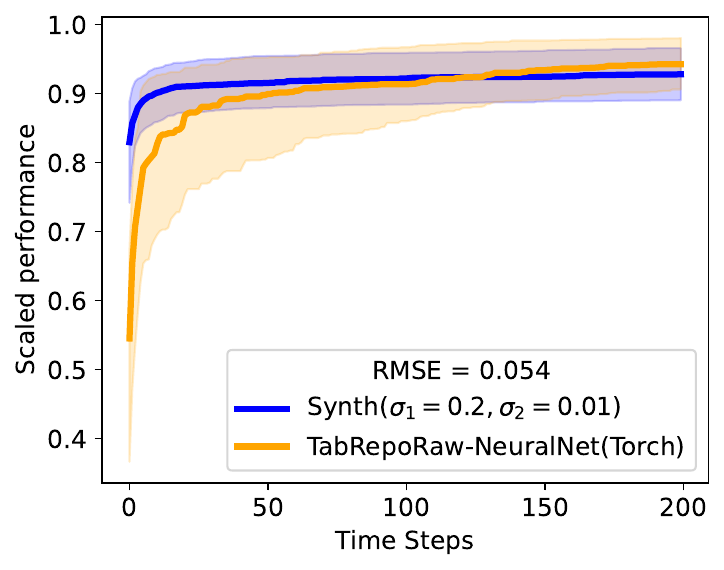}
    & \fbox{ \includegraphics[clip, trim=0.8cm 0cm 0.0cm 0.0cm,height=3.2cm]{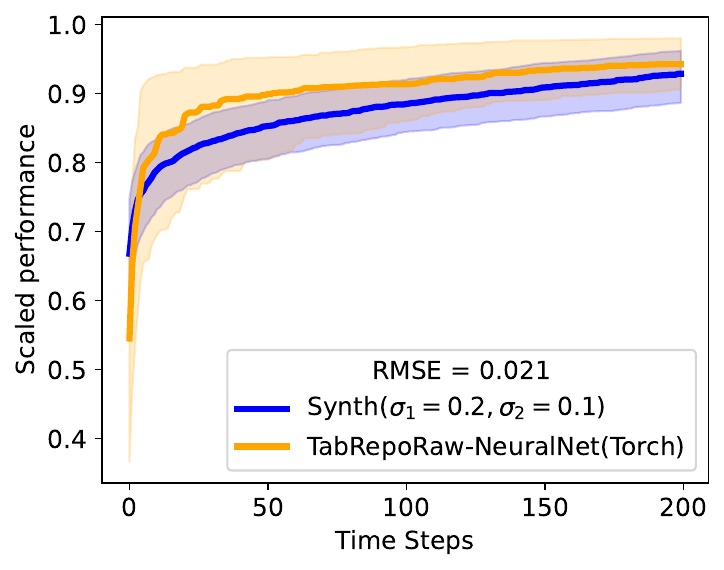}}\\
    \includegraphics[clip,  trim=0.0cm 0cm 0.0cm 0.0cm,height=3.2cm]{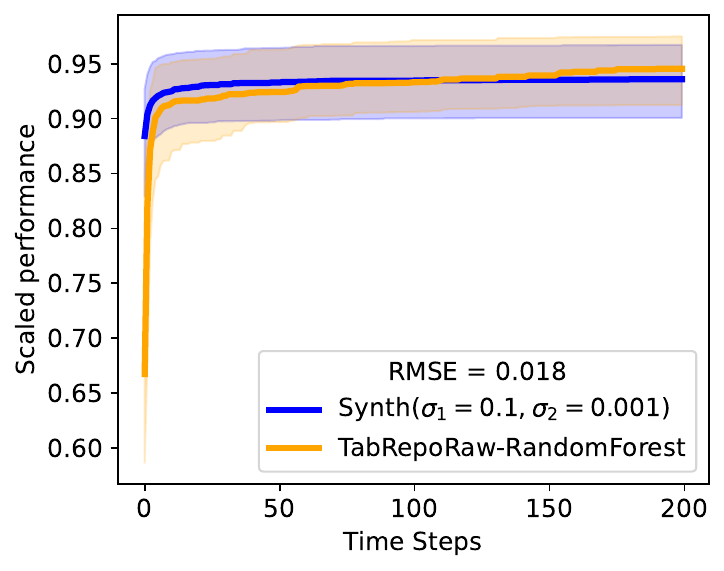}
    & \fbox{\includegraphics[clip, trim=0.8cm 0cm 0.0cm 0.0cm,height=3.2cm]{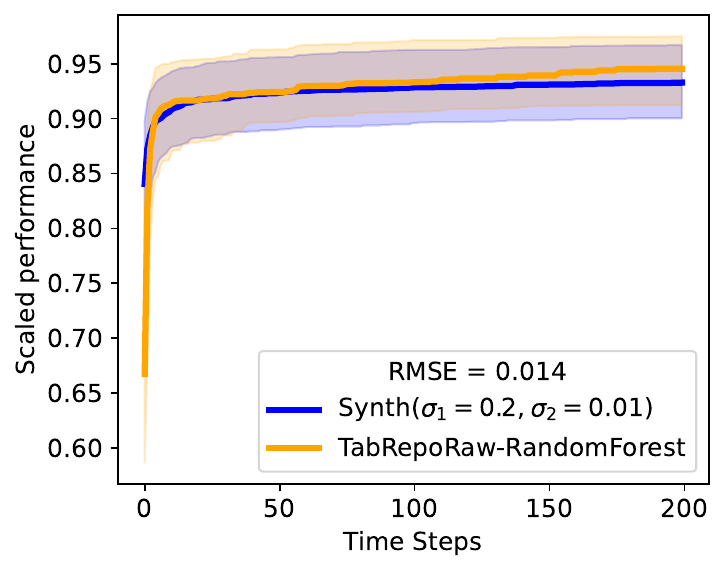}}
    & \includegraphics[clip, trim=0.8cm 0cm 0.0cm 0.0cm,height=3.2cm]{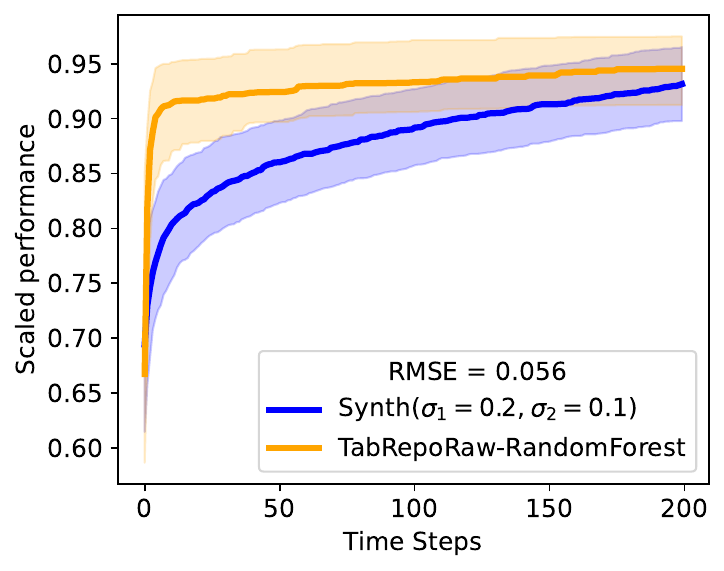}\\
     \fbox{\includegraphics[clip,  trim=0.0cm 0cm 0.0cm 0.0cm,height=3.2cm]{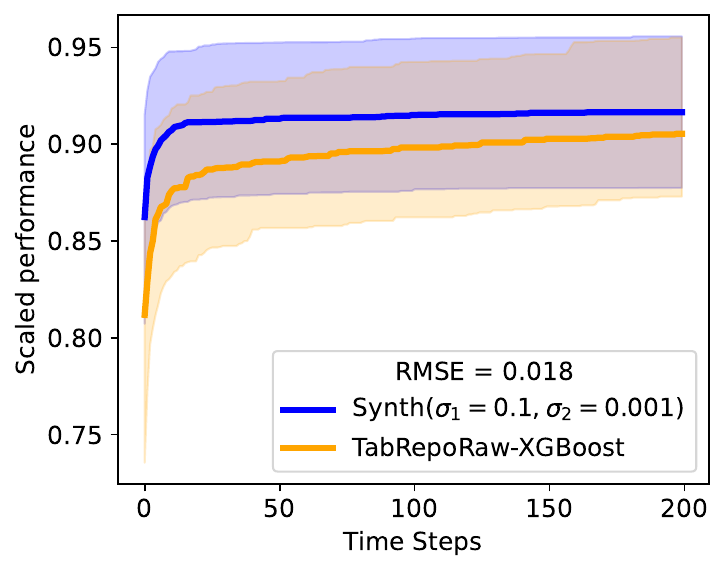}}
    &\includegraphics[clip, trim=0.8cm 0cm 0.0cm 0.0cm,height=3.2cm]{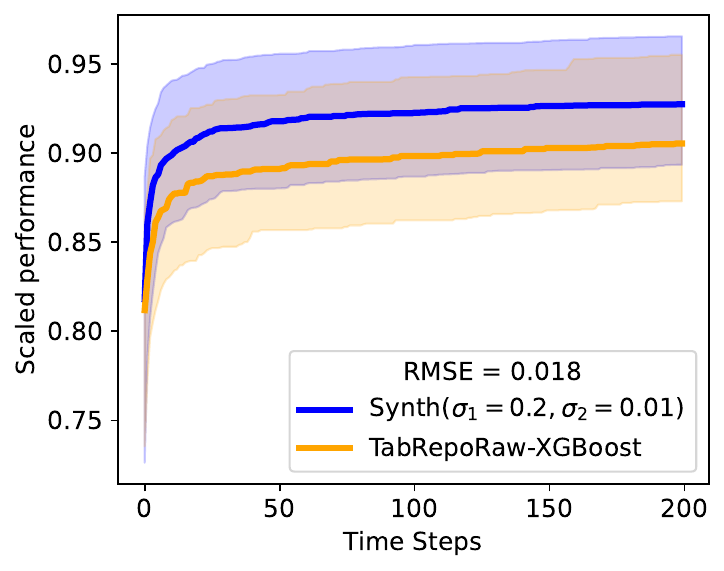}
    & \includegraphics[clip, trim=0.8cm 0cm 0.0cm 0.0cm,height=3.2cm]{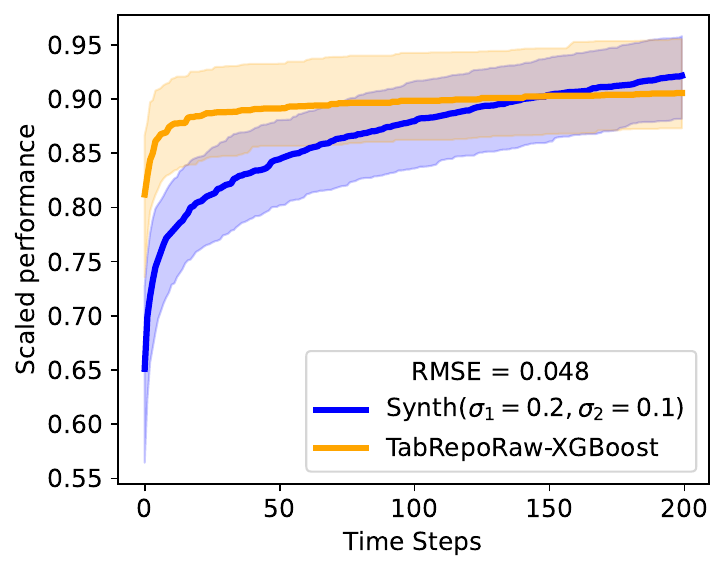}\\
\end{tabular}
\caption{Priors for \Tabreporaw{}}
\label{app:prior_selection_tabreporaw}
\end{figure*}

\begin{figure*}[htb]
\centering
    \begin{tabular}{c c c}
    \includegraphics[clip, trim=0.0cm 0cm 0.0cm 0.0cm,height=3.5cm]{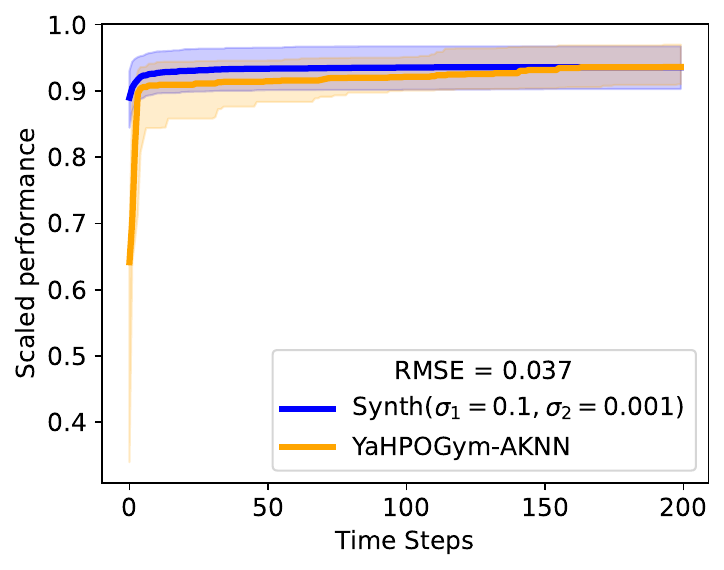}
    & \includegraphics[clip, trim=0.8cm 0cm 0.0cm 0.0cm,height=3.5cm]{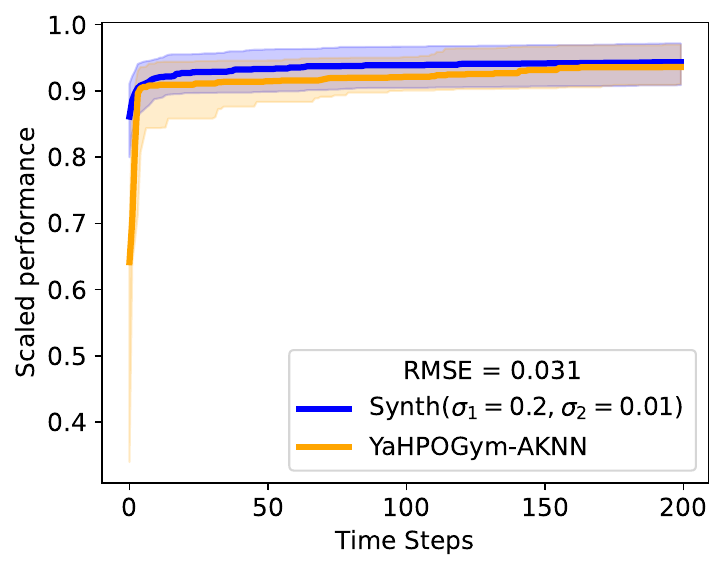}
    &  \fbox{ \includegraphics[clip,  trim=0.8cm 0cm 0.0cm 0.0cm,height=3.5cm]{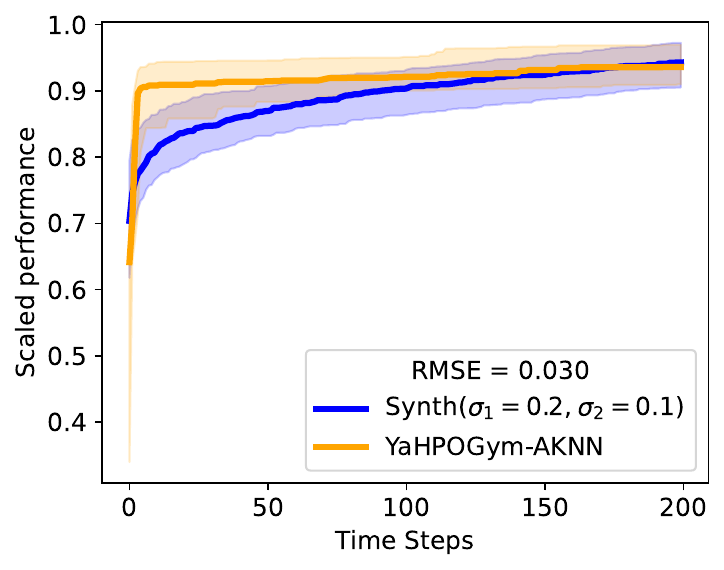}}\\
    \includegraphics[clip,  trim=0.0cm 0cm 0.0cm 0.0cm,height=3.5cm]{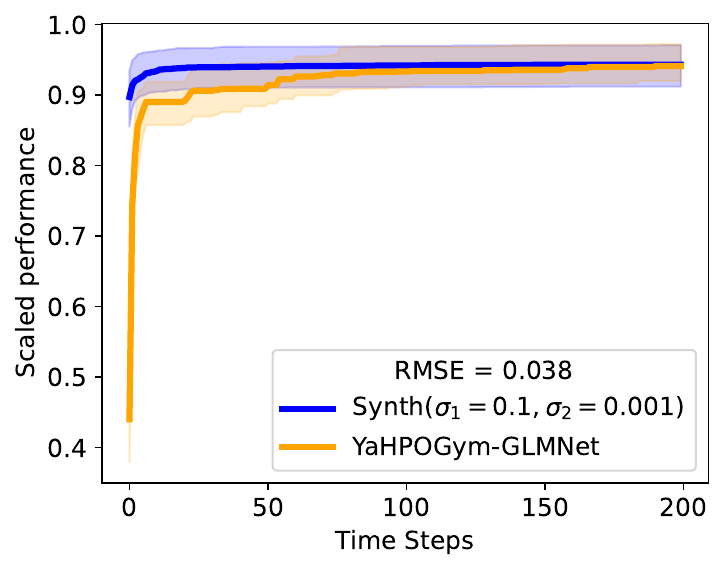}
    & \fbox{ \includegraphics[clip, trim=0.8cm 0cm 0.0cm 0.0cm,height=3.5cm]{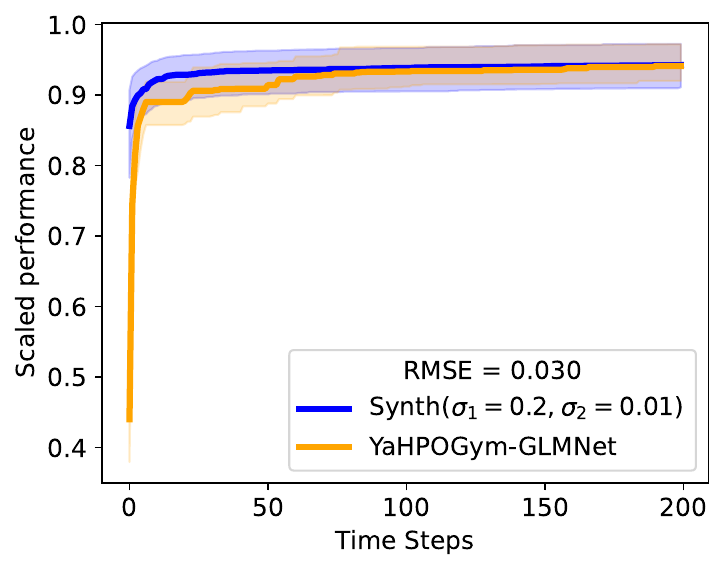}}
    & \includegraphics[clip, trim=0.8cm 0cm 0.0cm 0.0cm,height=3.5cm]{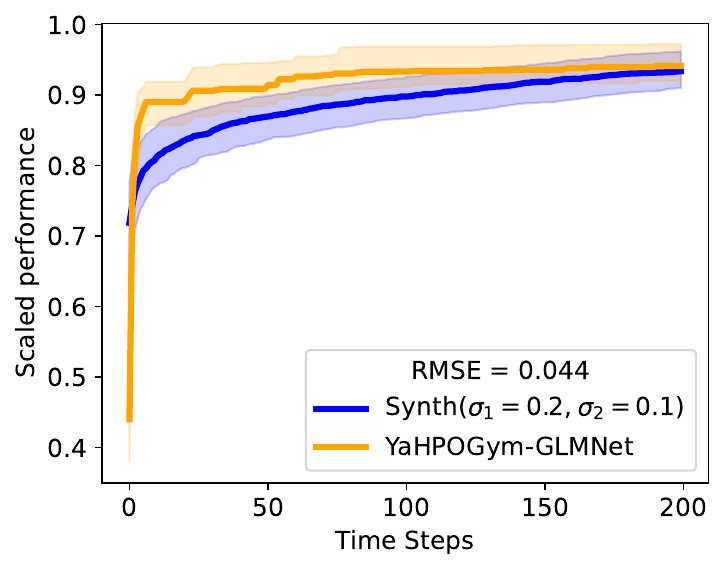}\\
    \includegraphics[clip,  trim=0.0cm 0cm 0.0cm 0.0cm,height=3.5cm]{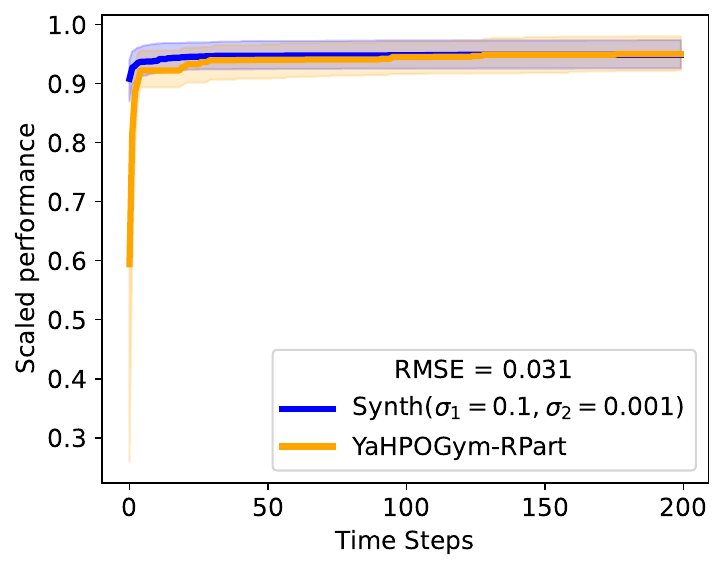}
    & \fbox{ \includegraphics[clip, trim=0.8cm 0cm 0.0cm 0.0cm,height=3.5cm]{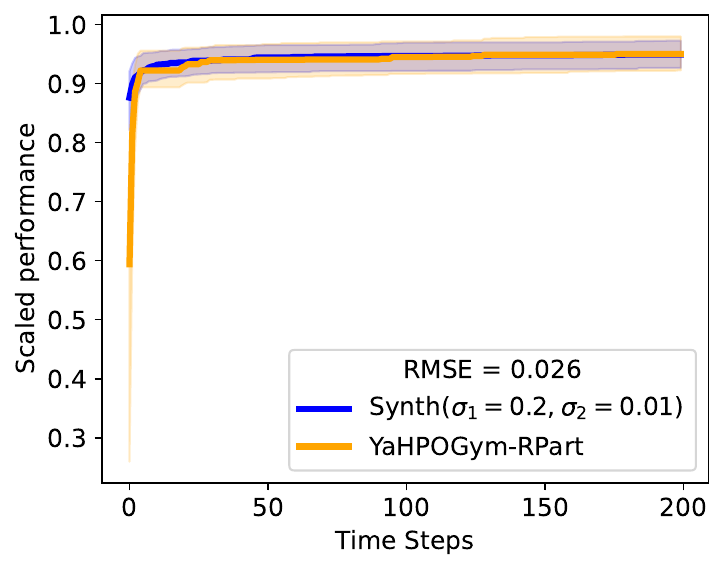}}
    &  \includegraphics[clip, trim=0.8cm 0cm 0.0cm 0.0cm,height=3.5cm]{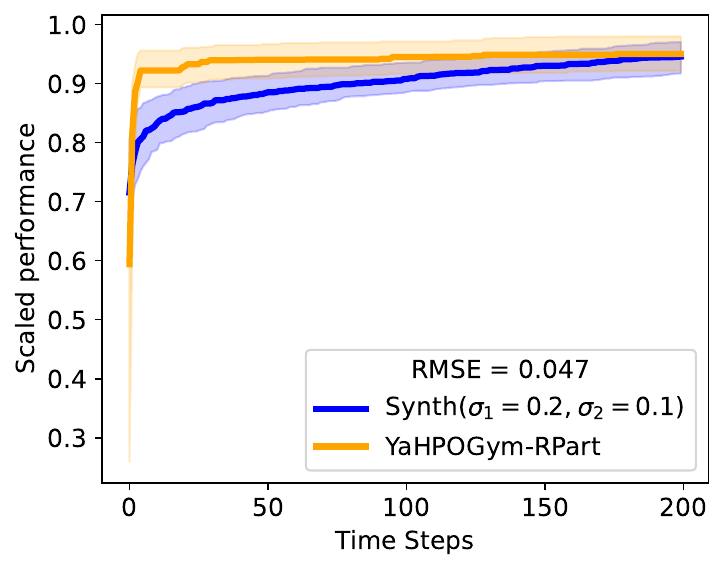}\\
    \includegraphics[clip,  trim=0.0cm 0cm 0.0cm 0.0cm,height=3.5cm]{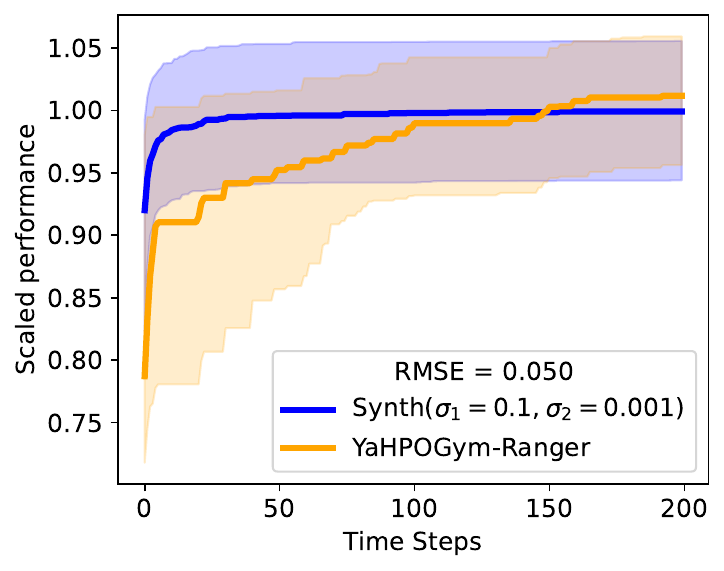}
    & \includegraphics[clip, trim=0.8cm 0cm 0.0cm 0.0cm,height=3.5cm]{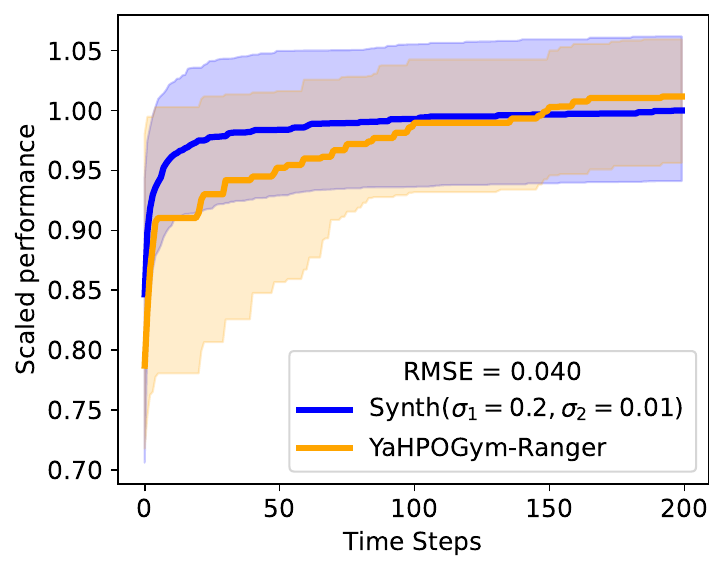}
    &  \fbox{ \includegraphics[clip, trim=0.8cm 0cm 0.0cm 0.0cm,height=3.5cm]{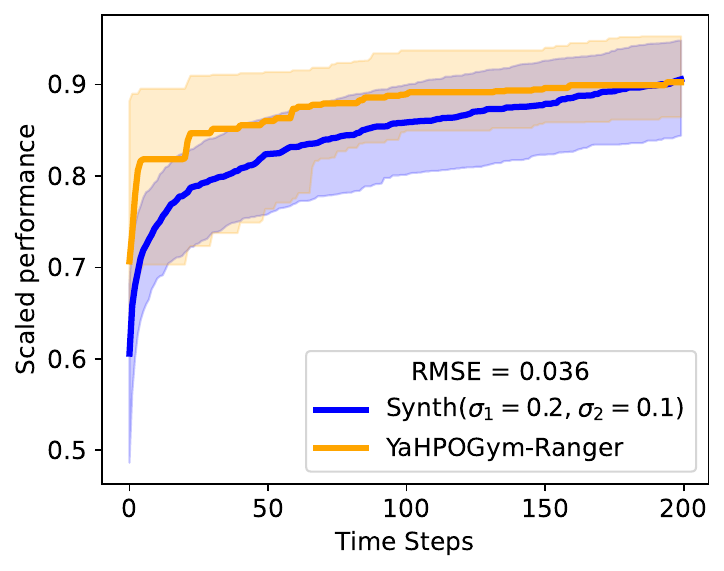}}\\
    \includegraphics[clip,  trim=0.0cm 0cm 0.0cm 0.0cm,height=3.5cm]{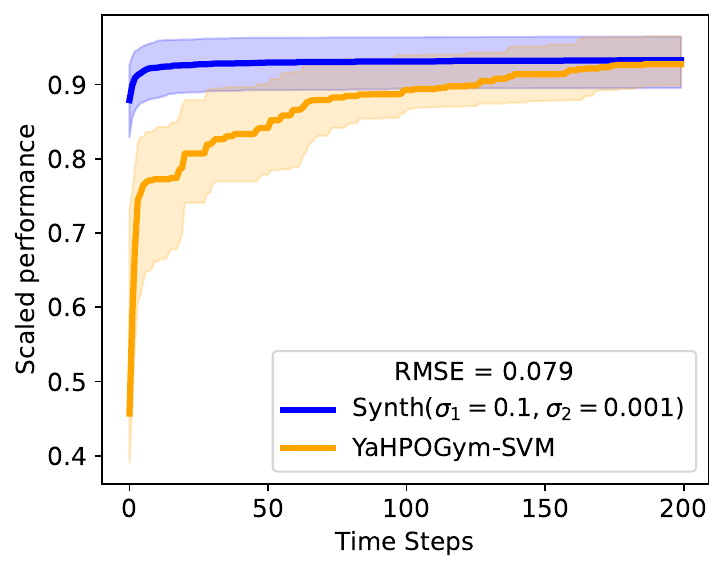}
    &\includegraphics[clip, trim=0.8cm 0cm 0.0cm 0.0cm,height=3.5cm]{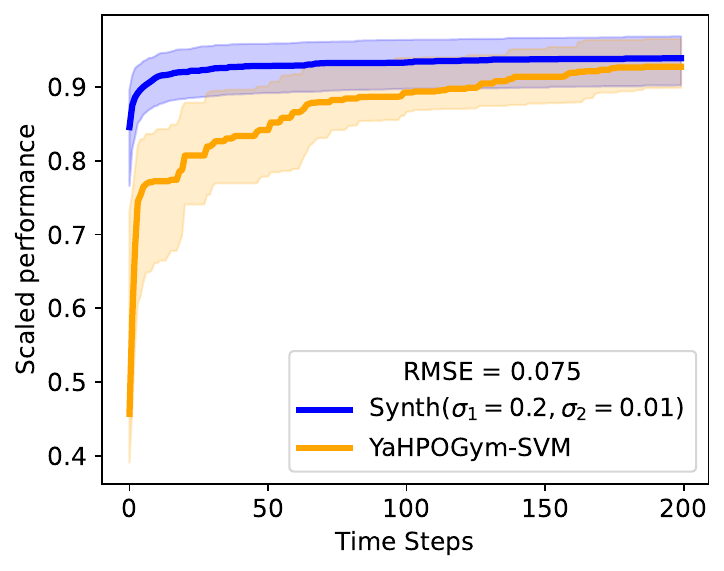}
    &  \fbox{ \includegraphics[clip, trim=0.8cm 0cm 0.0cm 0.0cm,height=3.5cm]{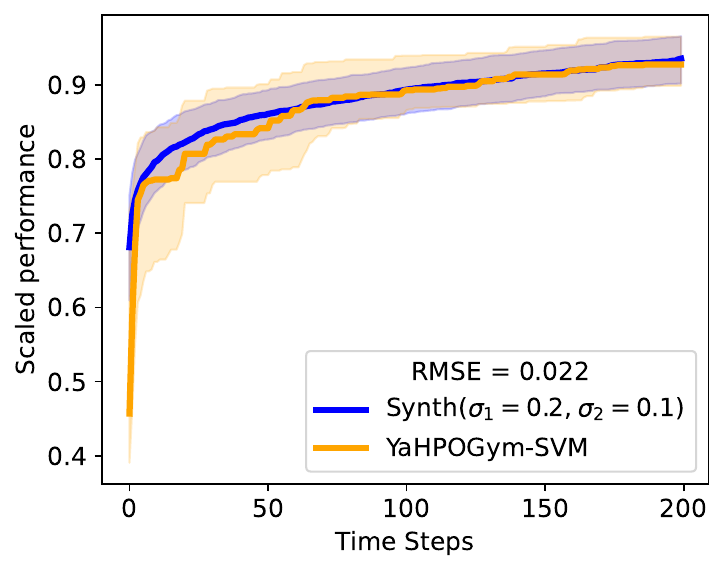}}\\
    \includegraphics[clip,  trim=0.0cm 0cm 0.0cm 0.0cm,height=3.5cm]{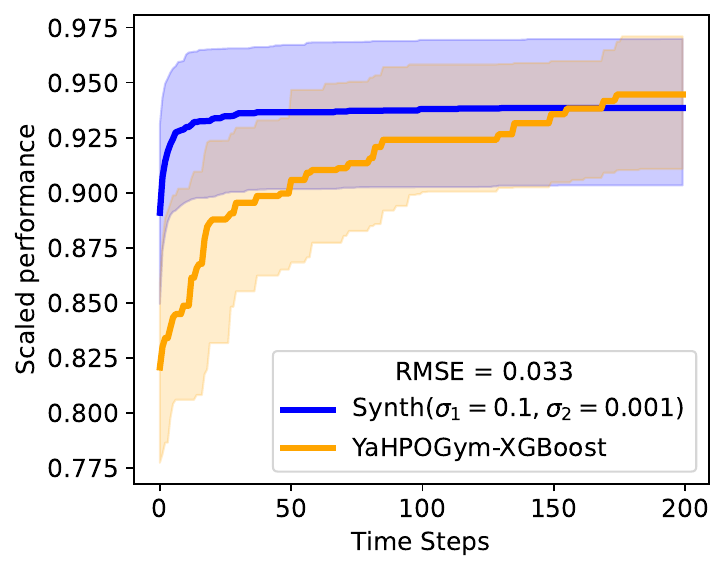}
    &\includegraphics[clip, trim=0.8cm 0cm 0.0cm 0.0cm,height=3.5cm]{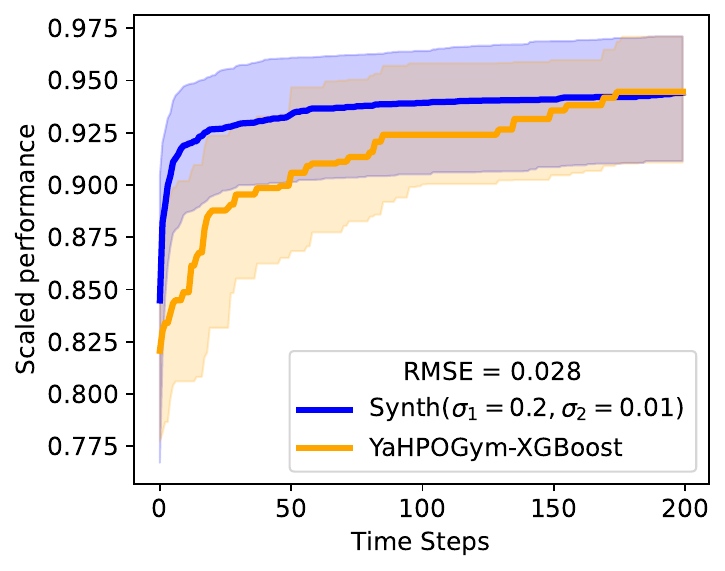}
    &  \fbox{ \includegraphics[clip, trim=0.8cm 0cm 0.0cm 0.0cm,height=3.5cm]{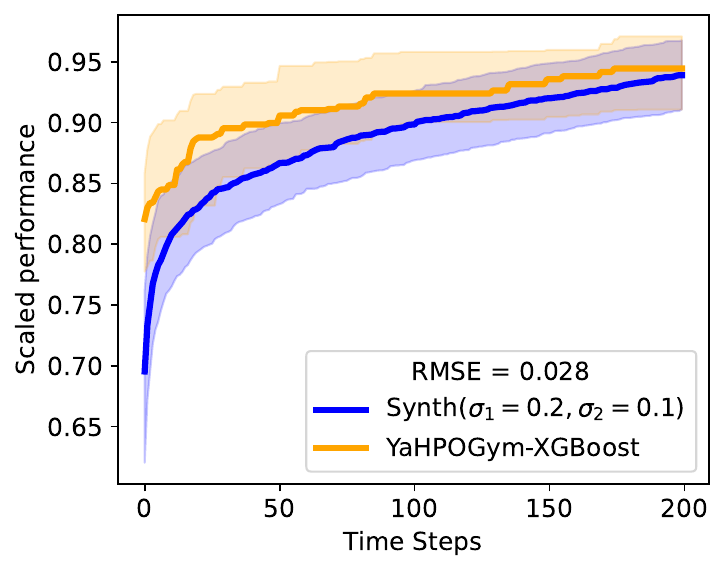}}\\
\end{tabular}
\caption{Priors for \Yahpogym{}}
\label{app:prior_selection_yahpogym}
\end{figure*}

\paragraph{Priors for \OurAlgoPFNs{}.} Our methodology allows using different priors for each arm and benchmark. We use this advantage for \OurAlgoPFNs{}, as shown in Table~\ref{app:tab:prior_types_PFNs}. We note that the choice of priors is critical, since they should match expected real-world data while simultaneously impacting the exploration behavior of the agent. For example, choosing a \textit{curved} prior instead of a \textit{semi-flat} or \textit{flat} one for the same arm leads to increased exploration of that arm (independently of whether this is the optimal choice for this arm). This implies that arms with a \textit{curved} prior will be sampled more often regardless of the true trajectory type. If such arms are consistently optimal, overall performance will improve. However, from another perspective, the possibility of prioritizing arms that are more likely to be optimal can be important. For instance, in \Tabreporaw{}, XGBoost has flat trajectories while it outperforms other methods more frequently. As a result, assigning a \textit{flat} prior to XGBoost reduces overall performance. For our evaluation, we selected the prior that most closely matched the trajectory shape on the holdout datasets, as measured by the root mean square error (RMSE), as seen in Figure \ref{app:prior_selection_tabreporaw} and 
\ref{app:prior_selection_yahpogym}, but we emphasize that investigating this trade-off is a research direction for future work.

\begin{table}[h]
\centering
\begin{tabular}{|c|c|c|}
\hline
\textbf{Benchmark} & \textbf{Model Class (Arm)} & \textbf{PFN Type} \\
\hline
\multirow{5}{*}{Complex} 
& XTab & semi-flat \\
& FLAML & semi-flat \\
& RealMLP & flat \\
& TabForestPFN & semi-flat \\
& TabPFN\_v2 & semi-flat \\
\hline
\multirow{7}{*}{TabRepoRaw} 
& CatBoost & semi-flat \\
& ExtraTrees & semi-flat \\
& LightGBM & curved \\
& NeuralNet(FastAI) & curved \\
& NeuralNet(Torch) & curved \\
& RandomForest & semi-flat \\
& XGBoost & flat \\
\hline
\multirow{6}{*}{YaHPOGym} 
& AKNN &  curved \\
& GLMNet & semi-flat  \\
& RPart & semi-flat \\
& Ranger & curved \\
& SVM & curved \\
& XGBoost & curved \\
\hline
\end{tabular}
\caption{PFNs types for each arm across different benchmarks.}
\label{app:tab:prior_types_PFNs}
\end{table}

\section{Results in details}
\label{app:more_results}

\begin{figure*}[htb]
    \begin{tabular}{c c c c c}
    \hspace{-0.5em}\small\Complex{}
    & \hspace{-1em}\small\Tabreporaw{[SMAC]}
    &\hspace{-1.5em}\small\Yahpogym{[SMAC]}
    & \\
    \includegraphics[clip, trim=0.2cm 0cm 0.2cm 0.25cm ,height=4.2cm]{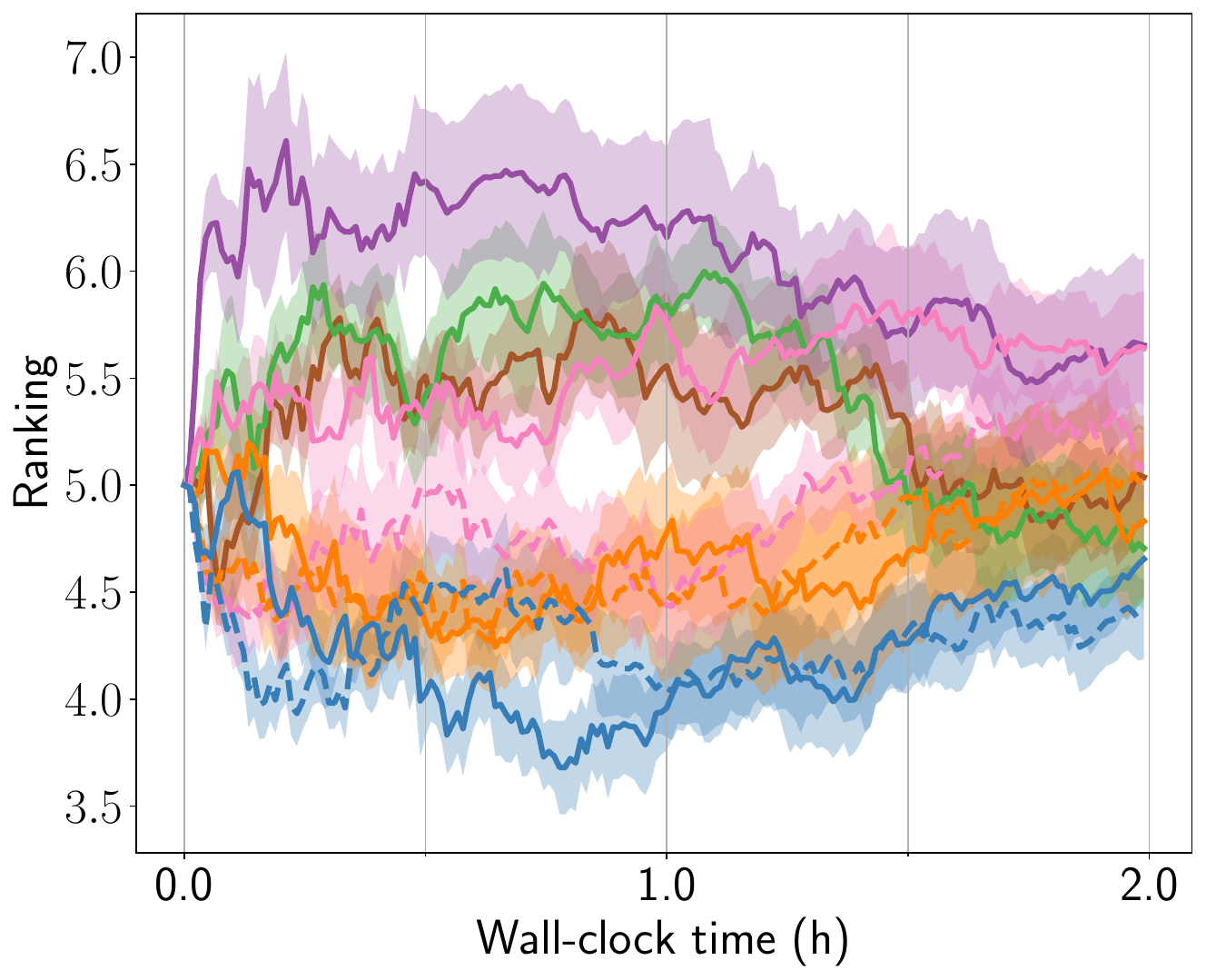}
    & \hspace{-1em} \includegraphics[clip, trim=1.1cm 0cm 0cm 0.25cm, height=4.2cm]{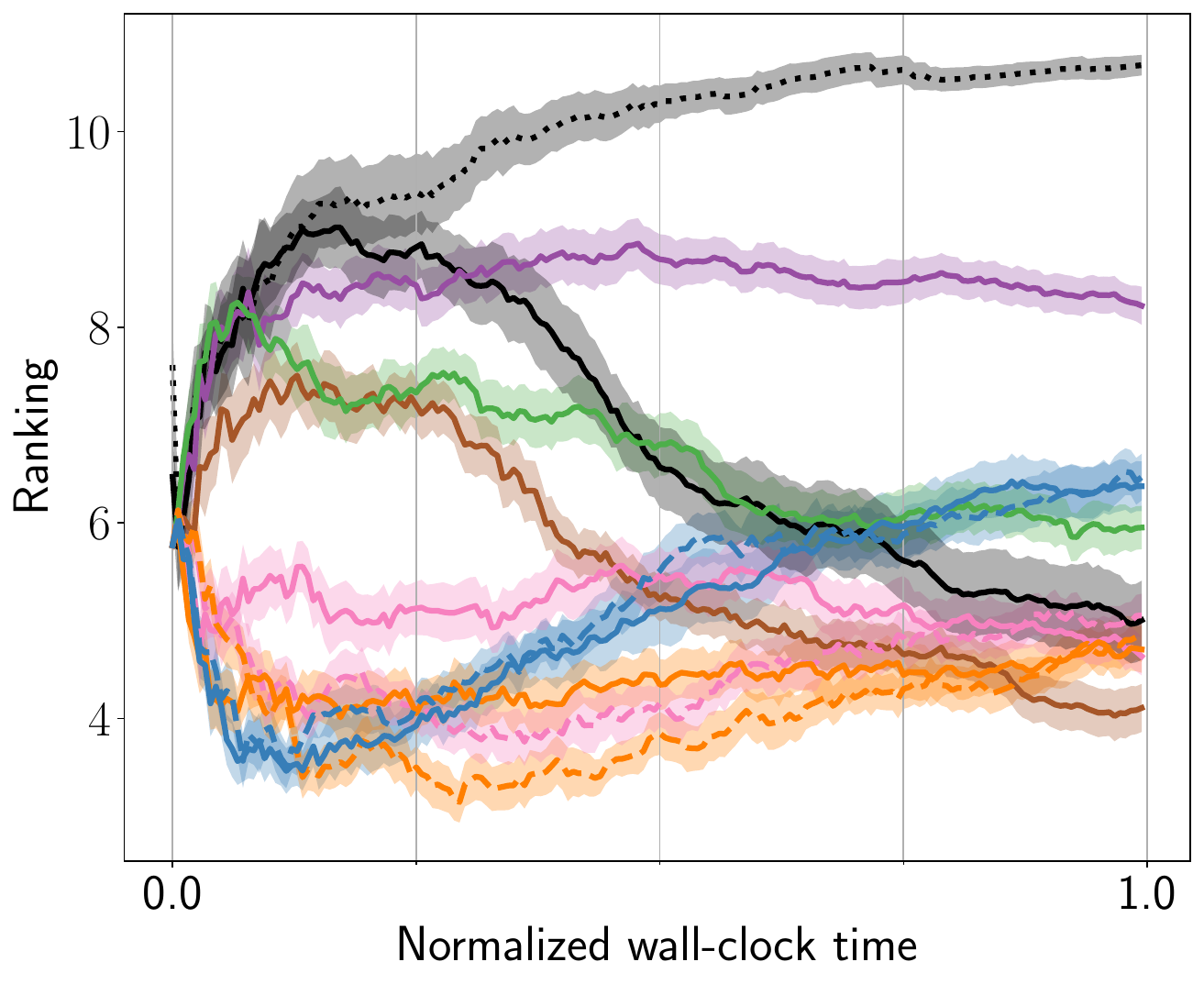}
    &  \hspace{-1em}\includegraphics[clip, trim=1.1cm 0cm 0cm 0.25cm, height=4.2cm]{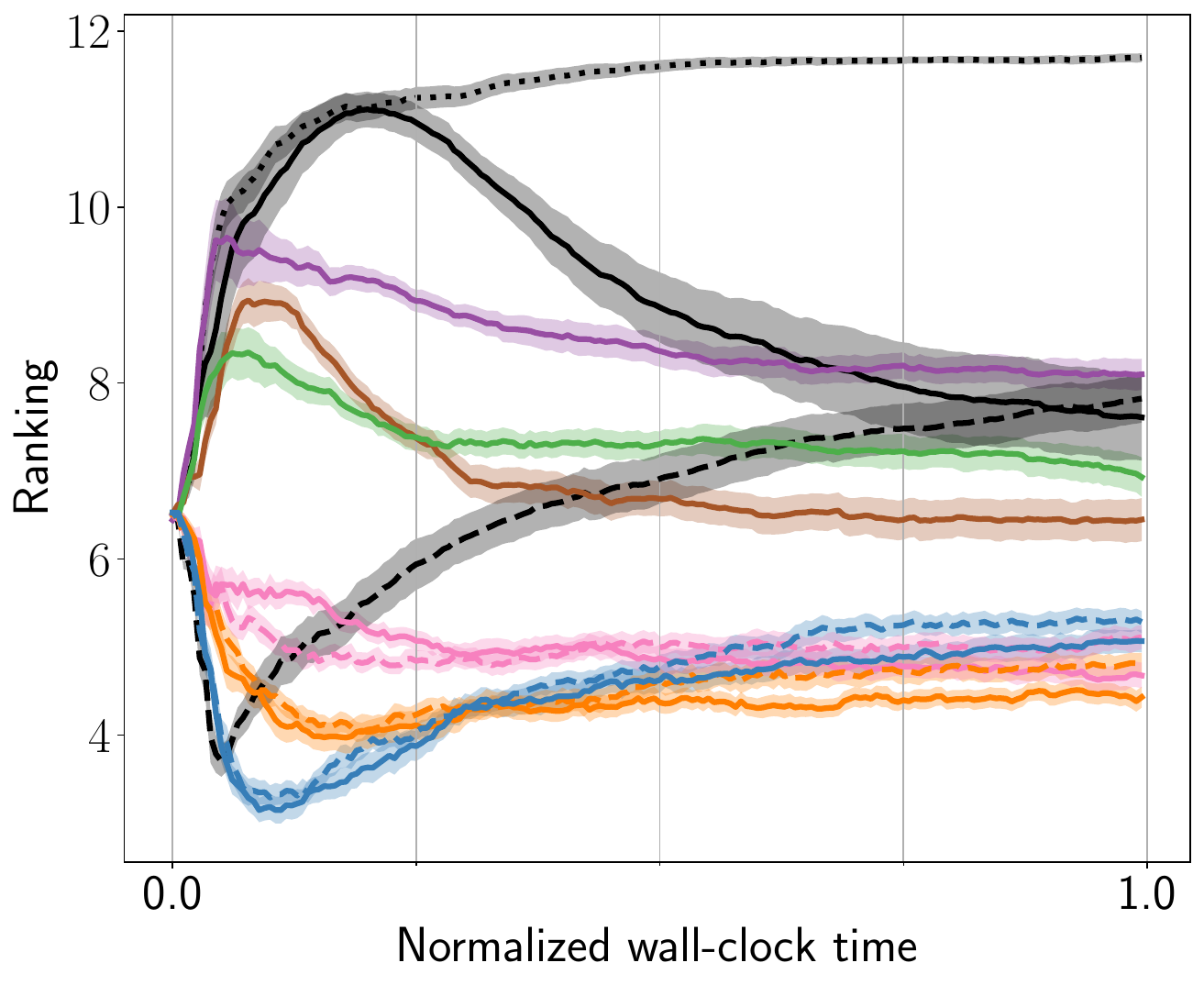}
    &\hspace{-1.4em} \includegraphics[clip, trim=0.3cm -3.5cm 0.0cm 0.0cm, height=3.8cm]{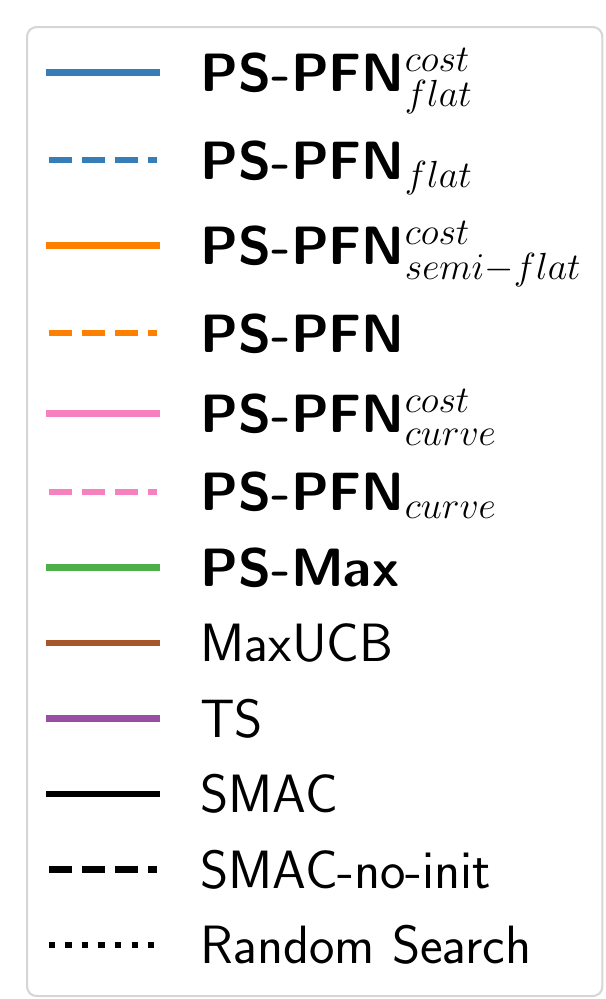}
    \end{tabular}
\caption{Ranking of algorithms on different benchmarks, lower is better. \SMAC{} and \randomsearch{} perform \combinedsearch{} across the joint space.}
\label{app:fig:average_ranking_different_priors}

\end{figure*}

\begin{figure*}[htb]
    \begin{tabular}{c c c c c}
    \hspace{-0.5em}\small\Complex{}
    & \hspace{-1em}\small\Tabreporaw{[SMAC]}
    &\hspace{-1.5em}\small\Yahpogym{[SMAC]}
    & \\
    \includegraphics[clip, trim=0.2cm 0cm 0.2cm 0.25cm ,height=4.2cm]{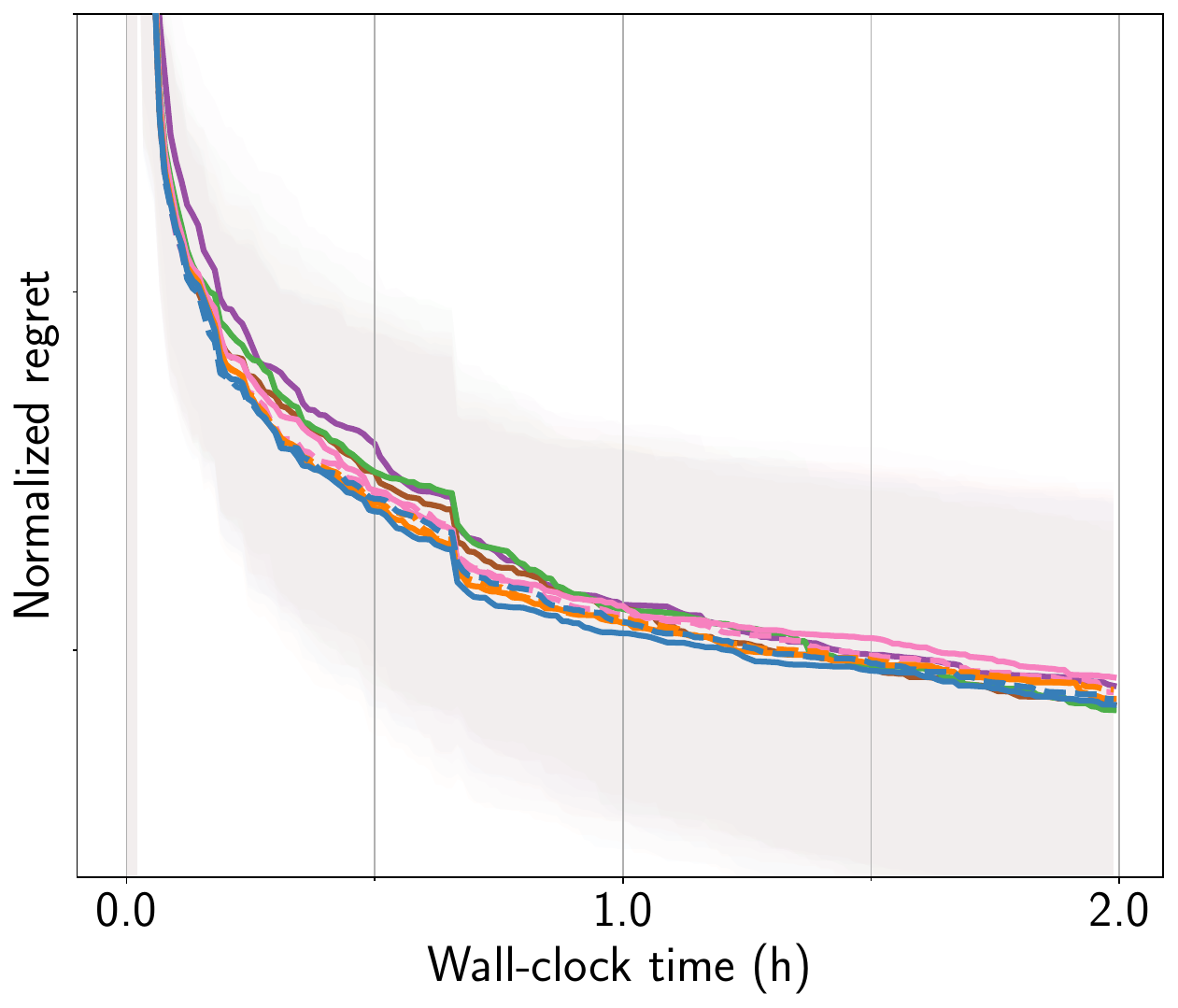}
    & \hspace{-1em} \includegraphics[clip, trim=1.1cm 0cm 0cm 0.25cm, height=4.2cm]{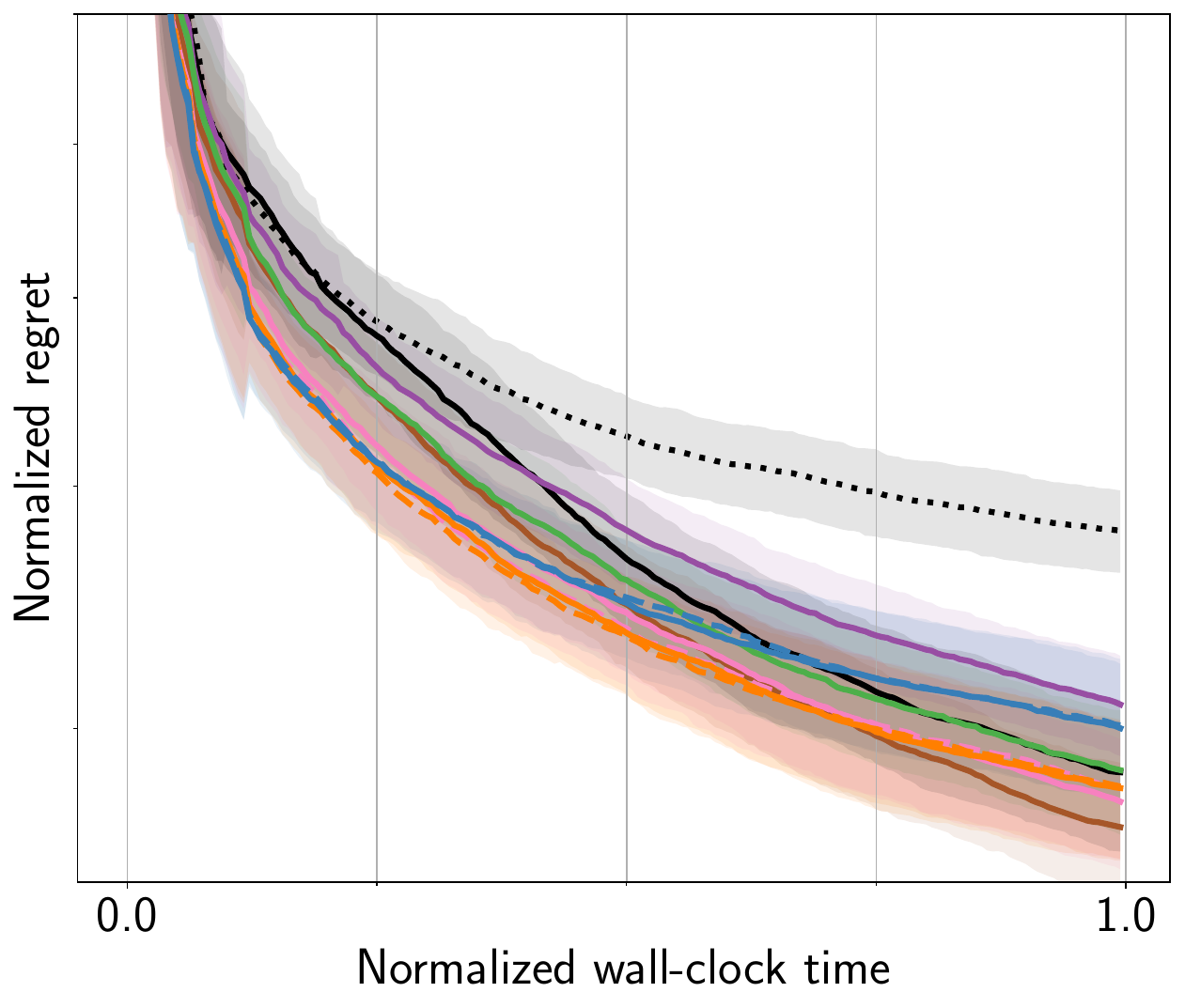}
    &  \hspace{-1em}\includegraphics[clip, trim=1.1cm 0cm 0cm 0.25cm, height=4.2cm]{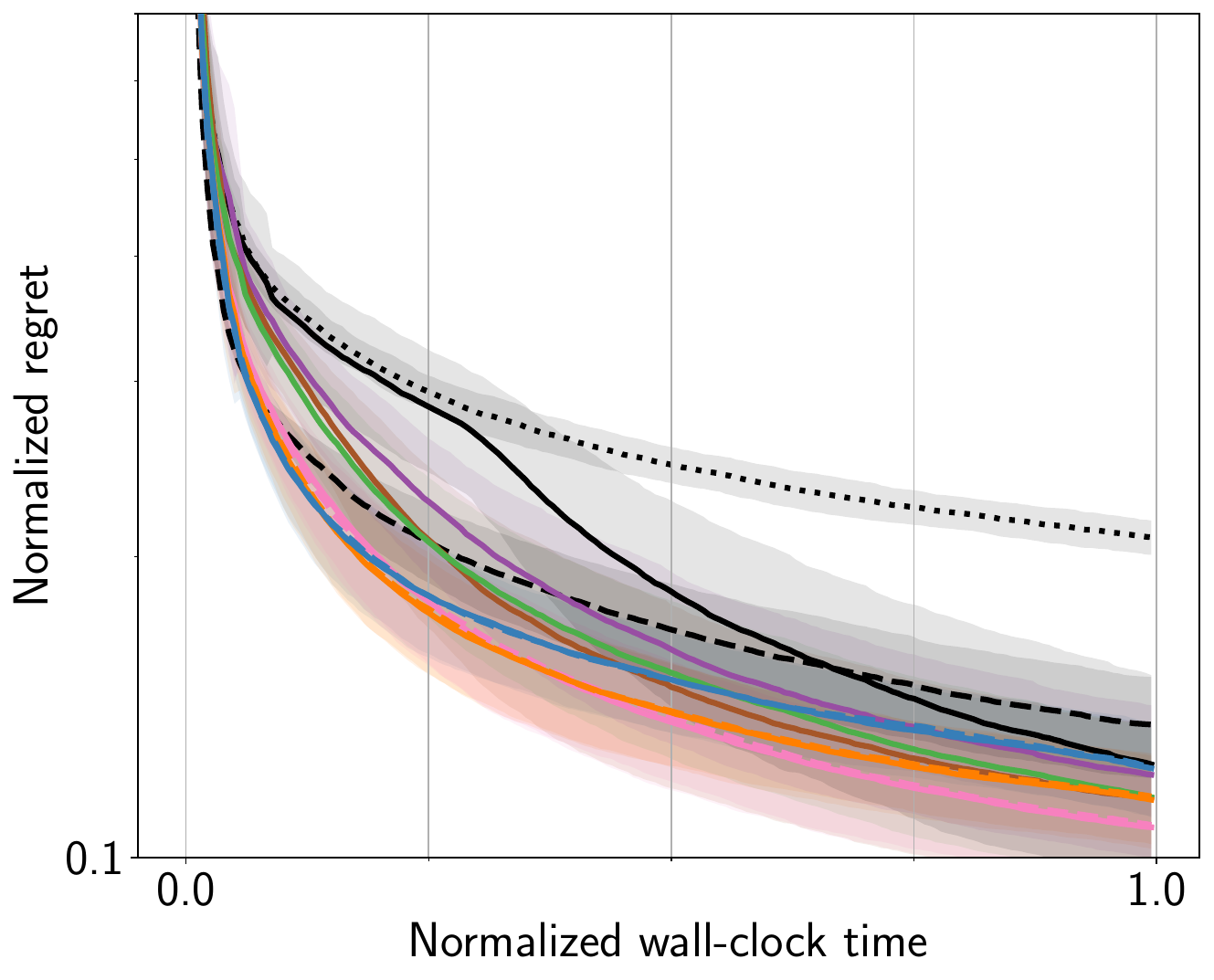}
    &\hspace{-1.4em} \includegraphics[clip, trim=0.3cm -3.5cm 0.0cm 0.0cm, height=3.8cm]{Extras_YaHPOGym_100_legend.pdf}
    \end{tabular}
\caption{Regret of algorithms on different benchmarks, lower is better. \SMAC{} and \randomsearch{} perform \combinedsearch{} across the joint space.}
\label{app:fig:average_regret_different_priors}
\end{figure*}

We compare the performance of \OurAlgo{} under different priors by showing the ranking plot in Figure~\ref{app:fig:average_ranking_different_priors} and the normalized loss (regret) over time in Figure~\ref{app:fig:average_regret_different_priors}.


\begin{figure*}[htb]
    \begin{tabular}{c c c c c}
    \hspace{-0.5em}\small\Complex{}
    & \hspace{-1em}\small\Tabreporaw{[SMAC]}
    &\hspace{-1.5em}\small\Yahpogym{[SMAC]}
    & \\
    \includegraphics[clip, trim=0.2cm 0cm 0.2cm 0.25cm ,height=4.3cm]{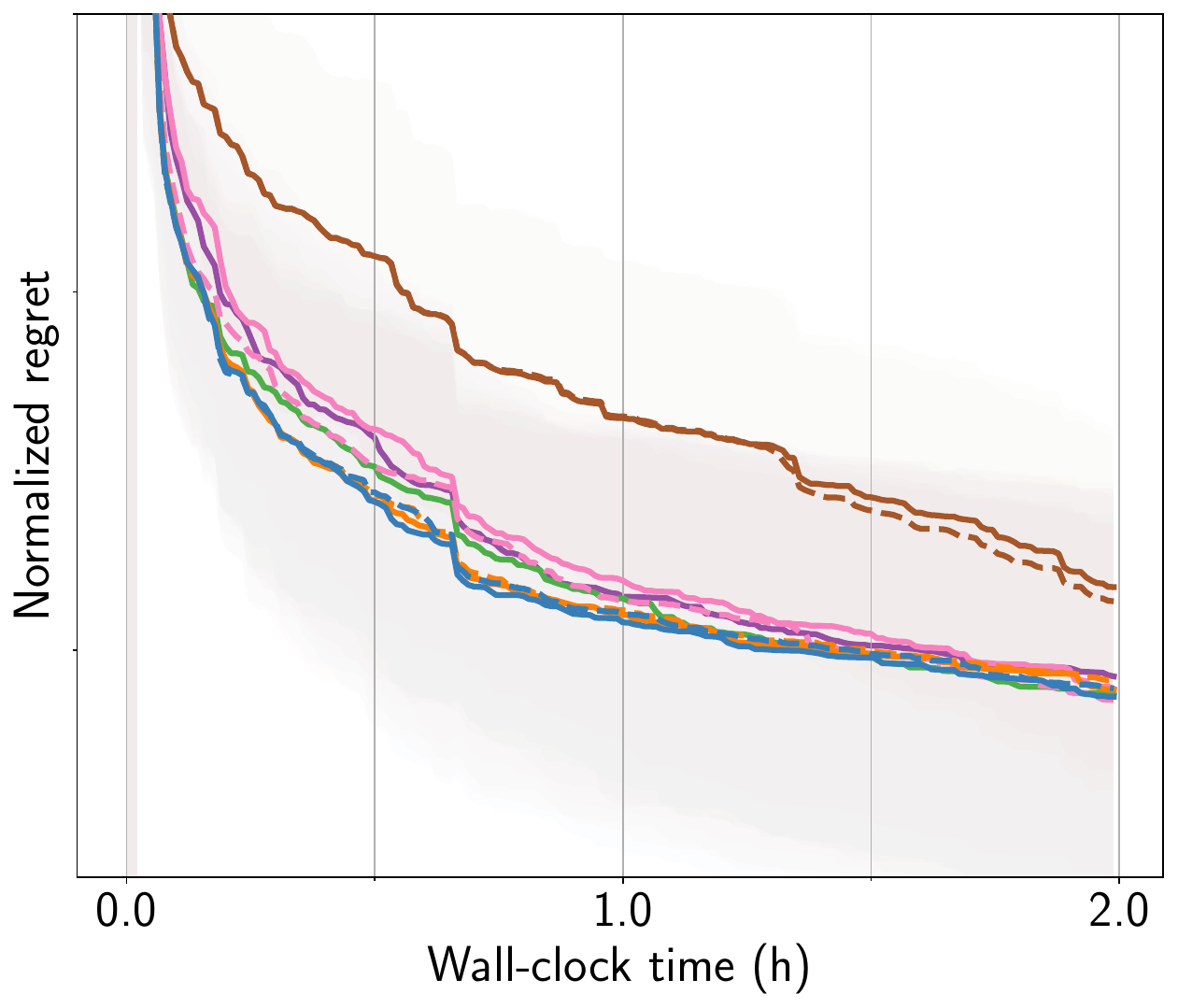}
    & \hspace{-1em} \includegraphics[clip, trim=1.1cm 0cm 0cm 0.25cm, height=4.3cm]{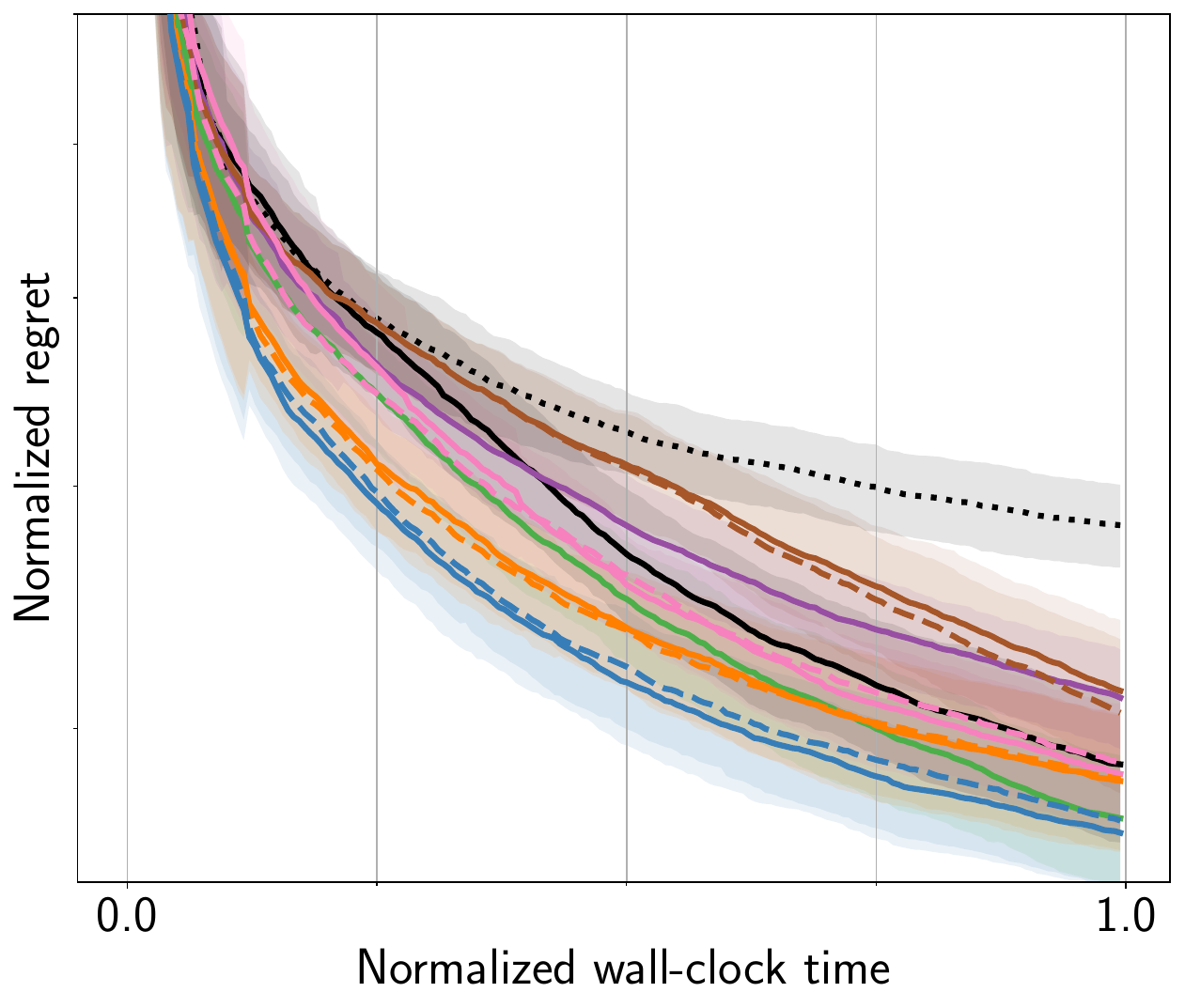}
    &  \hspace{-1em}\includegraphics[clip, trim=1.1cm 0cm 0cm 0.25cm, height=4.3cm]{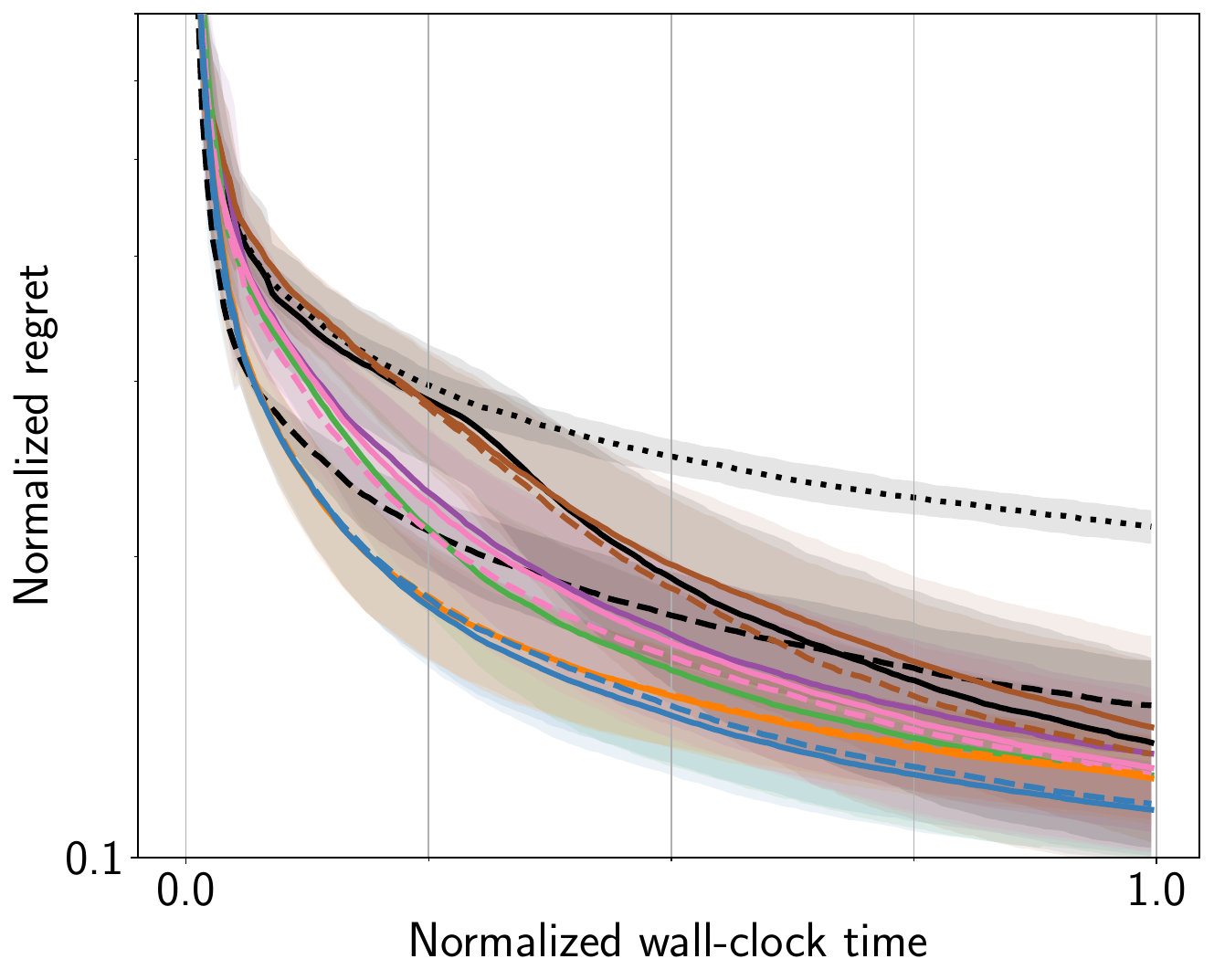}
    &\hspace{-1.4em} \includegraphics[clip, trim=0.3cm -3.5cm 0.0cm 0.0cm, height=4.2cm]{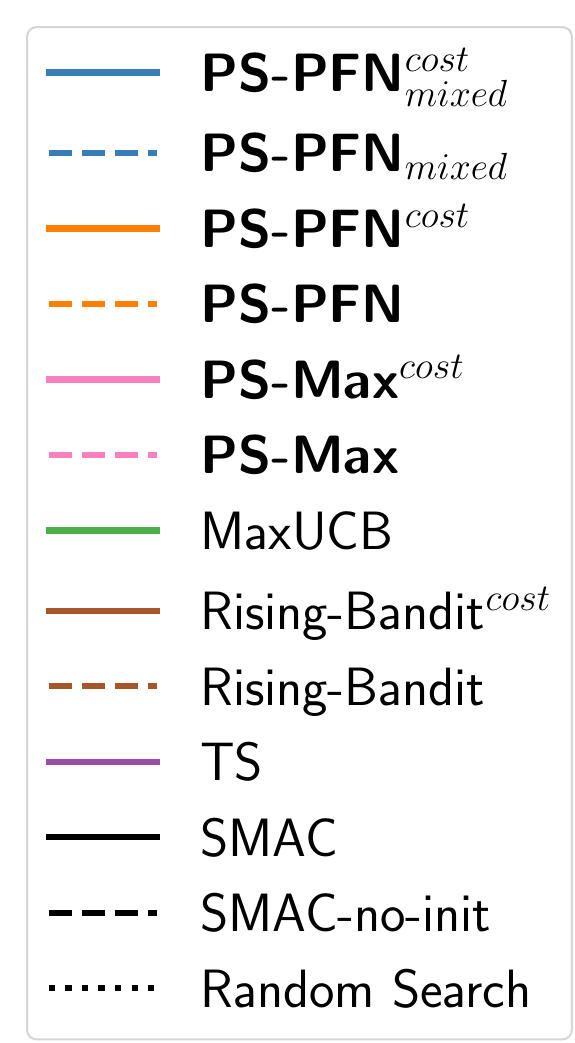}
    \end{tabular}
\caption{Regret of algorithms on different benchmarks, lower is better. \SMAC{} and \randomsearch{} perform \combinedsearch{} across the joint space.}
\label{app:fig:average_regret}
\end{figure*}

\begin{figure}[htb]
\centering
\includegraphics[height=5cm]{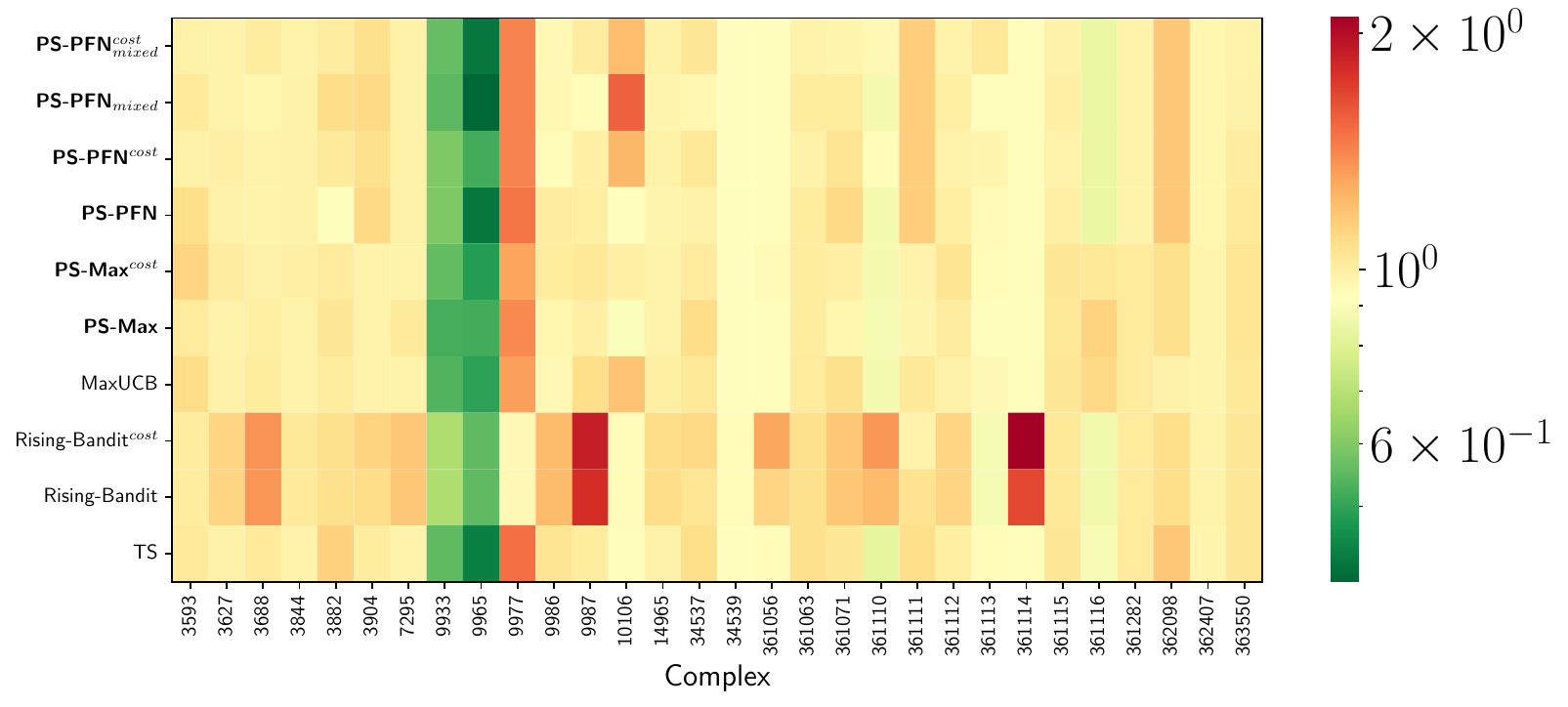}
\caption{Heatmap of regret of algorithms on different datasets of \Complex{} benchmark, lower (green) is better. }
\label{app:fig:heamap_regret_Complex}
\end{figure}

\begin{figure}[htb]
\centering
\includegraphics[height=9cm]{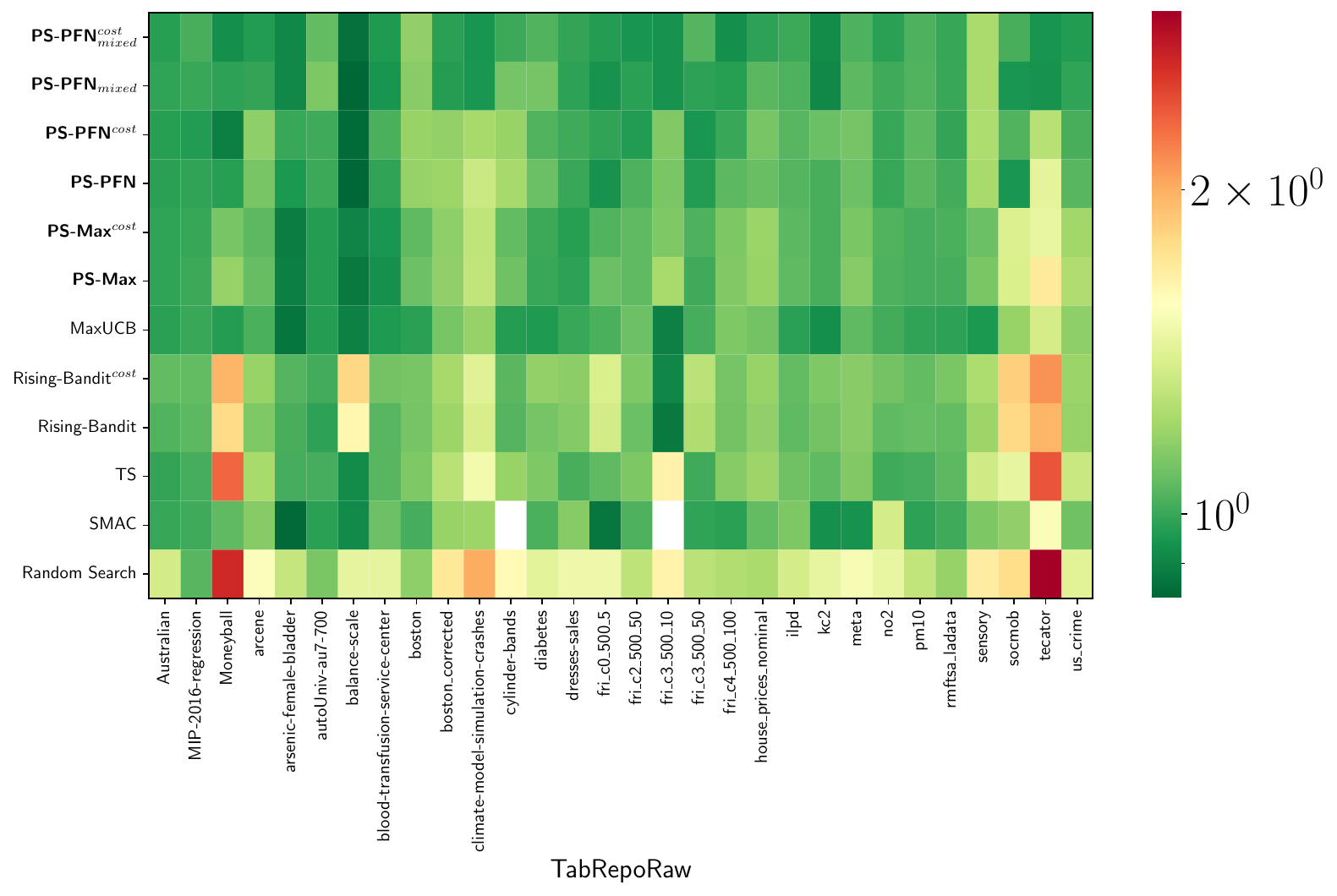}
\caption{Heatmap of regret of algorithms on different datasets of \Tabreporaw{} benchmark, lower (green) is better. }
\label{app:fig:heamap_regret_TabRepoRaw}
\end{figure}

\begin{figure}[htb]
\centering
\includegraphics[height=3.5cm]{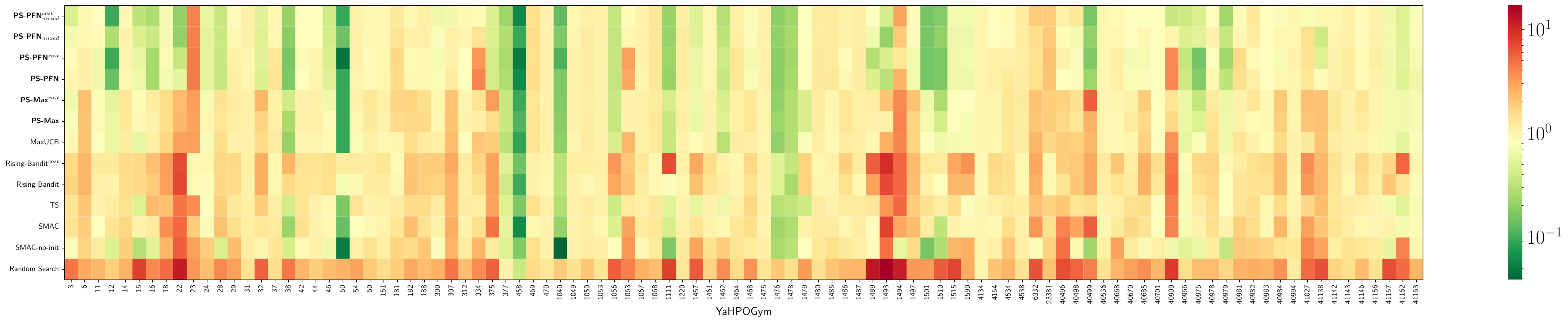}
\caption{Heatmap of regret of algorithms on different datasets \Yahpogym{} benchmark, lower (green) is better. }
\label{app:fig:heamap_regret_YaHPOGym}
\end{figure}

In Figure~\ref{app:fig:average_regret}, we show the normalized loss or regret over time. Figure~\ref{app:fig:heamap_regret_Complex} displays the regret heatmap for each dataset in the \Complex{} benchmark. Similarly, Figures~\ref{app:fig:heamap_regret_TabRepoRaw} and~\ref{app:fig:heamap_regret_YaHPOGym} present the regret heatmaps for the \Tabreporaw{} and \Yahpogym{} benchmarks, respectively.

\end{document}